\documentclass{article}

\usepackage[eandd, final]{neurips_2026}

\usepackage[utf8]{inputenc} 
\usepackage[T1]{fontenc}    
\usepackage{hyperref}       
\usepackage{url}            
\usepackage{booktabs}       
\usepackage{amsfonts}       
\usepackage{nicefrac}       
\usepackage{microtype}      
\usepackage{xcolor}         
\usepackage{natbib}
\usepackage{graphicx}
\usepackage{algorithm}
\usepackage{algorithmic}
\usepackage{amsmath}
\usepackage{amssymb}
\usepackage{array}
\usepackage{bm}
\usepackage{caption}
\usepackage{multirow}
\usepackage{needspace}
\usepackage{paracol}
\usepackage{ragged2e}
\usepackage{wrapfig}
\usepackage{makecell}
\usepackage{subcaption}

\newcommand{\xhdr}[1]{\noindent\textbf{#1.}}
\definecolor{tred}{RGB}{227, 11, 92}

\title{Aligning Language Model Benchmarks\\ with Pairwise Preferences}

\author{
Marco Gutierrez\\
University of Virginia\\
\texttt{sgw3fy@virginia.edu}
\And
Xinyi Leng\\
University of Virginia\\
\texttt{eam5kk@virginia.edu}
\And
Hannah Cyberey\\
University of Virginia\\
\texttt{yc4dx@virginia.edu}
\And
Jonathan Richard Schwarz\\
Imperial College London\\
\texttt{jschwarz@imperial.ac.uk}
\And
Ahmed Alaa\\
UC Berkeley \& UCSF \\
\texttt{amalaa@berkeley.edu}
\And
Thomas Hartvigsen\\
University of Virginia\\
\texttt{hartvigsen@virginia.edu}\\
\\
}

\begin{document}

\maketitle

\begin{abstract}
Language model benchmarks are pervasive and computationally-efficient proxies for real-world downstream performance. However, many recent works find that benchmarks often fail to predict downstream utility. While some works have begun diagnosing sources of misalignment, there remain no ways to systematically update benchmarks to align their scores with downstream usage. 
Towards bridging this gap, we introduce and study \textit{benchmark alignment}, where we use information about downstream model performance to automatically update benchmarks, specifically aiming to update static benchmarks so they generalizably rank models according to new pairwise preferences.
Our experiments involving 4576 language models and 6 benchmarks show that reweighting benchmark items can successfully rank unseen models, even generalizing across model scales in most cases.
And while naive alignment unsurprisingly requires large numbers of models and benchmark questions, an oracle experiment suggests this could be reduced to as few as 20 well-chosen models.
Overall, our work takes a step towards efficiently aligning benchmark development with downstream tasks.\footnote{All of our code, models, and data are publicly-available: \url{https://github.com/hartvigsen-group/benchalign}.}
%


\end{abstract}

\section{Introduction}\label{sec:intro}
Benchmarks are the standard way to track progress in language modeling, operating on the assumption that aggregated test scores directly predict downstream performance. For example, the top-scoring model on a \textit{language understanding benchmark} is presumed to be the \textit{best language understanding model}.
However, many recent works have found that the best models on a benchmark often underperform in deployment, even for the same task~\citep{Alaa2025MedicalValidity, Salaudeen2025ImageNot:Rankings, Zhang2025Train-before-TestRankings}. Similarly, general-purpose models, despite excelling on general-capability benchmarks, often perform worse than smaller task-specific models~\citep{Pecher2025ComparingPerformance, Zhou2025GeneralistOculomics}, suggesting that benchmark scores are poor predictors of real-world utility. 


Recent works have begun studying ways to improve benchmarks, many of which refine benchmarks to better capture model capabilities. 
These methods select informative subsets of test items ~\citep{Polo2024TinyBenchmarks:Examples,vivek-etal-2024-anchor,Kipnis2025Metabench:Models}, often drawing from psychometrics using item response theory (IRT)~\citep{vania-etal-2021-comparing,hofmann2025fluid, Kipnis2025Metabench:Models}.
The resulting subsets can then be inspected for patterns that explain what the benchmark measures. While understanding the signals captured by benchmarks is important, these approaches leave two important, open questions: 1) \textit{Do their rankings reflect real-world utility?} and 2) \textit{Can they be updated to better predict utility for specific downstream uses?}


Downstream utility is extremely challenging to estimate.
One common approach is to use pairwise preferences from downstream users: Given two models, practitioners choose which they prefer for their specific task~\citep{Wang2023HelpSteer:SteerLM, Cui2024UltraFeedback:Feedback}. 
For benchmarking, the most important signal is how models rank against one another rather than the raw performance. 
For example, in high-stakes settings like healthcare, models that score well on standard benchmarks can still produce outputs that are inaccurate or unsafe when evaluated against clinician preferences~\citep{Arora2025HealthBench:Health, Busch2025CurrentReview}. Yet there remains no systematic way to assess whether and how existing benchmarks can be aligned to reflect these preferences.


We introduce \textit{benchmark alignment}: given a benchmark dataset and a set of language models with an externally specified preference ranking (e.g., derived from human evaluations), we ask whether existing benchmarks can be reweighted to induce rankings that better reflect these preferences while generalizing to new models. This task is inherently challenging, as most existing benchmarks are not designed to encode external preferences and are static, even if they fail to predict utility. To study whether benchmark alignment is feasible, we propose a simple learning-to-rank approach as a diagnostic tool. Our approach learns preference-aligned weights for individual test items by predicting model pairwise rankings. These weights are interpretable and quantify the relevance of each test item to the target preferences. The reweighted benchmark items then produce an aggregate score that can be used to rank unseen models. Because this method is straightforward, it lets us examine relationships between alignment, preference data volume, and model selection.

In experiments spanning 4576 models, 6 benchmarks and 7 reward models, we find that benchmark alignment is surprisingly feasible. Aligned benchmarks accurately rank models unseen during training, including substantially larger models (30-70B+ parameters), across preferences of helpfulness and honesty.
However, large volumes of models and questions are required to achieve alignment naively, highlighting data requirements as a key challenge for benchmark alignment. Interestingly, we find that an oracle can use as few as 20 models to achieve high-quality alignment, suggesting that identifying them ahead of time could be an important future step.


In summary, our contributions are as follows:
\vspace{-8pt}
\begin{itemize}
    \item We propose aligning benchmarks with pairwise preferences, alongside a straightforward method to study benchmark alignability (Section~\ref{sec:proposed_method}).
    \item We show that benchmark alignment is feasible using only small and medium-sized model data, with re-weighted benchmarks generalizing rankings to substantially larger models (Section~\ref{sec:perf_model_size_splits})
    \item We find that naive alignment is data-intensive, requiring around 1,000 models and 5,000 questions, though as few as 20 carefully chosen models can suffice (Section~\ref{sec:perf_num_models}).
    \item We show that benchmark distillation methods can reduce the data requirements for alignment to using only 1,500 questions ($\sim$7\% of OpenLLMLeaderboard), with minimal loss in ranking accuracy. (Section~\ref{sec:perf_num_questions})
\end{itemize}

\section{Related Work}\label{sec:related_work}
\xhdr{Benchmark Validity and Utility} Static benchmarks have long been the standard for assessing model capabilities and tracking progress in the field. However, numerous studies have raised concerns about their \emph{reliability}--whether they yield consistent results--and \emph{validity}--whether they measure what they purport to measure~\citep{bowman-dahl-2021-will,subramonian-etal-2023-takes,reuel2024betterbench,saxon2024benchmarks}. In particular, current benchmarks often lack \emph{external validity} (or predictive validity) where improvements in benchmark performance do not reliably translate into improved real-world utility~\citep{liao2021we}. For instance, \citet{Alaa2025MedicalValidity} empirically show that performance on MedQA, a benchmark intended to evaluate medical knowledge, correlates weakly with models' actual clinical performance. This evaluation gap undermines the use of benchmarks for model selection, as practitioners cannot confidently assume that higher benchmark rankings correspond to better use experiences or downstream performance.

\xhdr{Refining Benchmarks} To improve the predictive validity of benchmarks, recent work has proposed methods to refine existing benchmarks. Several studies apply Item Response Theory (IRT), a psychometric technique that models item difficulty and discrimination, to identify smaller subsets of informative benchmark questions that more efficiently and accurately measure model capabilities~\citep{Polo2024TinyBenchmarks:Examples,Kipnis2025Metabench:Models}. \citet{hofmann2025fluid} introduce fluid benchmarking, which combines IRT with adaptive testing to dynamically select evaluation items matched to a model's capability. While these approaches focus on improving the measurement of latent model capabilities, our work instead aligns benchmarks with downstream user preferences that may not be fully captured by capability-driven metrics.

\xhdr{Alignment Evaluation and System Ranking}
While human annotations remain the gold standard for evaluation, constructing a large-scale benchmark or leaderboard like Chatbot Arena that relies on actual human ratings is costly. The use of LLM-based judges, which uses frontier models to approximate human judgment, has become widely adopted, especially for evaluating model alignment in open-ended tasks~\citep{zheng2023judging,dubois2023alpacafarm,li2024generative,kim-etal-2024-prometheus}. To compare overall model quality, a system-level ranking is typically derived by aggregating over multiple instance-level evaluations produced by these judges~\citep{gera2025justrank}. However, prior works on LLM-based evaluators have only assessed their performance at the instance level. \citet{gao2025re} find that LLM judges that perform well on per-example evaluations can produce substantially different and sometimes misaligned system-level rankings. Moreover, LLM judges are prone to systematic biases, such as positional bias~\citep{wang-etal-2024-large-language-models-fair} and self-preference bias~\citep{liu2024llms}. Our work also targets system-level ranking, but instead of relying on LLM judges, we apply learning-to-rank methods to learn preference-aligned weightings for benchmark questions, optimizing predictive validity against human-preferred model rankings.

\section{Benchmark Alignment}\label{sec:methods}

\subsection{Problem Definition}
\label{sec:problem_definition}
Let $\mathcal{F} = \{f_{1}, f_{2}, ..., f_{K}\}$ be a set of language models and $\mathcal{D}=(Q,s)$ be a benchmark, which contains a set of textual test items $Q=\{q_1,...,q_M\}$ and a scoring function $s$.
Drawing from educational testing~\citep{hively1968universe} and psychometrics~\citep{embretson1998cognitive}, we can treat each test item as a \textit{stimulus} designed to elicit model responses that carry information about the underlying construct (e.g., model abilities) we wish to measure for a given task.

The scoring function produces a scalar score $s(f_i, Q)\in\mathbb{R}$ for model $f_i$ that measures its overall performance on the benchmark, where higher is better (e.g., accuracy). Based on the scores $s(f_i, Q)$ produced for each model $f_i\in \mathcal{F}$, we can rank the models on benchmark $\mathcal{D}$ by sorting them based on their scores: $R_{\mathcal{D}}: \mathcal{F}\rightarrow \{1, 2,...,K\}$, where $R_{\mathcal{D}}(f_i)$ is $f_i$'s rank on benchmark $\mathcal{D}$ based on its score $s(f_i, Q)$. For example, if $R_{\mathcal{D}}(f_i)=4$, then model $f_i$ is the 4\textsuperscript{th} best model.

Now suppose there exists a target pairwise preference over the models in $\mathcal{F}$ (e.g., perceived helpfulness, honesty), denoted by $f_{i} \succ_{T} f_{j}$, such that it induces a target pairwise ranking $R_{T}(f_{i})$. Our goal is to construct a new benchmark $\hat{\mathcal{D}}=(\hat{Q}, s)$ that accurately ranks models according to the target preferences, such that the induced ranking $R_{\hat{D}}(f_i) = R_{T}(f_i)$. A benchmark $\hat{\mathcal{D}}$ that ranks models according to $R_T$ is thus \textit{aligned} with the target rankings. The central question we study is: \textit{to what extent can existing benchmarks be aligned with target preferences, and under what conditions does this alignment hold?}

\subsection{Studying Alignment via Learning to Rank}
\label{sec:proposed_method}

\begin{figure*}[t]
    \centering
    \includegraphics[width=\textwidth]{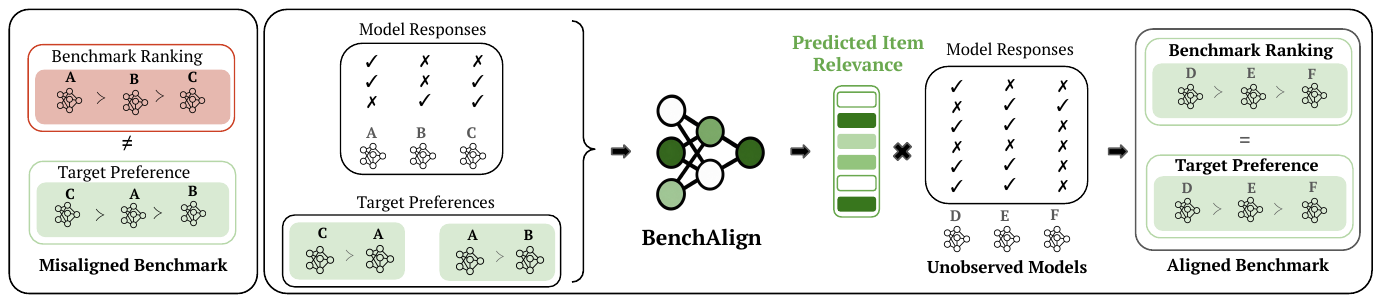}
    \caption{Overview of \textsc{BenchAlign} as a diagnostic tool for studying benchmark alignment. Given a benchmark whose ranking of target models is misaligned with a target preference, we use source models and their pairwise preferences to learn question weights via learning-to-rank. The reweighted benchmark is then evaluated to assess whether the induced ranking aligns with the target preference.}
    \label{fig:bench_align_overview}
\end{figure*}

While many approaches could be used to study alignment, the objective is not to propose a definitive solution to benchmark alignment. Instead, our goal is to examine whether preference-aligned weightings for benchmark questions can be learned from limited data, and whether the resulting rankings generalize to unseen models.
We refer to this approach as \textsc{BenchAlign}\footnote{Figure \ref{fig:bench_align_overview} provides a brief overview of benchmark alignment with \textsc{BenchAlign}.}.

\xhdr{Benchmark Construction} 
Standard benchmarks typically assume equal relevance across all test items and compute an unweighted sum or average to derive the final score~\citep{wang-etal-2024-large-language-models-fair, Hendrycks2021MeasuringDataset, Rein2023GPQA:Benchmark}. To formulate this, for a set of test items $Q$, we define the performance of a model $f_i$ as a feature vector $\bm{x}_i \in \mathbb{R}^{M}$, where the $m$-th element $x_{i,m}=\mathbf{a}(f_i, q_m)$ represents the instance-level score based on the model's response to $q$. Rather than assuming uniform relevance, we reformulate the scoring function with a continuous weight vector $\bm{\hat{w}} \in \mathbb{R}^M$, where each element $w_m$ indicates the learned relevance of test item $q_m$ to the target preference.

\xhdr{Learning-to-Rank} We approach benchmark alignment as a learning-to-rank problem~\citep{li2011short}, which aims to train a model to automatically produce an ordering of a set of objects such that a given utility function is maximized. We focus on the pairwise ranking approach, which learns to predict the relative order of a pair of objects. Pairwise comparisons simplify the evaluation to a binary classification task and have been shown to produce more consistent annotations than absolute rating scales~\citep{thurstone1927law,kiritchenko-mohammad-2017-best,carterette2008here}. Furthermore, this approach aligns with the objective of reward modeling used in reinforcement learning from human feedback, which has been shown to effectively model human preferences.\footnote{We provide further details on our algorithm in Appendix \ref{ap:methods_expanded}}.

Let $\mathcal{F}$ be the set of models accessible to obtain their output responses on test items $Q$ and their preference ranking $R_T(f_i)$. Given model $f_i\in\mathcal{F}$, we denote $\bm{x}_i\in\mathbb{R}^M$ by a feature vector containing the instance-level scores $\mathbf{a}(f_i,q)$ computed based on the model's response to all prompts $q\in Q$. We train a pairwise ranking model on inputs ($\bm{x}_i, \bm{x}_j$) for each pair of models $(f_i,f_j)$, where $f_i,f_j\in\mathcal{F}$ and $i\not=j$, minimizing the following pairwise ranking loss~\citep{Burges2005LearningDescent, Wang2018TheOptimization}:
\begin{equation}\label{eq:ltr_loss}
    \mathcal{L}(\bm{w}) = \sum^{K}_{i=1}\sum^{K}_{j=1} \mathbb{I}_{r_i < r_j}
\log\!\left(1 + e^{-(s_i - s_j)}\right)
\end{equation}
where $\mathbb{I}$ is the indicator function, and the term $\log\!\left(1 + e^{-(s_i - s_j)}\right)$ represents the individual pairwise loss $\mathcal{L}_{pair}(s_i,s_j)$. Here, $r_i$ and $r_j$ denote the target ranks for models $f_i$ and $f_j$ respectively, and $s_i = \bm{w}^{\mathsf{T}}\bm{x}_i$ is the predicted relevance score for model $f_i$ given its performance feature vector $\bm{x}_i$. The loss function $\mathcal{L}(\bm{w})$ penalizes misaligned pairs where model $f_i$ is preferred over $f_j$ (i.e., $r_i < r_j$) but the predicted relevance score indicates $s_i < s_j$.

\xhdr{Using \textsc{BenchAlign} as a Diagnostic Tool} After minimizing the pairwise ranking loss, we obtain the optimal learned weight vector $\bm{\hat{w}}$ that captures the relevance of each benchmark question to the target preferences. We use these weights to construct a reweighted benchmark $\mathcal{\hat{D}} = (\hat{Q}, s)$, which applies the reformulated scoring function $s_{\bm{\hat{w}}}(f_i, Q)$ over the original test items $Q$. We then evaluate whether the induced ranking $R_{\mathcal{\hat{D}}}$ generalizes to a target set of models $\mathcal{F}^{\prime}$ unseen during training (i.e., $\mathcal{F} \cap \mathcal{F}^{\prime} = \varnothing$). 

Specifically, we examine whether the learned weights $\bm{\hat{w}}$ transfer to unseen models, including substantially larger ones. We consider alignment feasible if $R_{\mathcal{\hat{D}}}(f^{\prime}) \approx  R_{T}(f^{\prime})$ for models in $\mathcal{F}^{\prime}$, and quantify this using Spearman rank correlation and pairwise ranking accuracy (Section~\ref{sec:setup}). We use the degree to which the alignment holds and the data required to achieve it as a measure of how easily existing benchmarks can be aligned.

\section{Experiments}\label{sec:experiments}
We empirically study the feasibility of benchmark alignment and characterize the conditions under which it succeeds. We first describe our experimental setup (Section~\ref{sec:setup}) and examine whether aligned benchmarks generalize to unseen models (Section~\ref{sec:perf_model_size_splits}). We then study the data requirements for alignment, analyzing how many training models (Section~\ref{sec:perf_num_models}) and benchmark questions (Section~\ref{sec:perf_num_questions}) are needed to achieve meaningful alignment.

\subsection{Experimental Setup}\label{sec:setup}
We describe the datasets, target preferences, baselines, and metrics used across all experiments.

\xhdr{Benchmarks and Language Models}
We primarily use the question-level responses from OpenLLMLeaderboard~\citep{Fourrier2024OpenV2}, which contains responses from 4576 models on 6 standard benchmarks: Big Bench Hard~\citep{Suzgun2022ChallengingThem}, MMLU Pro~\citep{Wang2024MMLU-Pro:Benchmark}, MuSR~\citep{Sprague2024MuSR:Reasoning}, MATH~\citep{Hendrycks2021MeasuringDataset}, GPQA~\citep{Rein2023GPQA:Benchmark}, and IFEval~\citep{Zhou2023Instruction-FollowingModels}. These benchmarks comprise 21,606 questions across 53 tasks. The dataset spans models from under 1B to over 70B parameters, covering a wide range of model scales. 

We keep 4364 models after filtering by availability of responses across all benchmarks\footnote{We split models into source and target sets per experiment as described individually below.}. Additionally, we holdout IFEval (541 questions) as this benchmark will be later used for obtaining our target preferences, leaving 52 tasks across 5 benchmarks (21065 questions) available for alignment.

\xhdr{Target Preferences}
We simulate target preferences using reward models trained separately on two datasets: (1) Helpsteer~\citep{Wang2023HelpSteer:SteerLM}, which consists of 10,459 single-turn prompts with human preferences for helpfulness, coherence, and complexity and (2) UltraFeedback~\citep{Cui2024UltraFeedback:Feedback}, which contains over 1 million GPT-4 human preferences for 250k user-assistant conversations across helpfulness, honesty, and verbosity. We focus on helpfulness for Helpsteer and honesty for UltraFeedback, denoting their reward models as $\text{Armo}$,  $\text{GPT2}$ and $\text{DPA}$.\footnote{All reward models have been used in already published work and are obtained from HuggingFace. Further details are explained in Appendix~\ref{ap:rm_human_prefs}. We evaluate our results using 4 additional reward models in Appendix \ref{ap:pref_heterog}} These reward models score model responses on IFEval to construct a ``ground truth'' preference ranking over all models. 

\xhdr{Baselines}
We compare \textsc{BenchAlign} against three simple baselines and two benchmark distillation methods: (1) \textsc{No Alignment},\footnote{We use \textsc{No Alignment} to refer specifically to the absence of question reweighting, not to imply that other baselines produce preference-aligned rankings.} which aggregates all benchmarks without reweighting, (2) \textsc{Individual}, which select the task (across 52) with the highest Spearman rank correlation with target preferences as a benchmark (3) \textsc{Random}, which randomly selects one OpenLLMLeaderboard benchmark, (4) \textsc{MetaBench}~\citep{Kipnis2025Metabench:Models}, which constructs a sparse benchmark by removing questions with redundant information, and (5) \textsc{TinyBenchmarks}~\citep{Polo2024TinyBenchmarks:Examples}, which distills a single benchmark into a smaller set of questions with the most predictive power for model performances. Further implementation details for all methods are described in Appendix~\ref{ap:baselines_implementation}. 

\xhdr{Metrics}
We use two metrics: 1) the Spearman rank correlation ($\rho$) measure how closely the predicted rankings match the actual rankings, and 2) \textit{pairwise ranking accuracy} counts how frequently relative model pairs are predicted correctly, as shown in Equation \ref{eq:pairwise_acc}, where $\mathcal{P} = \mathcal{F}\times\mathcal{F}$ denotes the set of all model pairs,
$r_{ij}$ indicates the ground-truth relative ordering between models $f_i$ and $f_j$, and $\hat{r}_{ij}$ is the predicted ordering. The indicator function $\mathbb{I}[\cdot]$ outputs 1 if the predicted order is equal to the real order, and 0 otherwise\footnote{Margins of error calculated at the 95\% confidence unless indicated otherwise}.
\begin{equation}
\text{Acc}_\text{pair} =
\frac{1}{|\mathcal{P}|}
\sum_{(i,j)\in\mathcal{P},i\not=j}
\mathbb{I}\!\left[\hat{r}_{ij} = r_{ij}\right]
\label{eq:pairwise_acc}
\end{equation}

\xhdr{Implementation Details}
We implement \textsc{BenchAlign} using a linear RankNet model trained with a pairwise ranking loss. We use the Adam optimizer with a learning rate of 1e-4, a margin of separation of $0$, and no weight decay unless otherwise indicated. Since the target ranking is given by the reward model's scores evaluated on IFEval, we exclude IFEval from \textsc{BenchAlign}'s training data\footnote{We provide further analysis on our learning-to-rank algorithm and compute resources in Appendix \ref{ap:ltr_choices}}.

\subsection{Can aligned benchmarks generalize across model sizes? (RQ 1)}
\label{sec:perf_model_size_splits}

With recent shifts towards larger general-purpose models, we first examine whether aligned benchmarks can generalize to larger unseen models. We therefore test whether benchmark weightings learned from smaller models can accurately predict the preference rankings of much larger models with parameter sizes unobserved during training.
We therefore our alignment approach to the baselines on three different source-target splits based on different parameter thresholds (13B, 30B, and 70B). Models above a given threshold are held out for testing, while the remaining smaller models are used for training.\footnote{Appendix~\ref{ap:model_scale_experiments} provides further details on the models used for these experiments.} 

\xhdr{Aligned benchmarks trained on smaller models generalize to larger, unseen models} 
As shown in Table~\ref{tab:model_size_splits_helpsteer_ultrafeedback}, our proposed method consistently achieves a higher rank correlation ($\rho$) and pairwise ranking accuracy ($Acc_{pair}$) than all baselines.
\textsc{MetaBench} and \textsc{TinyBenchmarks} show much lower and inconsistent correlations than the proposed method, which may be attributed to how both methods are designed to predict benchmark performance rather than human preferences. In fact, we find that they exhibit a similar level of correlations to the \textsc{No Alignment} baseline and sometimes even worse than the \textsc{Random} baseline. \textsc{Individual} is the second best baseline in almost all our settings, highlighting the importance of choosing a benchmark related to the target as a proxy of downstream performance. However, an important gap remains between benchmarks and preferences that only benchmark alignment can mitigate effectively.


\begin{table*}[t]
\centering
\small
\setlength{\tabcolsep}{5pt}
\renewcommand{\arraystretch}{1.05}

\resizebox{\textwidth}{!}{
\begin{tabular}{c c c c c c  c c c c}
\toprule
 \multirow{4}{*}{\makecell{\textbf{Target}\\\textbf{Models}}} & \multirow{4}{*}{\textbf{Method}} & \multicolumn{4}{c }{\textbf{Helpfulness}} & \multicolumn{4}{c}{\textbf{Honesty}} \\
\cmidrule(lr){3-6}  \cmidrule(lr){7-10} 

& & \multicolumn{2}{c}{$Acc_{pair}$}
& \multicolumn{2}{c}{$\rho$}
& \multicolumn{2}{ c}{$Acc_{pair}$}
& \multicolumn{2}{c}{$\rho$} \\
\cmidrule(lr){3-4}  \cmidrule(lr){5-6} \cmidrule(lr){7-8}  \cmidrule(lr){9-10} 

& & $\text{ArmoRM}$ & $\text{GPT2}$ & $\text{ArmoRM}$ & $\text{GPT2}$ & $\text{ArmoRM}$ & $\text{DPA}$ & $\text{ArmoRM}$ & $\text{DPA}$ \\ \midrule

\multirow{6}{*}{70B+}& \textsc{No Alignment}& $0.542\pm0.010$& $0.480\pm0.010$& $0.154\pm0.165$& $0.044\pm0.251$& $0.559\pm0.010$& $0.567\pm0.010$& $0.199\pm0.164$&$0.188\pm0.164$ \\
& \textsc{Random} & $0.538 \pm 0.010$& $0.468 \pm 0.010$& $0.087 \pm 0.161$& $-0.106 \pm 0.166$& $0.621 \pm 0.010$& $0.605 \pm 0.010$& $0.280 \pm 0.145$&$0.237 \pm 0.150$ \\
& \textsc{Individual}& 
$0.675\pm0.009$ & $\mathbf{0.621\pm0.010}$ & $0.492\pm0.136$ & $0.306\pm0.157$ & $0.694\pm0.009$& $0.679\pm0.012$ & $0.536\pm0.129$ & $0.490\pm0.136$\\
& \textsc{MetaBench} & $0.564\pm0.007$ & $0.484\pm0.007$& $0.200\pm0.158$& $-0.028\pm0.165$ &  $0.581\pm 0.007$ &  $0.575\pm 0.007$ &  $0.246\pm0.155$ &  $0.216\pm 0.157$ \\
& \textsc{TinyBenchmarks}& $0.540 \pm 0.010$ & $0.488 \pm 0.010$ & $0.147 \pm 0.161$ & $-0.024 \pm 0.165$ & $0.554 \pm 0.010$ & $0.563 \pm 0.010$ & $0.187 \pm 0.159$ & $0.187 \pm 0.159$ \\
& \textsc{BenchAlign} & $\mathbf{0.778 \pm 0.008}$& $0.620 \pm 0.010$& $\mathbf{0.707 \pm 0.074}$& $\mathbf{0.333 \pm 0.139}$& $\mathbf{0.774 \pm 0.008}$& $\mathbf{0.746 \pm 0.009}$& $\mathbf{0.706 \pm 0.074}$& $\mathbf{0.659 \pm 0.084}$\\
\midrule

\multirow{6}{*}{30B+}& \textsc{No Alignment}& $0.605\pm0.005$& $0.477\pm0.005$& $0.331\pm0.114$& $0.052\pm0.226$& $0.618\pm0.005$& $0.609\pm0.005$& $0.368\pm0.111$&$0.319\pm0.115$ \\
& \textsc{Random} & $0.551 \pm 0.005$& $0.452 \pm 0.005$& $0.108 \pm 0.120$& $-0.148 \pm 0.122$&  $0.650 \pm 0.005$&  $0.618 \pm 0.005$&  $0.407 \pm 0.098$&  $0.317 \pm 0.106$ \\
& \textsc{Individual}& $0.678\pm0.005$& $0.606\pm0.005$& $0.501\pm0.098$& $0.273\pm0.118$& $0.692\pm0.005$& $0.675\pm0.005$& $0.537\pm0.094$&$0.480\pm0.100$\\
& \textsc{MetaBench}  & $0.606\pm0.004$ & $0.497\pm0.004$ & $0.328\pm 0.110$& $0.007\pm0.123$& $0.613\pm 0.004$ & $0.604\pm 0.004$ & $0.346\pm 0.108$ & $0.302\pm0.112$  \\
& \textsc{TinyBenchmarks}& $0.605 \pm 0.005$ & $0.492 \pm 0.005$& $0.323 \pm 0.110$ & $-0.016 \pm 0.123$ & $0.615 \pm 0.005$ & $0.602 \pm 0.005$ & $0.354 \pm 0.108$ & $0.299 \pm 0.112$ \\
& \textsc{BenchAlign} & $\mathbf{0.765 \pm 0.005}$& $\mathbf{0.637 \pm 0.005}$& $\mathbf{0.710 \pm 0.056}$& $\mathbf{0.387 \pm 0.100}$&  $\mathbf{0.773 \pm 0.005}$&  $\mathbf{0.763 \pm 0.005}$&  $\mathbf{0.730 \pm 0.053}$&  $\mathbf{0.713 \pm 0.055}$\\
\midrule

\multirow{6}{*}{13B+}& \textsc{No Alignment}& $0.583\pm0.002$& $0.531\pm0.002$& $0.245\pm0.065$& $0.094\pm0.067$& $0.588\pm0.002$& $0.577\pm0.002$& $0.251\pm0.064$&$0.231\pm0.065$ \\ 
 & \textsc{Random} & $0.558 \pm 0.002$& $0.537 \pm 0.002$& $0.125 \pm 0.066$& $0.083 \pm 0.067$&  $0.603 \pm 0.002$&  $0.600 \pm 0.002$&  $0.280 \pm 0.061$&  $0.272 \pm 0.061$ \\ 
 & \textsc{Individual}& $0.660\pm0.002$& $0.608\pm0.002$& $0.430\pm0.057$& $0.278\pm0.063$& $0.661\pm0.002$& $0.669\pm0.002$& $0.432\pm0.057$&$0.451\pm0.055$\\
 
& \textsc{MetaBench} & $0.584\pm 0.001$ & $0.530\pm0.001$& $0.243\pm0.063$ & $0.089\pm0.067$& $0.586\pm 0.001$ & $0.579\pm 0.001$ & $0.248\pm 0.063$ & $0.229\pm 0.064$ \\
& \textsc{TinyBenchmarks}& $0.589 \pm 0.002$ & $0.543 \pm 0.002$ & $0.259 \pm 0.063$ & $0.128 \pm 0.066$ & $0.592 \pm 0.002$ & $0.583 \pm 0.002$ & $0.267 \pm 0.063$ & $0.243 \pm 0.063$ \\
& \textsc{BenchAlign} & $\mathbf{0.741 \pm 0.001}$& $\mathbf{0.696 \pm 0.002}$& $\mathbf{0.674 \pm 0.035}$& $\mathbf{0.552 \pm 0.045}$&  $\mathbf{0.748 \pm 0.001}$ & $\mathbf{0.749 \pm 0.001}$&  $\mathbf{0.686 \pm 0.034}$&  $\mathbf{0.691 \pm 0.034}$\\
\bottomrule
\end{tabular}
}
\caption{Helpsteer and UltraFeedback results under model scale splits.}
\label{tab:model_size_splits_helpsteer_ultrafeedback}
\end{table*}


The results in $\text{GPT2}$ further define the current frontier of benchmark alignment. The correlations achieved by \textsc{No Alignment}, \textsc{MetaBench}, and \textsc{TinyBenchmarks} are close to zero, indicating that overall accuracy across all benchmarks provides no meaningful information about the target preference. \textsc{Individual} and \textsc{BenchAlign} still exhibit a positive correlation, but this value is approximately $0.3$ and $0.4$ points lower than in other preference settings, respectively. This finding suggests that benchmark alignment is harder to achieve when benchmark data and target preferences are poorly correlated.

To summarize, our results show that with sufficient preference signals, aligned benchmarks can generalize across model scales, even when the source data contains only small to medium-sized models. We perform a series of ablations to show the robustness of our results in Appendix~\ref{ap:ablation}.

\subsection{How many training models are needed for benchmark alignment? (RQ 2)}
\label{sec:perf_num_models}

So far, we have applied benchmark alignment using all training models. In practice, collecting large-scale model evaluations and human preference rankings can be computationally and financially prohibitive. Moreover, the choice of the learning-to-rank algorithm may directly impact the number of models required for benchmark alignment, as different methods vary in how efficiently they extract signal from limited pairwise preference data. Therefore, we study how 
feasible is benchmark alignment under limited data settings. We implement two sets of experiments that use five learning-to-rank algorithms for alignment\footnote{The learning-to-rank algorithms are described further in Appendix~\ref{ap:other_ltr}.} on the setup from one of our best results of Section \ref{sec:perf_model_size_splits}: $\text{ArmoRM}$ (Helpful) as the target preference and $30$B+ target models.

\xhdr{Naive benchmark alignment requires a large number of smaller-sized models} We study a naive but realistic scenario in which practitioners incorporate model evaluations as they become available, prioritizing larger, more recent models, without prior knowledge of which model responses are most informative for benchmark alignment. To simulate this setting, we sort the source models by parameter count, train on the smallest 100 models, and progressively expand the training set in increments of 100 as larger models become available.

\begin{figure*}[t]
    \centering
    \begin{minipage}[t]{0.48\textwidth}
        \centering
        \includegraphics[width=\linewidth]{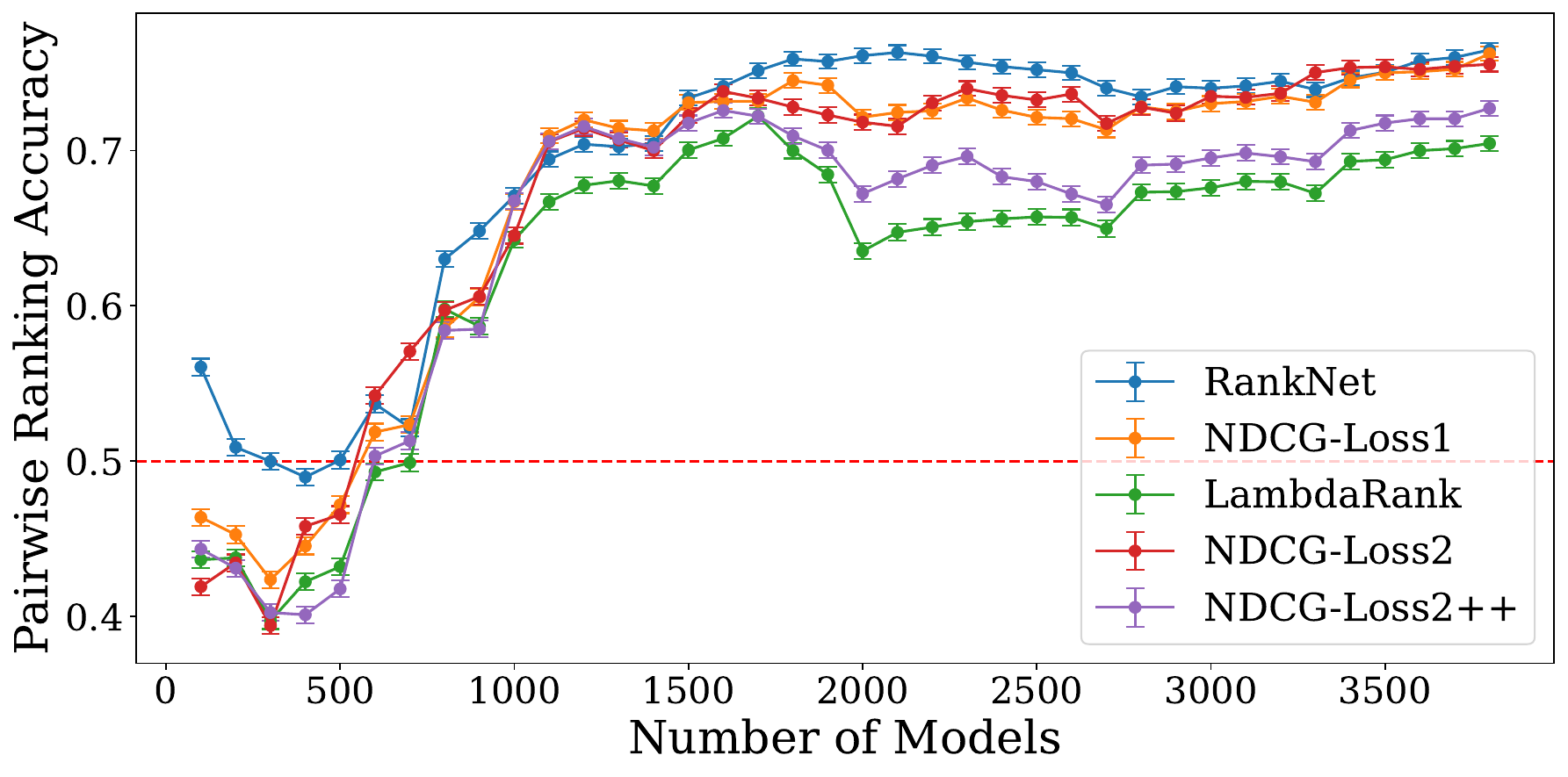}
        \vspace{2pt}
        {\small (a) Pairwise ranking accuracy ($Acc_{pair}$)}
    \end{minipage}\hfill
    \begin{minipage}[t]{0.48\textwidth}
        \centering
        \includegraphics[width=\linewidth]{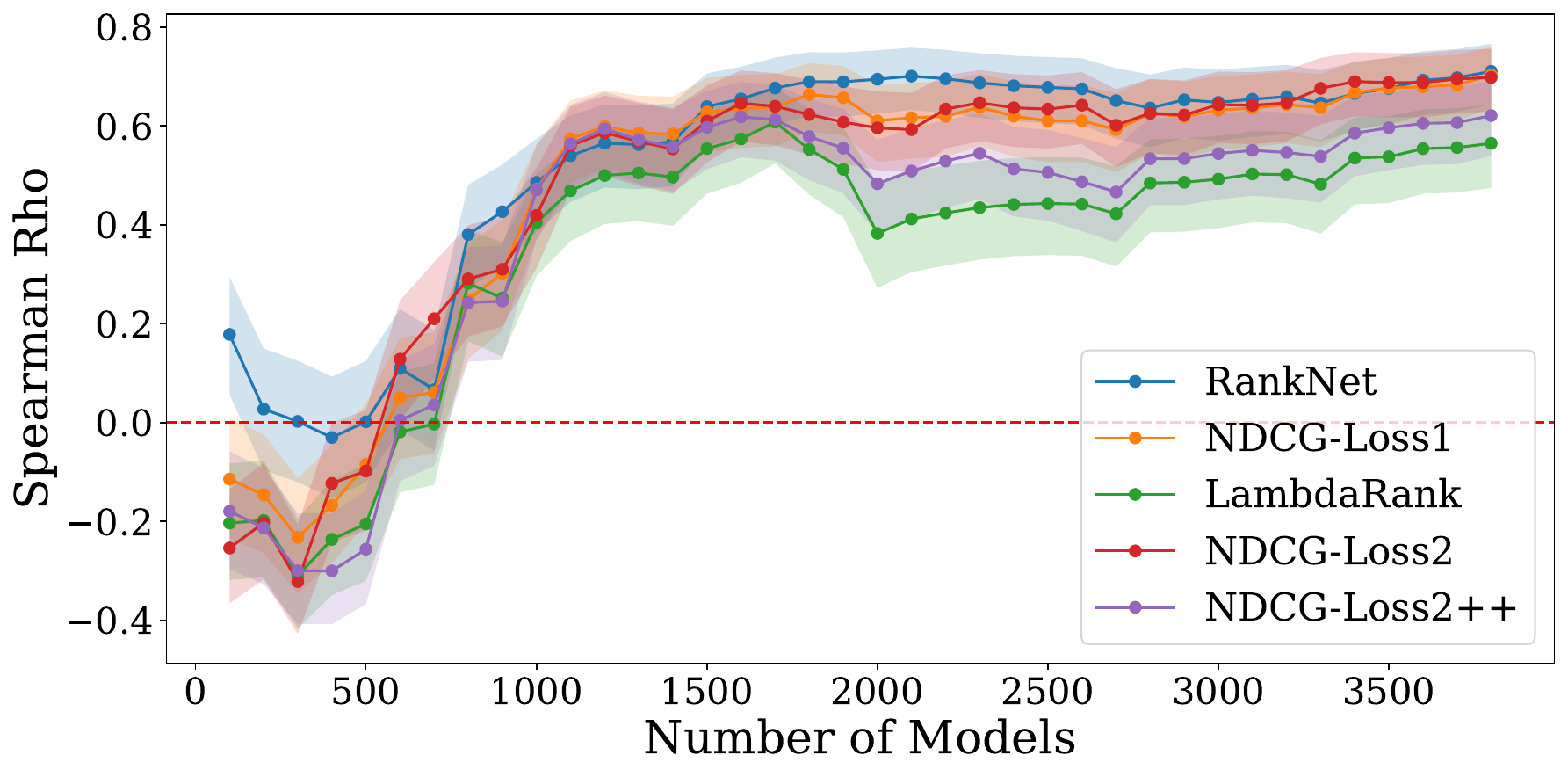}
        \vspace{2pt}
        {\small (b) Spearman rank coefficient ($\rho$)}
    \end{minipage}

    \caption{Benchmark alignment under different number of models and learning-to-rank algorithms. Error bars obtained using the number of pairs and sample size from the target set for $Acc_{pair}$ and $\rho$, respectively. The dotted line represents the results when models are ranked randomly.}
    \label{fig:bench_align_num_models}
\end{figure*}

Figure~\ref{fig:bench_align_num_models} shows that training with smaller models outputs benchmarks that rank larger models no differently than tossing a coin. This is expected as a considerably low number of parameters (less than $1$B parameters) might be insufficient to produce meaningful results on the OpenLLMLeaderboard benchmarks. While including larger models leads to an improvement in the $Acc_{pair}$ and $\rho$, we find that expanding the model pool with larger-sized models for training does not necessarily lead to more aligned benchmarks. Most implementations show some improvement beyond the first 25\% of all OpenLLMLeaderboard models which range between 1 to 7B parameters. However, the gains in rank correlations and pairwise ranking accuracy are marginal after further increments, indicating that smaller models already provide sufficient signals for learning weights that generalize across model scale.

Our results suggest that benchmark alignment needs a large number of small-sized models (less than 7B parameters) to infer the preference ranking of larger-sized models, and that the inclusion of larger models in training does not improve generalization significantly.

\xhdr{Informative subsets enable alignment with as few as $\sim$20 models} To study whether benchmark alignment can be simplified, we modify our previous setup to identify informative subsets of models ahead of time. Since our previous experiments showed that alignment signals vary across model scales, we first sort the source models by parameter count. We then select a fixed number of neighboring models from the ordered list, align our benchmarks using only those models, and repeat the process by shifting the selected subset across the list. This allows us to identify which subset provides the strongest alignment signals at each subset size. 

\begin{figure*}[t]
    \centering
    \begin{subfigure}[b]{0.49\linewidth}
        \centering
        \includegraphics[width=\linewidth]{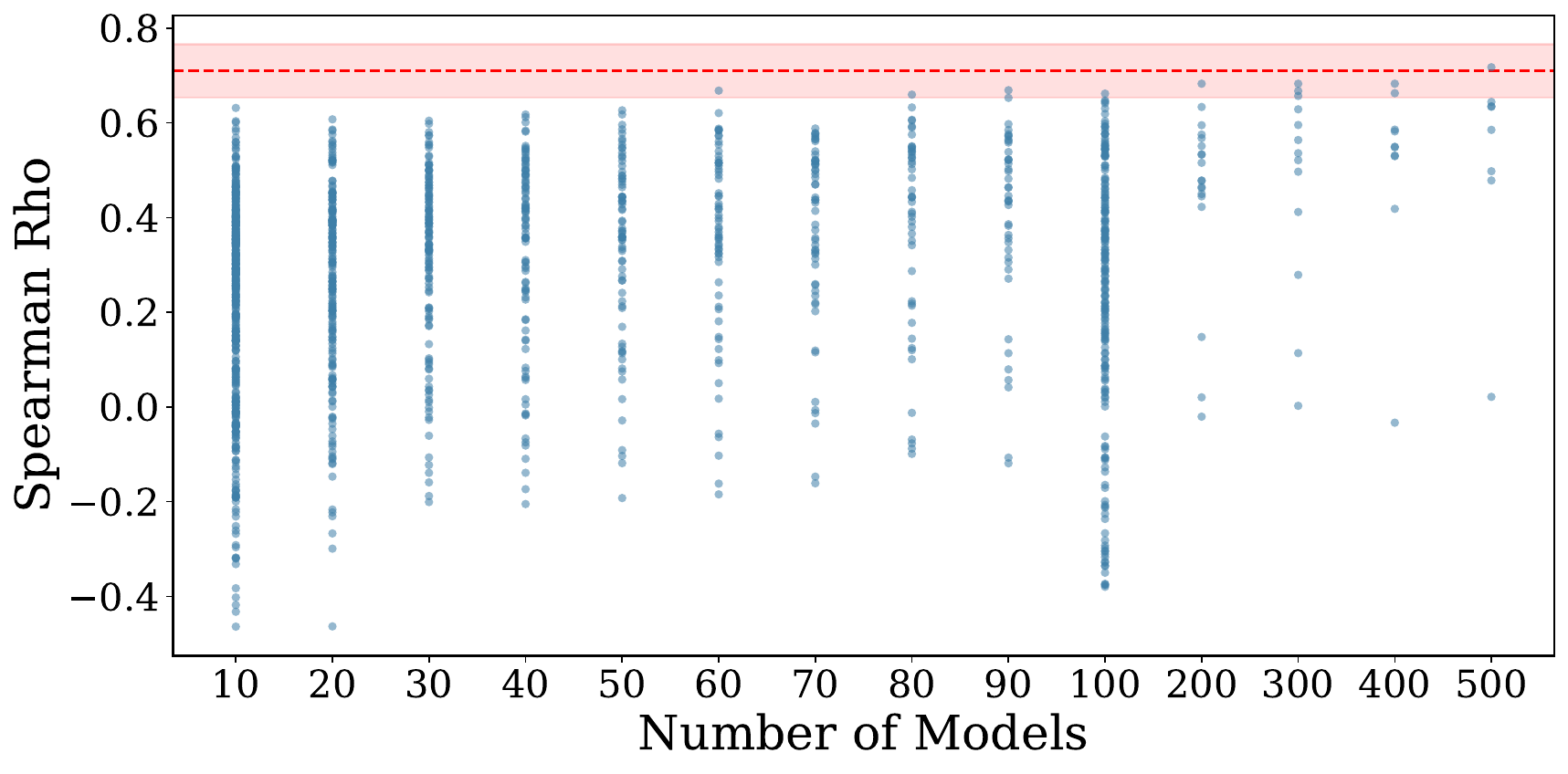}
        \caption{All model subsets}
        \label{fig:vary_window_size}
    \end{subfigure}
    \hfill
    \begin{subfigure}[b]{0.49\linewidth}
        \centering
        \includegraphics[width=\linewidth]{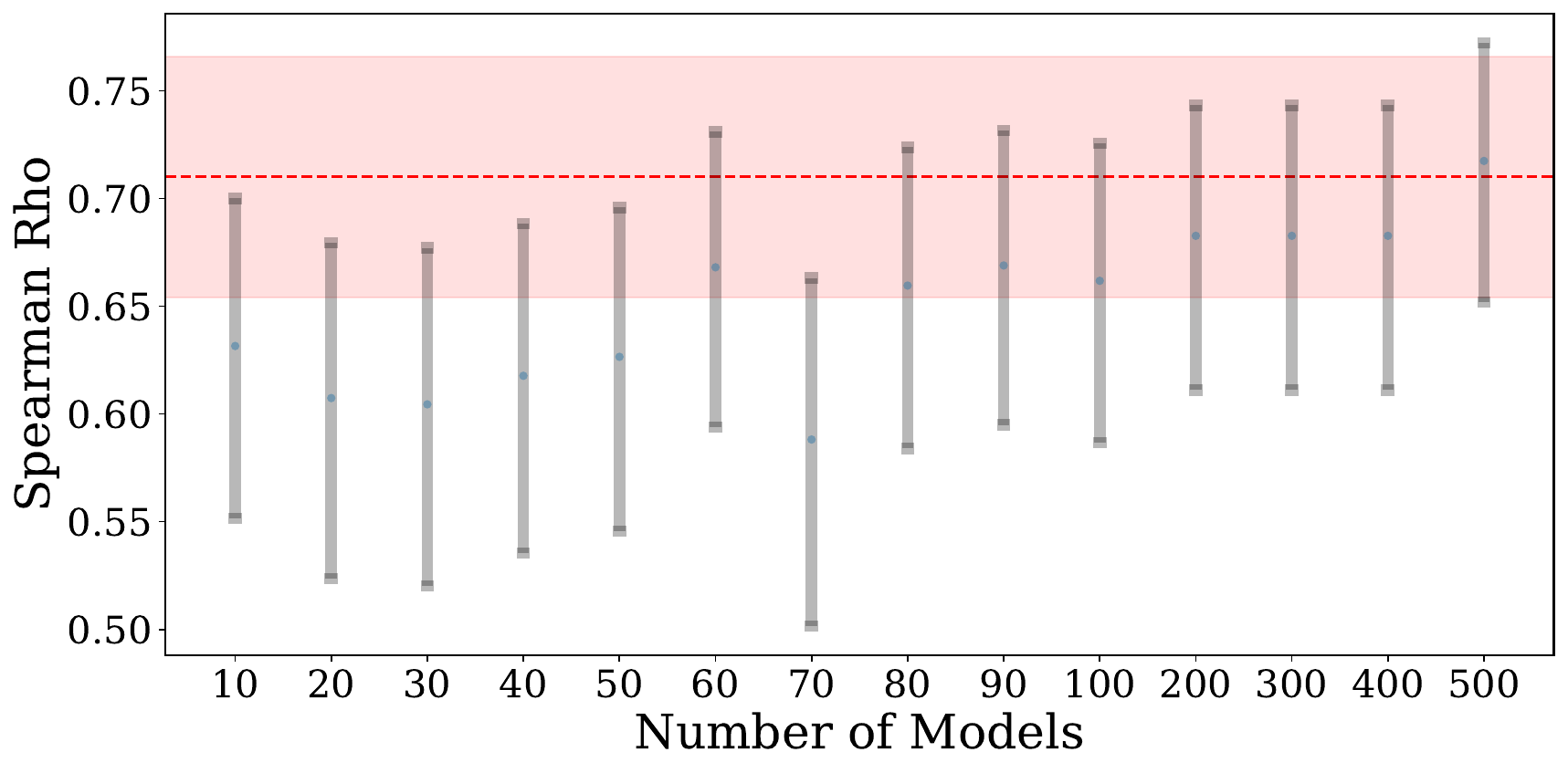}
        \caption{Best-performing model subset}
        \label{fig:best_per_window}
    \end{subfigure}
    \caption{Aligned benchmarks created with different numbers of models. In (a), each point shows the Spearman Rho for one subset of models. In (b), each point reports the results of the best subset of models by correlation. Error bars obtained using the sample size from the target set. The dotted line represents the results of alignment using all models.}
    \label{fig:bench_align_opt_num_models}
\end{figure*}

Figure \ref{fig:bench_align_opt_num_models} shows the results for $\text{ArmoRM}$ (Helpful), with a target of 30B+ models.\footnote{The results for the remaining reward models and target models are in Appendix \ref{ap:opt_num_models}.} We observe a direct relationship between the correlation and the number of models, consistent with our findings from Figure \ref{fig:bench_align_num_models}, as larger subsets allow for more variety in the model set for performing benchmark alignment. However, a surprising finding is that groups of fewer than 100 models still achieve results within the error margins of using the full training split. Figure~\ref{fig:vary_window_size} shows that most of the benchmarks created with at least 50 models have a correlation skewed upwards. Furthermore, a closer inspection of the best subset of models reveals that we can create aligned benchmarks with far less models. Figure~\ref{fig:best_per_window} indicates that this requirement can be minimized to 10 models, while showing that the best benchmarks created with similar sized groups are not different from using all models. 

Our results suggest that knowing the most informative subset of models could reduce the models needed for alignment to $\sim0.25\%$ of all models in OpenLLMLeaderboard. To further prove their robustness, we run the same experiment for all reward models and target models from Section \ref{sec:perf_model_size_splits}, finding that we can create aligned benchmarks with windows equal or smaller than 20 models for almost all settings (see Appendix \ref{ap:opt_num_models}). We also study how to select the best subset of models using the model family and generation in Appendix \ref{ap:best_subset_models}. However, this is infeasible on these data, suggesting a line of future work.

\subsection{How many questions are needed for benchmark alignment? (RQ 3)}
\label{sec:perf_num_questions}

Our results from Section \ref{sec:perf_model_size_splits} suggest that a large number of questions may be required for alignment to be feasible in this exact setting. This would often be prohibitively expensive as it requires significant computing and evaluation resources. However, existing benchmarks may be highly redundant, leading to an artificial overestimate of this number~\citep{Kipnis2025Metabench:Models, Polo2024TinyBenchmarks:Examples}. We therefore test whether these data requirements can be further minimized on two sets of experiments using the same experimental setup as Section \ref{sec:perf_num_models}.

\xhdr{Naive benchmark alignment requires a large fraction of all benchmark questions} Similar to Section \ref{sec:perf_num_models}, we design a naive scenario that corresponds to a practitioner without information on the best subset of questions. We first randomly sort the questions and train our alignment implementations with the benchmark responses of all training models on a random subset of 100 questions. Then, we increase the number of questions in steps of 100 until all questions are included.

\begin{figure*}[t]
    \centering
    \begin{minipage}[t]{0.48\textwidth}
        \centering
        \includegraphics[width=\linewidth]{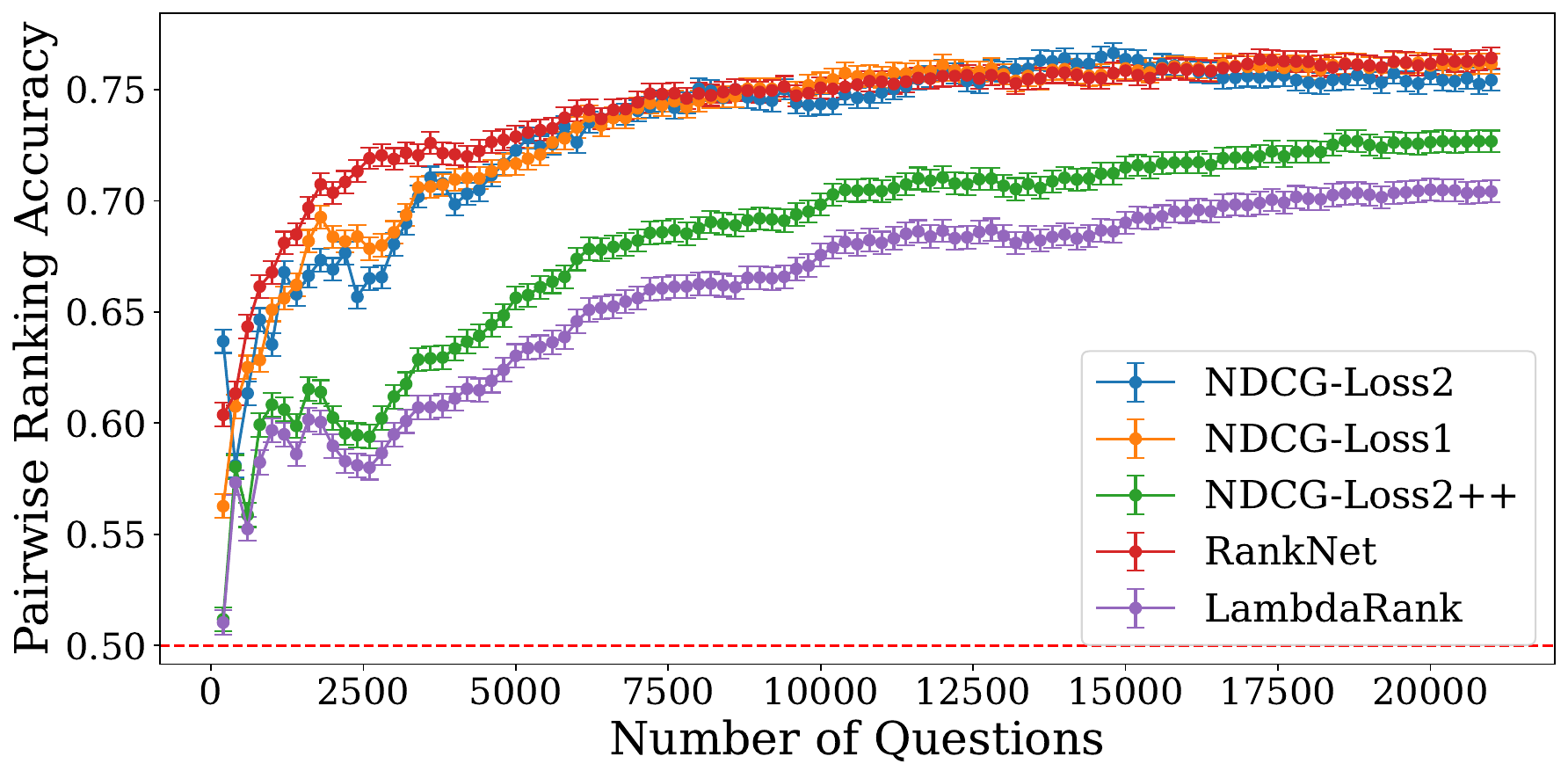}
        \vspace{2pt}
        {\small (a) Pairwise ranking accuracy ($\text{Acc}_\text{pair}$)}
    \end{minipage}\hfill
    \begin{minipage}[t]{0.48\textwidth}
        \centering
        \includegraphics[width=\linewidth]{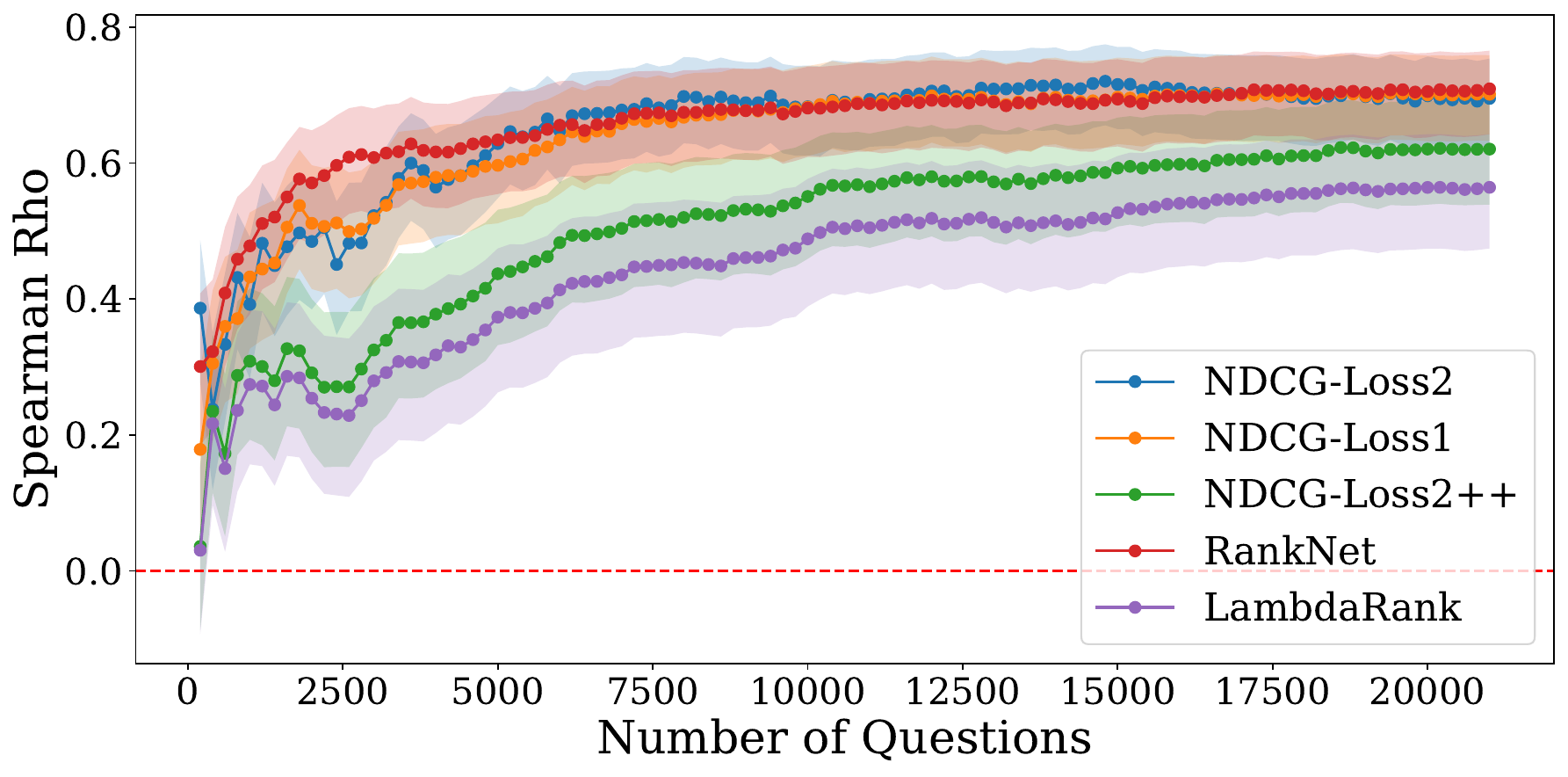}
        \vspace{2pt}
        {\small (b) Spearman rank coefficient ($\rho$)}
    \end{minipage}

    \caption{Benchmark alignment under different numbers of questions and learning-to-rank algorithms. Error bars obtained using the number of pairs and sample size from the target set for $Acc_{pair}$ and $\rho$, respectively. The dotted line represents the results when models are ranked randomly.}
    \label{fig:bench_align_num_questions}
\end{figure*}

Figure~\ref{fig:bench_align_num_questions} shows that we can detect signals about model preferences with a limited number of questions, even when using 100 questions across most implementations. Moreover, increasing the number of available questions results in large increases, as shown by the upward trend in both $Acc_{pair}$ and $\rho$ over the first 2500 questions. This result is expected, as augmenting the number of questions increases the probability of the training data capturing a wider variety of questions across different preference levels. This allows the alignment implementations to learn which questions discriminate better according to the target preference. However, the trend reverses when using around 2500 questions for training, except for \textsc{RankNet}. Among all learning-to-rank algorithms, \textsc{RankNet} shows more consistent results on both $Acc_{pair}$ and $\rho$ across varying numbers of questions included in training.

Once we increase the number of questions to more than 5000, all benchmark alignment implementations begin to show more steady improvements on both metrics. However, three out of the five implementations, including \textsc{RankNet}, \textsc{NDCG-Loss1}, and \textsc{NDCG-Loss2}, show plateaus in $Acc_{pair}$ and $\rho$. On the other hand, \textsc{LambdaRank} and \textsc{NDCG-Loss2++} show lower performance on both metrics, but then steadily improve and ultimately converge to performance comparable to the other implementations. These results indicate that the choice of the learning-to-rank algorithm may affect how the alignment of benchmarks scales with the number of available questions.

Overall, our results show that benchmark alignment needs around 5000 questions ($\sim 25\%$ of all OpenLLMLeaderboard questions) to achieve its best results.

\xhdr{Benchmark distillation methods can shrink question set requirements} We create a setting where a practitioner knows in advance which are the most predictive questions among all benchmarks by using distillation methods. Specifically, we use \textsc{Metabench} and \textsc{TinyBenchmarks} on all benchmarks jointly to find the set of most predictive questions of benchmark performance and then apply alignment using the responses of all models from the training set on these questions. Table~\ref{tab:distill_benchalign} shows the results for $\text{ArmoRM}$ (Helpful) and $\text{ArmoRM}$ (Honest), with a target of 13B+ models\footnote{We choose RankNet as it is the best learning-to-rank algorithm found in Section~\ref{sec:perf_num_models}.}

\Needspace{12\baselineskip}
\begin{wraptable}[10]{r}{0.50\textwidth}
\centering
\setlength{\tabcolsep}{2pt}
\renewcommand{\arraystretch}{0.92}
\resizebox{\linewidth}{!}{%
\begin{tabular}{cccc}
\toprule
RM & Method & $Acc_{pair}$ & $\rho$ \\
\midrule
\multirow{4}{*}{$\text{ArmoRM}$}
& \textsc{No Alignment} & $0.583\pm0.002$ & $0.245\pm0.065$ \\
& \textsc{BenchAlign} & $0.741 \pm 0.001$ & $0.674 \pm 0.035$ \\
& \textsc{MetaBench + BenchAlign} & $0.744\pm0.001$ & $0.677\pm0.035$ \\
& \textsc{TinyBenchmarks + BenchAlign} & $\mathbf{0.754\pm0.001}$ & $\mathbf{0.700\pm0.033}$ \\
\midrule
\multirow{4}{*}{$\text{ArmoRM}$}
& \textsc{No Alignment} & $0.588\pm0.002$ & $0.251\pm0.064$ \\
& \textsc{BenchAlign} & $0.748 \pm 0.001$ & $0.686 \pm 0.034$ \\
& \textsc{MetaBench + BenchAlign} & $0.757\pm0.001$ & $0.706\pm0.032$ \\
& \textsc{TinyBenchmarks + BenchAlign} & $\mathbf{0.770\pm0.001}$ & $\mathbf{0.735\pm0.030}$ \\
\bottomrule
\end{tabular}%
}
\caption{Distillation methods plus benchmark alignment.}
\label{tab:distill_benchalign}
\end{wraptable}

In all cases, the combination of a benchmark distillation method with benchmark alignment matches or outperforms alignment alone, despite using a reduced set of test items. Particularly, \textsc{TinyBenchmarks} achieves the best results. This may be explained by the difference in the number of questions used by the two benchmarks: \textsc{TinyBenchmarks} finds 5000 of the most predictive questions, whereas \textsc{Metabench} uses only 1500. These results not only show that the evaluation costs of aligned benchmarks can be reduced, but also how we may achieve that. We can create aligned benchmarks with as few as 1500 questions, $\sim$7\% of all OpenLLMLeaderboard questions, by relying on the questions most predictive of benchmark performance.

We further inspect our findings by replicating this experiment on all reward models and target models from Section \ref{sec:perf_model_size_splits} in Appendix~\ref{ap:opt_num_questions}, finding that applying alignment on the subset obtained by a benchmark distillation method produces similar results to applying alignment on all test items.

\section{Limitations}\label{sec:limitations}
There are three key limitations. The first regards reward models as proxies of real world utility. Given the limited availability of large-scale alternatives for capturing human preferences, we rely on reward models trained directly from human feedback~\citep{Ouyang2022TrainingFeedback, Christiano2017DeepPreferences}. However, they may not capture the full heterogeneity of human preferences. We show robustness to such cases on alternative reward models and datasets (Appendix \ref{ap:pref_heterog}), but incorporating improved proxies of preferences remains an important direction for future work. 

The second concerns reliance on ground-truth rankings: alignment requires reliable targets, and if these already exist, the need for a new benchmark is unclear. We acknowledge this limitation, however, this is not our goal. Our approach is designed as a simple tool to study the lack of external validity from existing benchmarks and as a step towards developing novel benchmark alignment methods. Moreover, our results in Sections \ref{sec:perf_num_models} and \ref{sec:perf_num_questions} are particularly useful for future works as they show the minimal data requirements for achieving alignment. 

Finally, our results show that we need benchmarks that already have a positive correlation with the target preferences for benchmark alignment. We consider this an important finding as it highlights that practitioners should not aim for aligning arbitrary benchmarks with their targets. Rather, they should aim to align existing benchmarks that are not capturing the utility they were designed to measure.\footnote{We discuss our \textbf{broader impacts} in Appendix \ref{ap:broader_impacts}.} 

\section{Conclusions}\label{sec:conclusion}
In this paper, we introduce the problem of \textit{benchmark alignment} to address the gap between benchmark performance and its utility for predicting models' downstream preferences. We propose a simple tool and evaluation setup that identifies the limitations for aligning existing benchmarks and show that we can produce new benchmarks that generalize to unseen larger models.

Our work is not intended to replace the development of new benchmarks for tasks where existing evaluations fail to reflect real-world performance. Rather, it aims to study the external validity gap in current evaluations and to establish benchmark alignment as a meaningful direction for future research, while also informing practitioners on its requirements for downstream deployments.

\begin{ack}\label{sec:acknowledgments}
This work was sponsored in part by NIH Award
\#1OT2OD038079-01. We thank the University of Virginia's Research Computing team for maintaining excellent high-performance computing resources that allowed us to conduct this research.
\end{ack}

\bibliographystyle{apalike}
\bibliography{bibliography, references}

\newpage
\appendix
\section{Broader Impacts}\label{ap:broader_impacts}
This paper analyses whether introduces the problem of benchmark alignment using limited benchmark data and rankings derived from pairwise preferences. Our goal is to advance the field of machine learning by creating new benchmarks from existing ones that better reflect a set of target preferences. 

We consider that model selection for downstream deployment is an application of our work with potential societal consequences. Aligned benchmarks might be used on high-stakes settings such as healthcare, where rankings of language models in real-world clinical tasks are scarce. This also poses the risk of misuse as using the wrong subset of models can negatively impact on a patients health.

In addition, benchmark alignment could be used as a tool for benchmark diagnosis as our work offers a framework to determine whether existing benchmarks are aligned with real-world utility based on the pairwise ranking accuracy and the spearman rank correlation.

Finally, aligned benchmarks could be used to construct leaderboards of models that have yet to be tested for our target ranking. Our results suggest we can build generalizable benchmarks for larger models relying solely on small and medium sized model data.

\section{Reward Models} \label{ap:rm_human_prefs}

The main preferences we consider are Helfulness and Honesty as they are the standard in the model alignment literature to evaluate whether generations are consistent with human preferences.

We use three reward models from HuggingFace trained on Helpsteer-Helpfulness~\citep{Wang2023HelpSteer:SteerLM} and UltraFeedback-Honesty~\citep{Cui2024UltraFeedback:Feedback} as described in the following list:

\begin{itemize}
    \item $\text{ArmoRM}$ (Helpful): ArmoRM, Helpsteer-Helpfulness dimension~\citep{Wang2024InterpretableMixture-of-Experts}
    \item $\text{GPT2}$: GPT2-Large-helpful-reward-model~\citep{Yang2024Rewards-in-Context:Adjustment}
    \item $\text{ArmoRM}$ (Honest): ArmoRM, UltraFeedback-Honesty dimension~\citep{Wang2024InterpretableMixture-of-Experts}
    \item $\text{DPA}$: RewardModel-Mistral-7b-for-DPA-v1~\citep{Wang2024ArithmeticRewards}
\end{itemize}

ArmoRM~\citep{Wang2024InterpretableMixture-of-Experts} is a reward model trained to output a vector of rewards, with each dimension corresponding to a category of human values datasets such as HelpSteer and UltraFeedback. We used the Helpsteer-Helpfulness and UltraFeedback-Honesty as shown above.

Six models used to train HelpSteer and UltraFeedback were also present in our OpenLLMLeaderboard dataset and were therefore excluded from our experiments. These models are: ``wizardlmteam/wizardlm-13b-v1.2," ``meta-llama/llama-2-7b-hf," ``meta-llama/llama-2-13b-hf," ``meta-llama/llama-2-70b-chat-hf," ``tiiuae/falcon-40b-instruct," and ``eleutherai/pythia-12b."

To proof the robustness of our results, we also consider the Coherence, Complexity and Verbosity preferences from ArmoRM in Tables \ref{tab:model_size_splits_other_dimensions} and \ref{tab:random_splits_other_dimensions}.

\begin{table*}[t]
\centering
\small
\setlength{\tabcolsep}{5pt}
\renewcommand{\arraystretch}{1.05}

\resizebox{\textwidth}{!}{
\begin{tabular}{c c c c c c c c }
\toprule
 \multirow{3}{*}{\makecell{\textbf{Target}\\\textbf{Models}}} & \multirow{3}{*}{\textbf{Method}} & \multicolumn{2}{c}{\textbf{Coherence}} & \multicolumn{2}{c}{\textbf{Complexity}} & \multicolumn{2}{c}{\textbf{Verbosity}} \\
\cmidrule(lr){3-4} \cmidrule(lr){5-6} \cmidrule(lr){7-8}

& & $Acc_{pair}$ & $\rho$ & $Acc_{pair}$ & $\rho$ & $Acc_{pair}$ & $\rho$ \\
\midrule

\multirow{4}{*}{70B+} & \textsc{No Alignment} & $0.519\pm0.010$ & $0.080\pm0.162$ & $0.569\pm0.010$ & $0.205\pm0.153$ & $0.536\pm0.010$ & $0.104\pm0.160$ \\
 & \textsc{Random} & $0.576\pm0.010$ & $0.210\pm0.152$ & $0.585\pm0.010$ & $0.223\pm0.151$ & $0.552\pm0.010$ & $0.103\pm0.160$ \\
 & \textsc{Individual} & $0.653\pm0.009$ & $0.426\pm0.126$ & $0.674\pm0.009$ & $0.448\pm0.123$ & $\mathbf{0.638\pm0.009}$ & $\mathbf{0.355\pm0.136}$ \\
 & \textsc{Metabench} & $0.542\pm0.007$ & $0.127\pm0.162$ & $0.587\pm0.007$ & $0.251\pm0.155$ & $0.555\pm0.007$ & $0.149\pm0.161$ \\
 & \textsc{TinyBenchmarks} & $0.517\pm0.007$ & $0.076\pm0.164$ & $0.562\pm0.007$ & $0.180\pm0.160$ & $0.527\pm0.007$ & $0.080\pm0.164$ \\
 & \textsc{BenchAlign} & $\mathbf{0.765\pm0.008}$ & $\mathbf{0.681\pm0.079}$ & $\mathbf{0.701\pm0.009}$ & $\mathbf{0.540\pm0.107}$ & $0.482\pm0.010$ & $-0.037\pm0.166$ \\
\midrule
\multirow{4}{*}{30B+} & \textsc{No Alignment} & $0.586\pm0.005$ & $0.272\pm0.110$ & $0.609\pm0.005$ & $0.319\pm0.106$ & $0.541\pm0.005$ & $0.119\pm0.119$ \\
 & \textsc{Random} & $0.611\pm0.005$ & $0.320\pm0.106$ & $0.608\pm0.005$ & $0.295\pm0.108$ & $0.547\pm0.005$ & $0.113\pm0.120$ \\
 & \textsc{Individual} & $0.670\pm0.005$ & $0.462\pm0.091$ & $0.666\pm0.005$ & $0.445\pm0.093$ & $\mathbf{0.594\pm0.005}$ & $\mathbf{0.263\pm0.111}$ \\
 & \textsc{Metabench} & $0.592\pm0.004$ & $0.282\pm0.113$ & $0.590\pm0.004$ & $0.269\pm0.114$ & $0.556\pm0.004$ & $0.158\pm0.120$ \\
 & \textsc{TinyBenchmarks} & $0.588\pm0.004$ & $0.269\pm0.114$ & $0.605\pm0.004$ & $0.305\pm0.112$ & $0.542\pm0.004$ & $0.125\pm0.121$ \\
 & \textsc{BenchAlign} & $\mathbf{0.763\pm0.005}$ & $\mathbf{0.707\pm0.057}$ & $\mathbf{0.718\pm0.005}$ & $\mathbf{0.600\pm0.073}$ & $0.573\pm0.005$ & $0.222\pm0.114$ \\
\midrule
\multirow{4}{*}{13B+} & \textsc{No Alignment} & $0.574\pm0.002$ & $0.215\pm0.063$ & $0.582\pm0.002$ & $0.239\pm0.063$ & $0.570\pm0.002$ & $0.207\pm0.064$ \\
 & \textsc{Random} & $0.636\pm0.002$ & $0.357\pm0.057$ & $0.572\pm0.002$ & $0.209\pm0.064$ & $0.543\pm0.002$ & $0.088\pm0.067$ \\
 & \textsc{Individual} & $0.655\pm0.002$ & $0.420\pm0.054$ & $0.650\pm0.002$ & $0.403\pm0.055$ & $0.608\pm0.002$ & $0.311\pm0.060$ \\
 & \textsc{Metabench} & $0.573\pm0.001$ & $0.211\pm0.064$ & $0.582\pm0.001$ & $0.238\pm0.064$ & $0.572\pm0.001$ & $0.215\pm0.064$ \\
 & \textsc{TinyBenchmarks} & $0.581\pm0.001$ & $0.237\pm0.064$ & $0.584\pm0.001$ & $0.245\pm0.063$ & $0.553\pm0.001$ & $0.155\pm0.066$ \\
 & \textsc{BenchAlign} & $\mathbf{0.745\pm0.001}$ & $\mathbf{0.683\pm0.034}$ & $\mathbf{0.726\pm0.001}$ & $\mathbf{0.622\pm0.040}$ & $\mathbf{0.614\pm0.002}$ & $\mathbf{0.330\pm0.059}$ \\
\bottomrule
\end{tabular}
}
\caption{Alignment with other preference dimensions under model scale splits.}
\label{tab:model_size_splits_other_dimensions}
\end{table*}

\begin{table*}[t]
\centering
\small
\setlength{\tabcolsep}{5pt}
\renewcommand{\arraystretch}{1.05}

\resizebox{\textwidth}{!}{
\begin{tabular}{c c c c c c c c }
\toprule
 \multirow{3}{*}{\makecell{\textbf{Target}\\\textbf{Models}}} & \multirow{3}{*}{\textbf{Method}} & \multicolumn{2}{c}{\textbf{Coherence}} & \multicolumn{2}{c}{\textbf{Complexity}} & \multicolumn{2}{c}{\textbf{Verbosity}} \\
\cmidrule(lr){3-4} \cmidrule(lr){5-6} \cmidrule(lr){7-8}

& & $Acc_{pair}$ & $\rho$ & $Acc_{pair}$ & $\rho$ & $Acc_{pair}$ & $\rho$ \\
\midrule

\multirow{4}{*}{2\%} & \textsc{No Alignment} & $0.711\pm0.015$ & $0.583\pm0.124$ & $0.697\pm0.015$ & $0.562\pm0.129$ & $0.575\pm0.016$ & $0.222\pm0.191$ \\
 & \textsc{Random} & $0.661\pm0.015$ & $0.468\pm0.150$ & $0.699\pm0.015$ & $0.561\pm0.129$ & $0.543\pm0.016$ & $0.108\pm0.204$ \\
 & \textsc{Individual} & $0.724\pm0.014$ & $0.608\pm0.118$ & $0.707\pm0.015$ & $0.576\pm0.125$ & $0.712\pm0.015$ & $0.425\pm0.158$ \\
 & \textsc{Metabench} & $0.717\pm0.010$ & $0.593\pm0.139$ & $0.702\pm0.010$ & $0.572\pm0.144$ & $0.573\pm0.011$ & $0.219\pm0.201$ \\
 & \textsc{TinyBenchmarks} & $0.704\pm0.010$ & $0.570\pm0.144$ & $0.684\pm0.011$ & $0.543\pm0.151$ & $0.576\pm0.011$ & $0.231\pm0.200$ \\
 & \textsc{BenchAlign} & $\mathbf{0.864\pm0.011}$ & $\mathbf{0.888\pm0.037}$ & $\mathbf{0.839\pm0.012}$ & $\mathbf{0.868\pm0.044}$ & $\mathbf{0.745\pm0.014}$ & $\mathbf{0.637\pm0.110}$ \\
\midrule
\multirow{4}{*}{5\%} & \textsc{No Alignment} & $0.696\pm0.006$ & $0.548\pm0.087$ & $0.703\pm0.006$ & $0.569\pm0.084$ & $0.532\pm0.006$ & $0.104\pm0.130$ \\
 & \textsc{Random} & $0.660\pm0.006$ & $0.457\pm0.099$ & $0.703\pm0.006$ & $0.566\pm0.084$ & $0.508\pm0.006$ & $0.005\pm0.133$ \\
 & \textsc{Individual} & $0.714\pm0.006$ & $0.584\pm0.081$ & $0.718\pm0.006$ & $0.585\pm0.081$ & $0.643\pm0.006$ & $0.256\pm0.120$ \\
 & \textsc{Metabench} & $0.697\pm0.004$ & $0.549\pm0.093$ & $0.703\pm0.004$ & $0.568\pm0.090$ & $0.534\pm0.004$ & $0.104\pm0.131$ \\
 & \textsc{TinyBenchmarks} & $0.691\pm0.004$ & $0.539\pm0.095$ & $0.697\pm0.004$ & $0.561\pm0.092$ & $0.539\pm0.004$ & $0.124\pm0.131$ \\
 & \textsc{BenchAlign} & $\mathbf{0.847\pm0.005}$ & $\mathbf{0.862\pm0.031}$ & $\mathbf{0.820\pm0.005}$ & $\mathbf{0.829\pm0.037}$ & $\mathbf{0.719\pm0.006}$ & $\mathbf{0.588\pm0.081}$ \\
\midrule
\multirow{4}{*}{10\%} & \textsc{No Alignment} & $0.664\pm0.003$ & $0.466\pm0.070$ & $0.678\pm0.003$ & $0.501\pm0.067$ & $0.525\pm0.003$ & $0.077\pm0.093$ \\
 & \textsc{Random} & $0.639\pm0.003$ & $0.396\pm0.076$ & $0.678\pm0.003$ & $0.495\pm0.068$ & $0.500\pm0.003$ & $-0.024\pm0.094$ \\
 & \textsc{Individual} & $0.691\pm0.003$ & $0.520\pm0.065$ & $0.688\pm0.003$ & $0.512\pm0.066$ & $0.584\pm0.003$ & $0.233\pm0.087$ \\
 & \textsc{Metabench} & $0.660\pm0.002$ & $0.457\pm0.074$ & $0.674\pm0.002$ & $0.492\pm0.071$ & $0.524\pm0.002$ & $0.072\pm0.093$ \\
 & \textsc{TinyBenchmarks} & $0.661\pm0.002$ & $0.457\pm0.074$ & $0.672\pm0.002$ & $0.489\pm0.072$ & $0.530\pm0.002$ & $0.092\pm0.093$ \\
 & \textsc{BenchAlign} & $\mathbf{0.850\pm0.002}$ & $\mathbf{0.866\pm0.022}$ & $\mathbf{0.832\pm0.002}$ & $\mathbf{0.845\pm0.025}$ & $\mathbf{0.767\pm0.003}$ & $\mathbf{0.700\pm0.045}$ \\
\bottomrule
\end{tabular}
}
\caption{Alignment with other preference dimensions under arbitrary model sets}
\label{tab:random_splits_other_dimensions}
\end{table*}

\section{BenchAlign Expanded Description}\label{ap:methods_expanded}

\subsection{A Framework for Benchmark Construction}

Similar to educational testing, previous work has argued that not all test items are equally informative for model evaluation~\citep{rodriguez-etal-2021-evaluation}. This idea has led to methods for improving benchmark predictability~\citep{zhang2025how}. The most prominent approach is benchmark distillation~\citep{vivek-etal-2024-anchor,Polo2024TinyBenchmarks:Examples,Kipnis2025Metabench:Models}, which constructs new benchmarks by selecting a smaller subset of test items that preserves the predictive power of model performances on the benchmark. This approach is equivalent to applying a binary weight vector $\bm{w} \in \{0,1\}^{M}$, where $w_m=1$ for selected items and $w_m=0$ for the rest. 

While benchmark distillation methods provide a more efficient way to evaluate model performance, recent work by \citet{rodriguez-etal-2021-evaluation} shows that the success of these methods primarily relies on \emph{model similarity}: they do well on similar models but fail to generalize when source models (used for constructing new benchmarks) and target models are drawn from different distributions.

Rather than assuming uniform relevance or applying binary weights to test items, we propose reformulating the benchmark scoring function with a continuous weight vector $\bm{w} \in \mathbb{R}$. Each element $w_m$ indicates the learned relevance of test item $q_m$ to the target preference. The new scoring function $s_{\bm{w}}$ computes the system-level score for model $f_i$ as the dot product of the normalized weights and the model's performance vector $\bm{x}_i$:
\begin{equation}\label{eq:base_accuracy}
    s_{\bm{w}}(f_i,Q)= \sum_{m=1}^{M} \tilde{w}_m x_{i,m} = \bm{\tilde{w}}^{\mathsf{T}}\bm{x}_i
\end{equation}
where $\bm{\tilde{w}}$ is a normalized weight vector, computed as $w_m / \sum_{j=1}^{M}w_j$. While the instance-level score $x_{i,m}$ varies by tasks, we assume a binary setting where $x_{i,m}\in\{0,1\}$, which includes common metrics like accuracy. If $x_{i,m}$ is determined by correctness, the scoring function $s$ corresponds to weighted accuracy. The system-level scores produced by $s$ are used to determine the final model ranking. By optimizing the continuous weight vector $\bm{\tilde{w}}$, we construct an aligned benchmark $\mathcal{\hat{D}}$ in which test items are scaled by their utility for predicting the target ranking. 

\subsection{Training BenchAlign}

Algorithm~\ref{alg:benchalign} summarizes our approach for training the \textsc{BenchAlign} model.

\begin{algorithm}[t]
\caption{BenchAlign}
\label{alg:benchalign}
  \begin{algorithmic}
    \STATE {\bfseries Input:} $X = [\bm{x}_{1}, ..., \bm{x}_{K}]$ feature vectors of all models
    \STATE {\bfseries Input:} $r = [r_{1}, ..., r_{K}]$ target ranking $R_{T}$ for all models
    \STATE {\bfseries Input:} $\mathcal{L}_{pair}$ loss function
    \STATE {\bfseries Input:} $b$ batch size
    \STATE {\bfseries Output:} $\bm{w}$ learned weight vector
    \STATE Initialize $\bm{w} \leftarrow \textsc{random}()$

    \STATE \# \textcolor{blue}{ordered pair generation}
    \STATE $P \leftarrow \emptyset$ \COMMENT{set of ordered pairs}
    \FOR{$i = 1$ to $K$}
      \FOR{$j = 1$ to $K$}
        \IF{$r_i < r_j$}
          \STATE $P \leftarrow P \cup \{(i,j)\}$
        \ENDIF
      \ENDFOR
    \ENDFOR

    \STATE \# \textcolor{blue}{learning-to-rank}
    \FOR{each epoch}
    \STATE Shuffle $P$
    \FOR{each batch $B \subset P$ of size $b$}
      \STATE $L \leftarrow 0$
      \FORALL{$(i,j) \in B$}
        \STATE Compute scores $s_i=\bm{w}^{\mathsf{T}}\bm{x}_i$ and $s_j=\bm{w}^{\mathsf{T}}\bm{x}_j$
        \STATE $L \leftarrow L + \mathcal{L}_{pair}(s_i, s_j)$
      \ENDFOR
      \STATE $L \leftarrow L / |B|$
      \STATE Update $\bm{w}$ by minimizing $L$
    \ENDFOR
  \ENDFOR
  
  \STATE \textbf{return} $\bm{w}$  
  \end{algorithmic}

\end{algorithm}


\section{Learning-to-rank Algorithm for Benchmark Alignment}\label{ap:ltr_choices}

\subsection{Number of Layers}

According to our problem definition, benchmark alignment can be performed in different ways. One of them could be designing a bigger neural network than the one used for our experiments, This architecture is a deliberate choice aimed to create better and interpretable evaluations using a simple approach. As shown in Section \ref{sec:perf_model_size_splits}, this architecture is sufficient to achieve generalization, and provides a starting point towards more complex methods. Multilayer models, e.g. reward models, would likely lead to better predictions results, but cost tractability and interpretation. We delve deeper on interpretability in Appendix \ref{ap:interpretability}.

\subsection{Hyperparameter Analysis}
\begin{figure*}[t]
    \centering
    \begin{subfigure}[b]{0.49\linewidth}
        \centering
        \includegraphics[width=\linewidth]{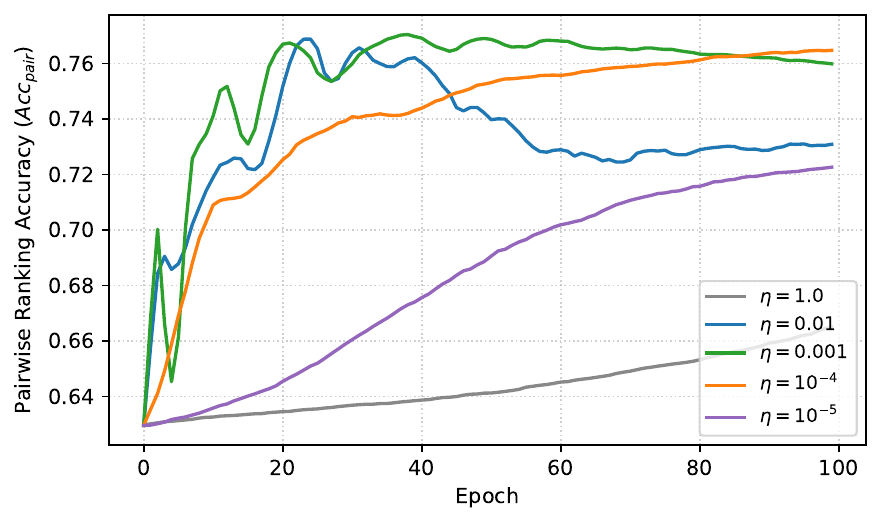}
        \caption{Pairwise ranking accuracy ($Acc_{pair}$)}
    \end{subfigure}
    \hfill
    \begin{subfigure}[b]{0.49\linewidth}
        \centering
        \includegraphics[width=\linewidth]{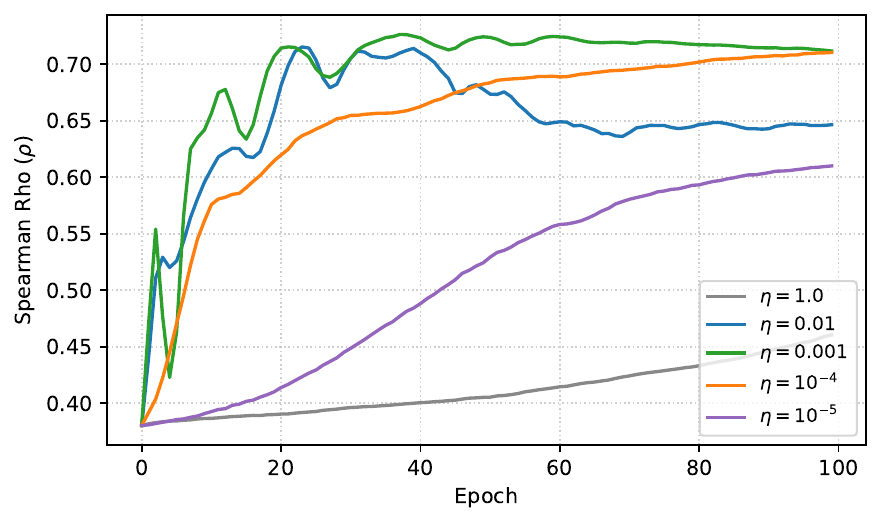}
        \caption{Spearman rank coefficient ($\rho$)}
    \end{subfigure}
    \caption{Effect of learning rate on \textsc{BenchAlign} test-time performance over $100$ epochs.}
    \label{fig:hyp_search_lr}
\end{figure*}

We complement the architectural analysis above with a hyperparameter study focused on the learning rate, the only optimization hyperparameter that materially affected \textsc{BenchAlign} performance in our experiments. Owing to the simplicity of the model, the optimization space is correspondingly small. We evaluate $\eta \in {1, 10^{-2}, 10^{-3}, 10^{-4}, 10^{-5}}$ on the $\text{ArmoRM}$ (Helpful) preference using the $30$B+ split, training each configuration for $100$ epochs and tracking test-set $Acc_{pair}$ and $\rho$ throughout training. Figure~\ref{fig:hyp_search_lr} summarizes the results. Both extremes of the sweep behave as expected. With $\eta=1$, optimization is unstable and the model quickly plateaus, while $\eta=10^{-5}$ yields updates that are too small for meaningful progress within $100$ epochs. The intermediate regime is comparatively robust: both $\eta=10^{-3}$ and $\eta=10^{-4}$ converge to roughly $Acc_{pair}\approx0.76$ and $\rho\approx0.71$. The primary difference between the two lies in convergence speed, with $\eta=10^{-3}$ reaching its plateau after roughly $30$ epochs and $\eta=10^{-4}$ requiring closer to $60$. Although $\eta=10^{-2}$ converges rapidly, it stabilizes at slightly worse final metrics, suggesting mild overshooting.

\subsection{Constraining Items}

\begin{figure*}[t]
    \refstepcounter{figure}
    \centering
    \begin{minipage}[t]{0.48\textwidth}
        \centering
        \includegraphics[trim=0 0 0 0.8cm, clip, width=\linewidth]{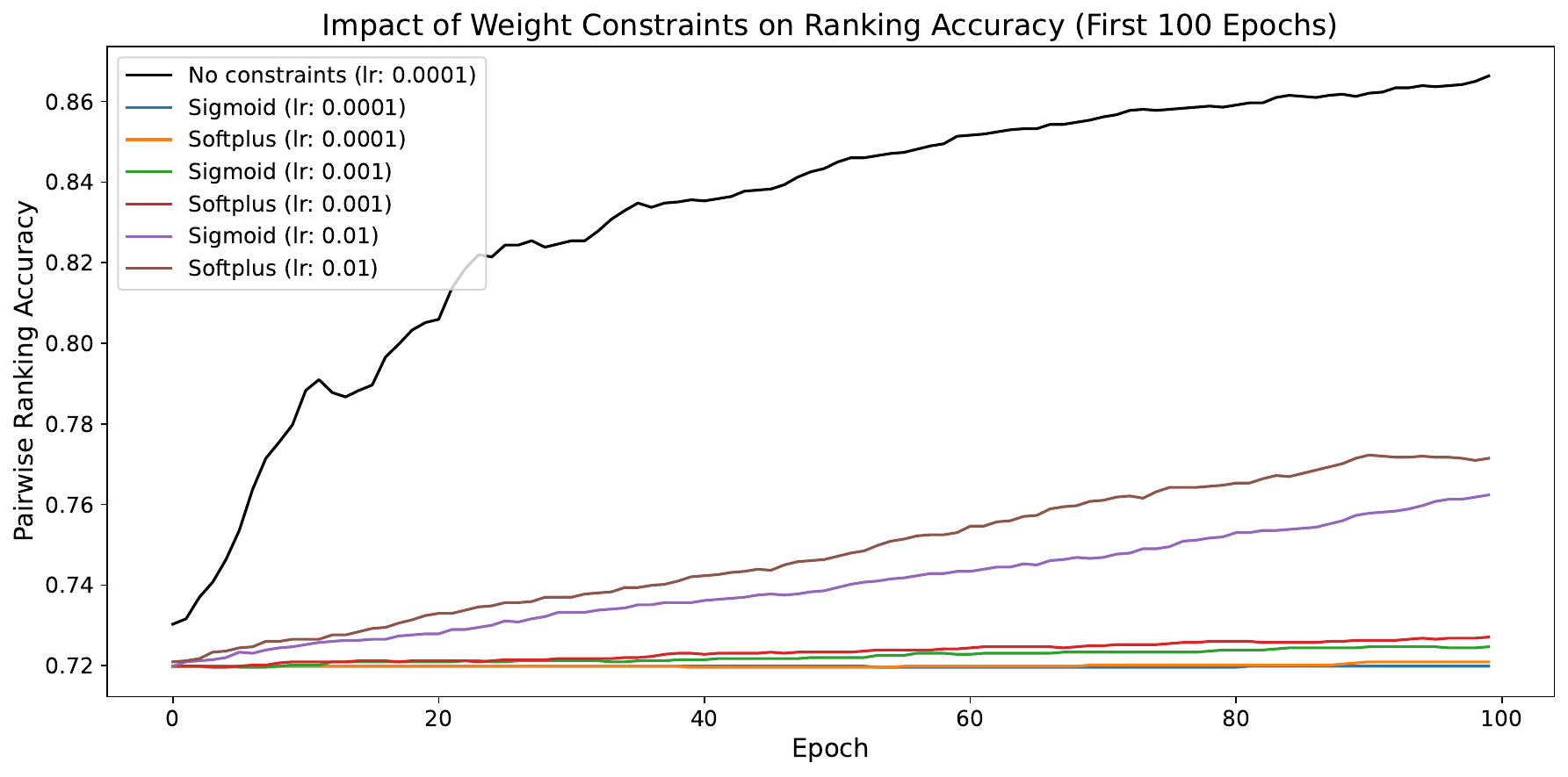}
        \vspace{2pt}
        {\small (a) Pairwise ranking accuracy ($Acc_{pair}$)}
    \end{minipage}\hfill
    \begin{minipage}[t]{0.48\textwidth}
        \centering
        \includegraphics[trim=0 0 0 0.8cm, clip, width=\linewidth]{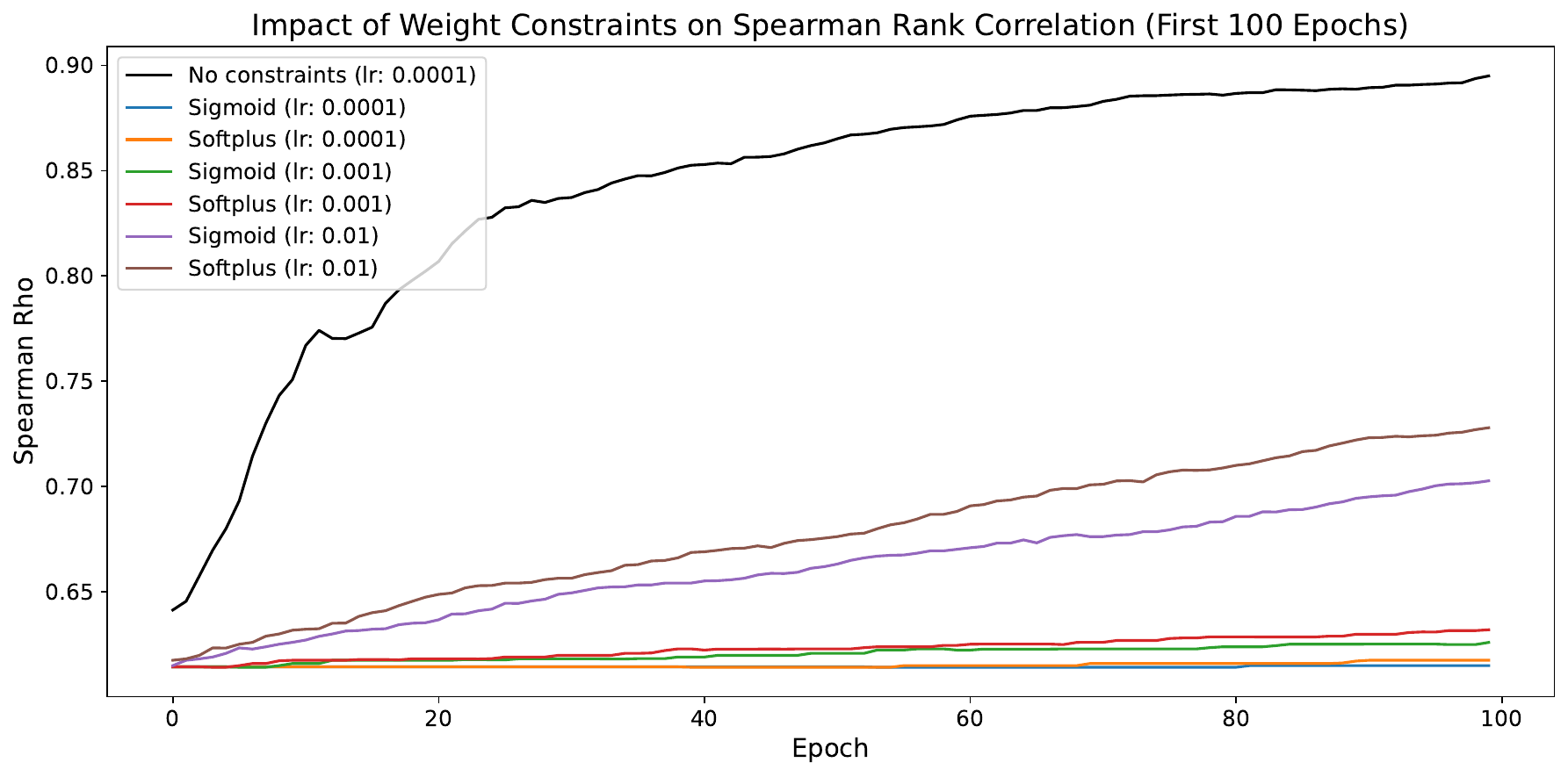}
        \vspace{2pt}
        {\small (b) Spearman rank coefficient ($\rho$)}
    \end{minipage}

    \vspace{4pt}

    \caption{Comparison of ranking performance of unconstrained vs. constrained weights over the first 100 training epochs.}

    \label{fig:constrained_weights}
\end{figure*}

We don’t impose non-negativity constraints as they remove useful signals from questions that are used for ranking a model lower according to the target preferences. To prove that these constraints would limit alignment, we replicate our $\text{ArmoRM}$ (Helpful) experiment on the 2\% target set from Table \ref{tab:random_splits_helpsteer_ultrafeedback} with 8 different configurations that impose a non-negativity constraint as shown in Figure \ref{fig:constrained_weights}

\subsection{Interpretability of Benchmark Alignment}\label{ap:interpretability}

We design a deliberately challenging setting that forces alignment to recover signal from latent relationships across available benchmark tasks. Specifically, we chose the ranking from MMLU-Pro Physics as our target preferences, using the 13B+ models as target\footnote{In this setup, we removed the ties from the target as RankNet based approaches are designed for a strict preference ordering. Hence, giving the tied models an arbitrary order would be a source of noise against \textsc{BenchAlign}} and six candidate tasks where only one, Math Geometry Hard, has a direct but mild relationship with the target\footnote{Math Geometry Hard has the lowest train correlation with the target preferences among all STEM benchmark tasks}. The remaining candidates are the five tasks with the lowest target correlation on the train split. The benchmarks and number of questions are indicated in Table \ref{tab:mmlu_pro_physics_benchmarks}

\begin{table}[H]
\centering
\caption{Benchmarks used for MMLU-Pro Physics target preferences}
\begin{tabular}{llc}
\toprule
Benchmark  &Task& Number of Questions \\
\midrule
IFEval&Instruction Following& 541 \\
MUSR&Object Placements& 256 \\
BBH&Sports Understanding& 250 \\
BBH&Tracking Shuffled Three Objects& 250 \\
MUSR&Team Allocation& 250 \\
MATH&Geometry Hard& 132 \\
\midrule
Total  && 1679 \\
\bottomrule
\end{tabular}
\label{tab:mmlu_pro_physics_benchmarks}
\end{table}

First, we analyze how benchmark alignment determines the least and most relevant tasks for benchmark alignment. As explained in Section \ref{sec:proposed_method}, \textsc{BenchAlign} learns weights that capture the relevance of every question with respect to the target preferences. These weights range between -1 and 1~\citep{Burges2005LearningDescent}, such that questions with weights close to the extremes are the most informative for discriminating between models and determining their ranking. In contrast, questions with weights close to 0 have little to no influence. For simplicity, we take the absolute value of these weights to focus on the least and most relevant questions.

\begin{figure*}[h]
    \centering
    \begin{minipage}[h]{0.5\textwidth}
        \centering
        \includegraphics[trim=0 0 0 2cm, clip, width=\linewidth]{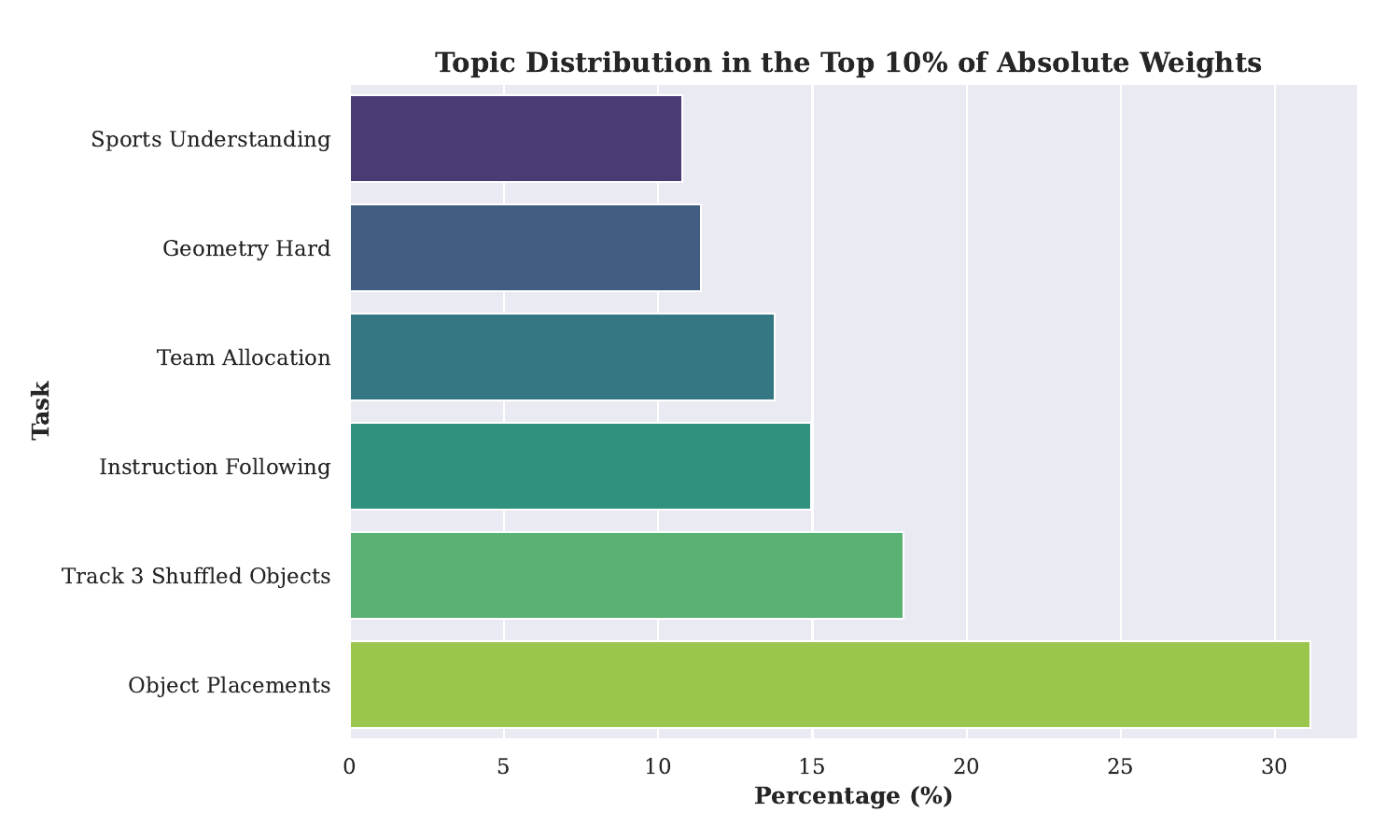}
        \vspace{2pt}
        {\small (a) Top 10\%}
    \end{minipage}\hfill
    \begin{minipage}[h]{0.5\textwidth}
        \centering
        \includegraphics[trim=0 0 0 2cm, clip, width=\linewidth]{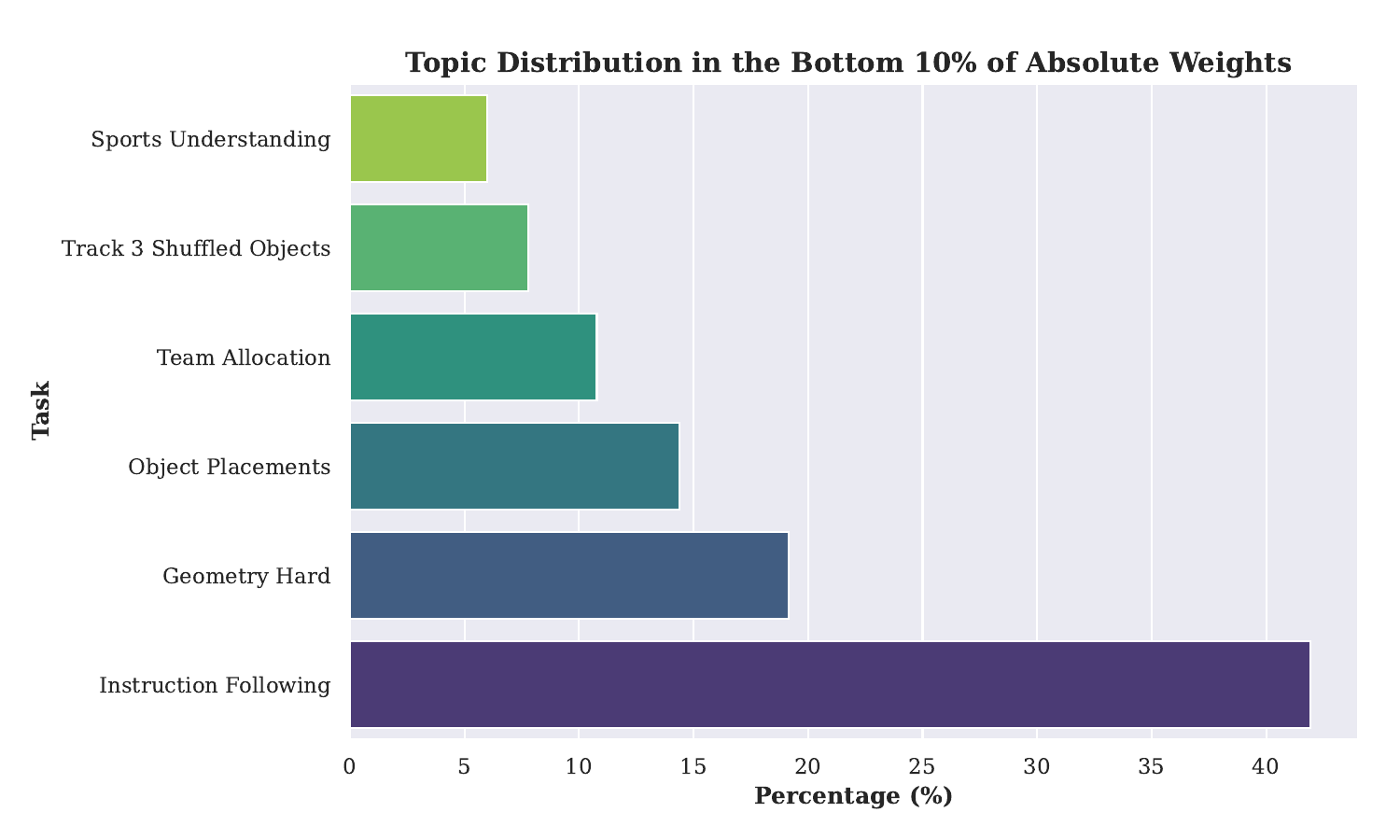}
        \vspace{2pt}
        {\small (b) Bottom 10\%}
    \end{minipage}

    \caption{Topic distribution of questions in the top 10\% and bottom 10\% by absolute weight learned by \textsc{BenchAlign}.}
    \label{fig:physics_case_study1}
\end{figure*}

Figure \ref{fig:physics_case_study1} shows the distribution of tasks among questions in the top and bottom $10\%$ of absolute weights. A surprising finding is that questions from all six benchmarks appear in both the top and bottom weight groups. While it would be expected to find questions from a task related to the target such as Geometry to be the dominant group within the top 10 \%, we find that Instruction Following, an unrelated benchmark, is ahead by a large margin. These results suggest that \textsc{BenchAlign} draws signal from questions across all available benchmarks, rather than relying on topic relationship between candidate benchmarks and the target preferences.

To further investigate this finding, we also perform a qualitative assessment of the least and most relevant questions in Table \ref{tab:physics_case_questions}. We do not observe an obvious semantic pattern indicating that highly weighted questions are more closely related to physics than low-weight questions. For example, two of the questions with the highest absolute weights involve narrative reasoning about the location of an object in a story, while another involves writing a rap song. These examples appear unrelated to physics. In contrast, several of the questions with near-zero weights are drawn from the Math Geometry Hard benchmark, which is arguably more semantically related to physics. These observations suggest that the learned weights do not capture surface-level semantic similarity between benchmarks and the target task.

\begin{table*}[t]
\centering
\small
\caption{Examples of questions with the highest and lowest absolute weights learned by \textsc{BenchAlign}.}
\resizebox{\columnwidth}{!}{%
\begin{tabular}{p{0.06\linewidth} lp{0.2\linewidth} p{0.74\linewidth}}
\toprule
$\lvert w \rvert$ &  Benchmark&Task& Question \\
\midrule
0.0118 &  MUSR&Object Placements &
In the quiet afternoon, Madison was preparing for an important parent–teacher conference when she realized her gradebook was missing. While searching the classroom with Rachel and the school custodian Alex, several objects were moved around the room... \textbf{Question:} Which location is the most likely place Madison would look
to find the gradebook? \\
0.0117 &  IFEval&Instruction Following &
Generate a list of 100 random names. Make sure that no name is repeated and every name is unique. All letters in your entire response should be capitalized. Italicize 5 of your favorite names. \\
0.0117 &  MUSR&Object Placements & Charlie had just finished his novel and stored the manuscript in the cupboard while working with his assistant Lisa and roommate Matthew. As they moved around the room during various tasks, the manuscript and other objects were relocated...
\textbf{Question:} Which location is the most likely place Lisa would look to find the manuscript? \\
0.0117 &  IFEval&Instruction Following & Can you write a rap that doesn't include the keywords ``Yo", ``check", and ``peace"? \\
0.0117 &  MATH&Geometry Hard& A 10-cm stick has a mark at each centimeter. By breaking the stick at two of these nine marks at random, the stick is split into three pieces, each of integer length. What is the probability that the three lengths could be the three side lengths of a triangle? Express your answer as a common fraction. \\
\midrule
0.0 &  MATH&Geometry Hard&
In acute triangle $ABC$, altitudes $AD$, $BE$, and $CF$ intersect at the orthocenter $H$.  If $BD = 5$, $CD = 9$, and $CE = 42/5$, then find the length of $HE$.\\
0.0 &  MUSR&Object Placements & Oliver was preparing for a music school audition while practicing violin with support from Peter and supervision from Evelyn. As they moved around the room during practice, instruments and music sheets were relocated between the case, stand, and piano bench... \textbf{Question:} Which location is the most likely place Peter would look to find the violin bow? \\
0.0 &  MATH&Geometry Hard& In triangle $ABC$, $AB = 17$, $AC = 8$, and $BC = 15$.  Let $D$ be the foot of the altitude from $C$ to $AB$.  Find the area of triangle $ACD$.\\
0.0 &  MATH&Geometry Hard& Let $ABCD$ be a regular tetrahedron with side length 2. The plane parallel to edges $AB$ and $CD$ and lying halfway between them cuts $ABCD$ into two pieces. Find the surface area of one of these pieces.\\
0.0 &  MATH&Geometry Hard& In $\triangle ABC$ we have $AB=7$, $AC=8$, and $BC=9$. Point $D$ is on the circumscribed circle of the triangle so that $\overline{AD}$ bisects $\angle BAC$. What is the value of $AD/CD$?\\
\bottomrule
\end{tabular}
}
\label{tab:physics_case_questions}
\end{table*}

\begin{figure*}[t]
    \centering
    \begin{minipage}[t]{0.48\textwidth}
        \centering
        \includegraphics[width=\linewidth]{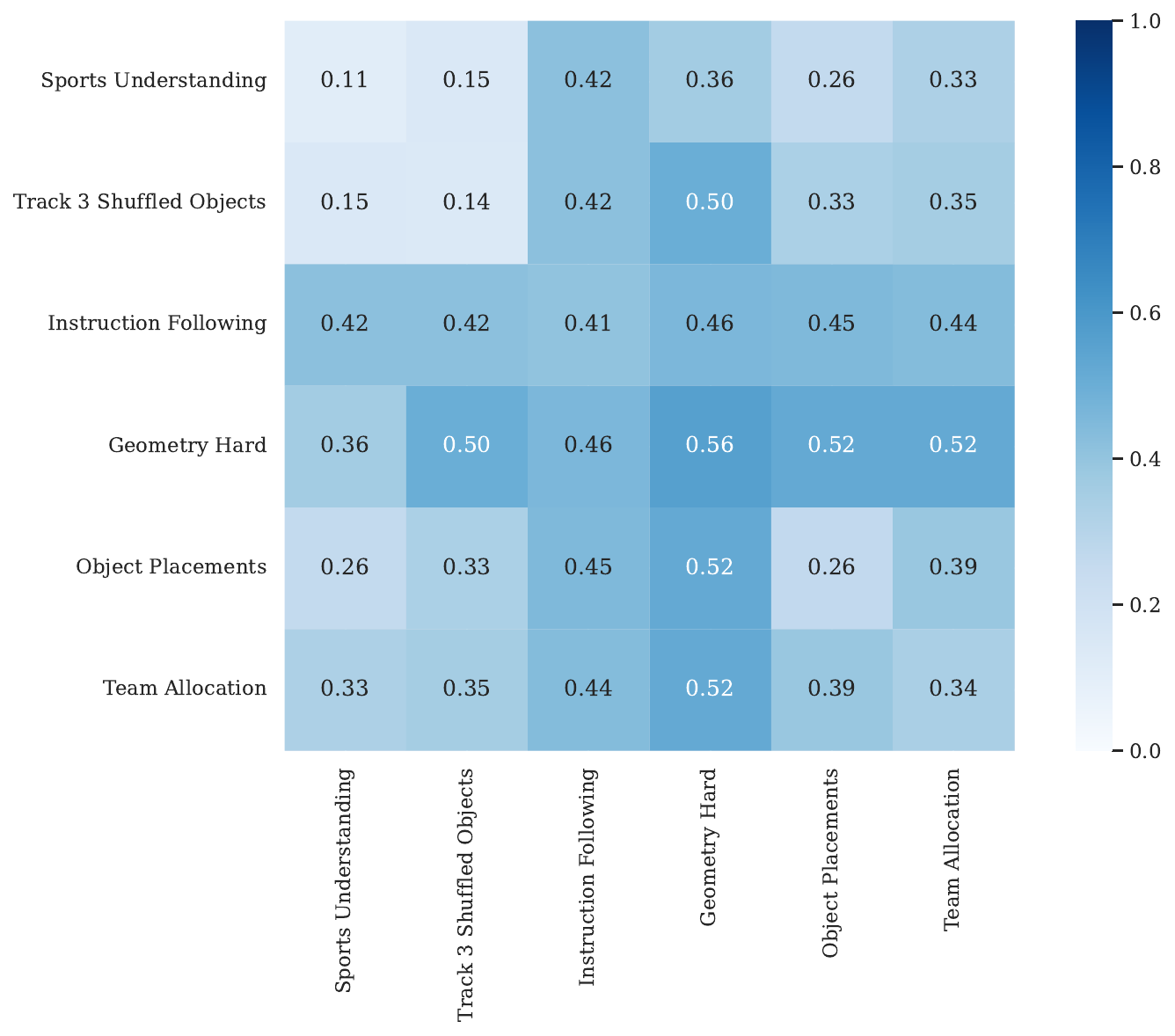}
        \vspace{2pt}
        {\small (a) Before \textsc{BenchAlign}}
    \end{minipage}\hfill
    \begin{minipage}[t]{0.48\textwidth}
        \centering
        \includegraphics[width=\linewidth]{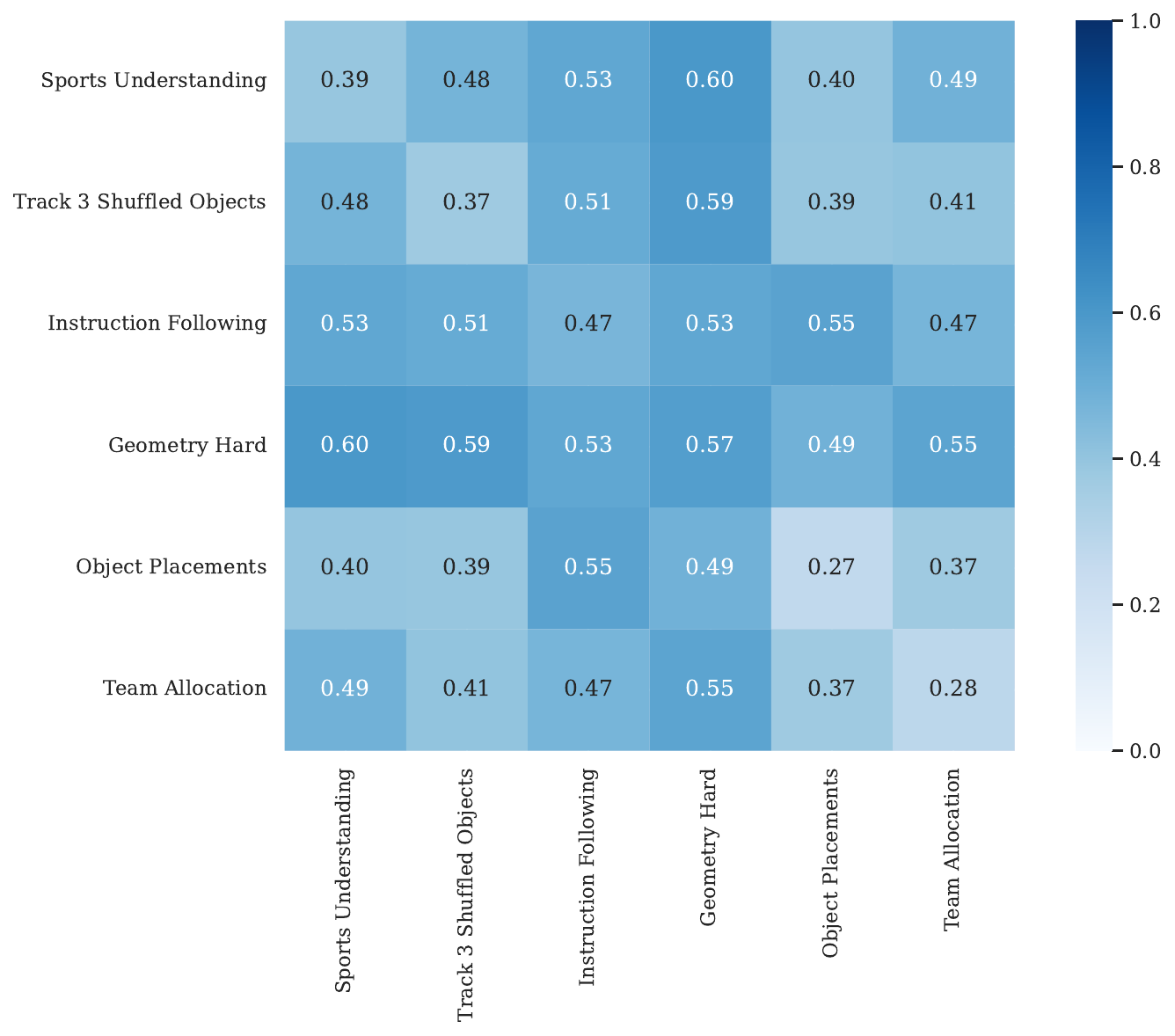}
        \vspace{2pt}
        {\small (b) After \textsc{BenchAlign}}
    \end{minipage}

    \caption{Target correlation between benchmark-induced model rankings and target ranking. The diagonal shows the correlation obtained when using each benchmark individually. Off-diagonal cells show the correlation obtained when using two joint benchmarks.}
    \label{fig:physics_case_study}
\end{figure*}

\begin{figure*}[t]
    \refstepcounter{figure}
    \centering
    \begin{minipage}[t]{0.48\textwidth}
        \centering
        \includegraphics[width=\linewidth]{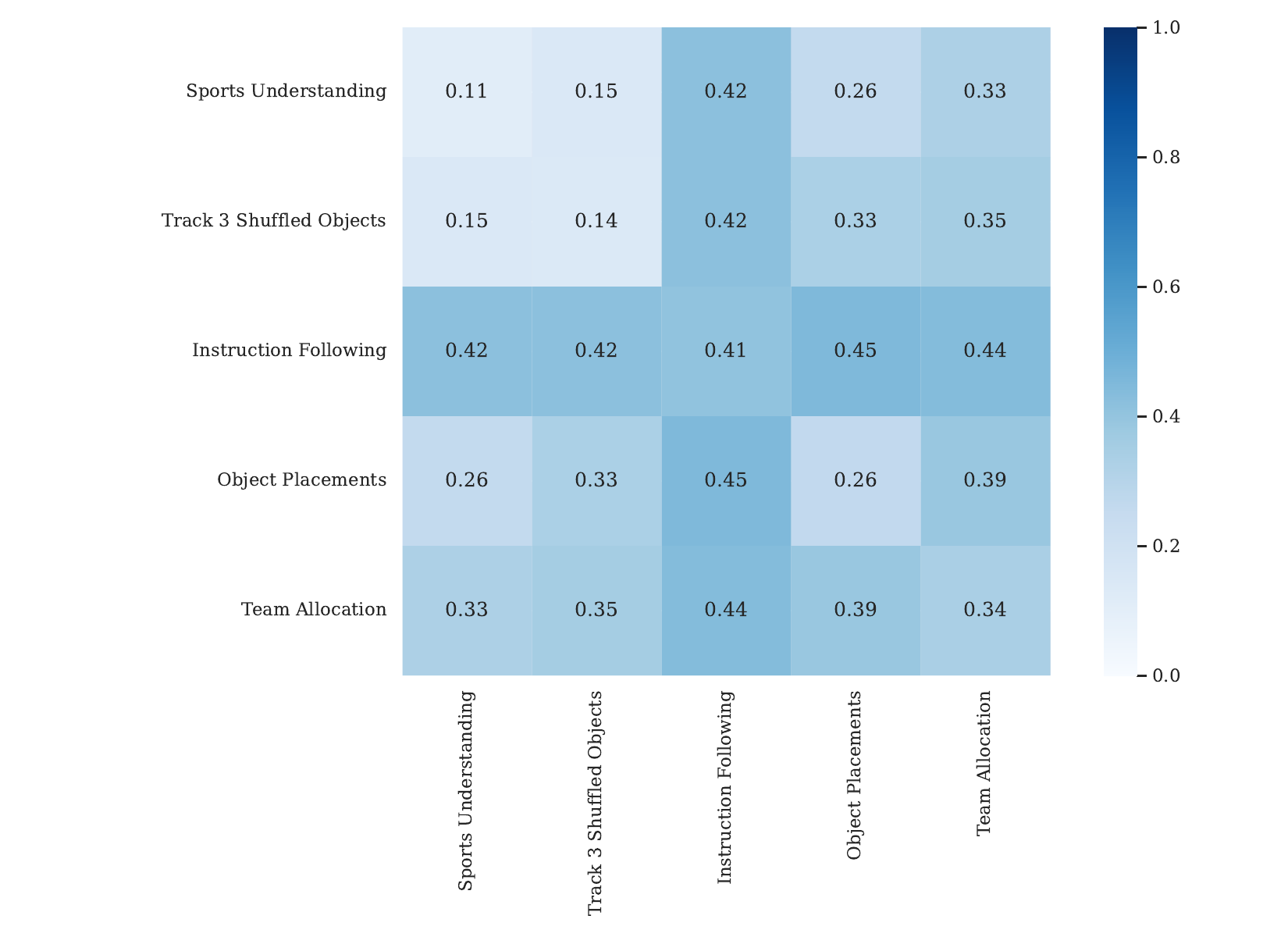}
        \vspace{2pt}
        {\small (a) Before \textsc{BenchAlign}}
    \end{minipage}\hfill
    \begin{minipage}[t]{0.48\textwidth}
        \centering
        \includegraphics[width=\linewidth]{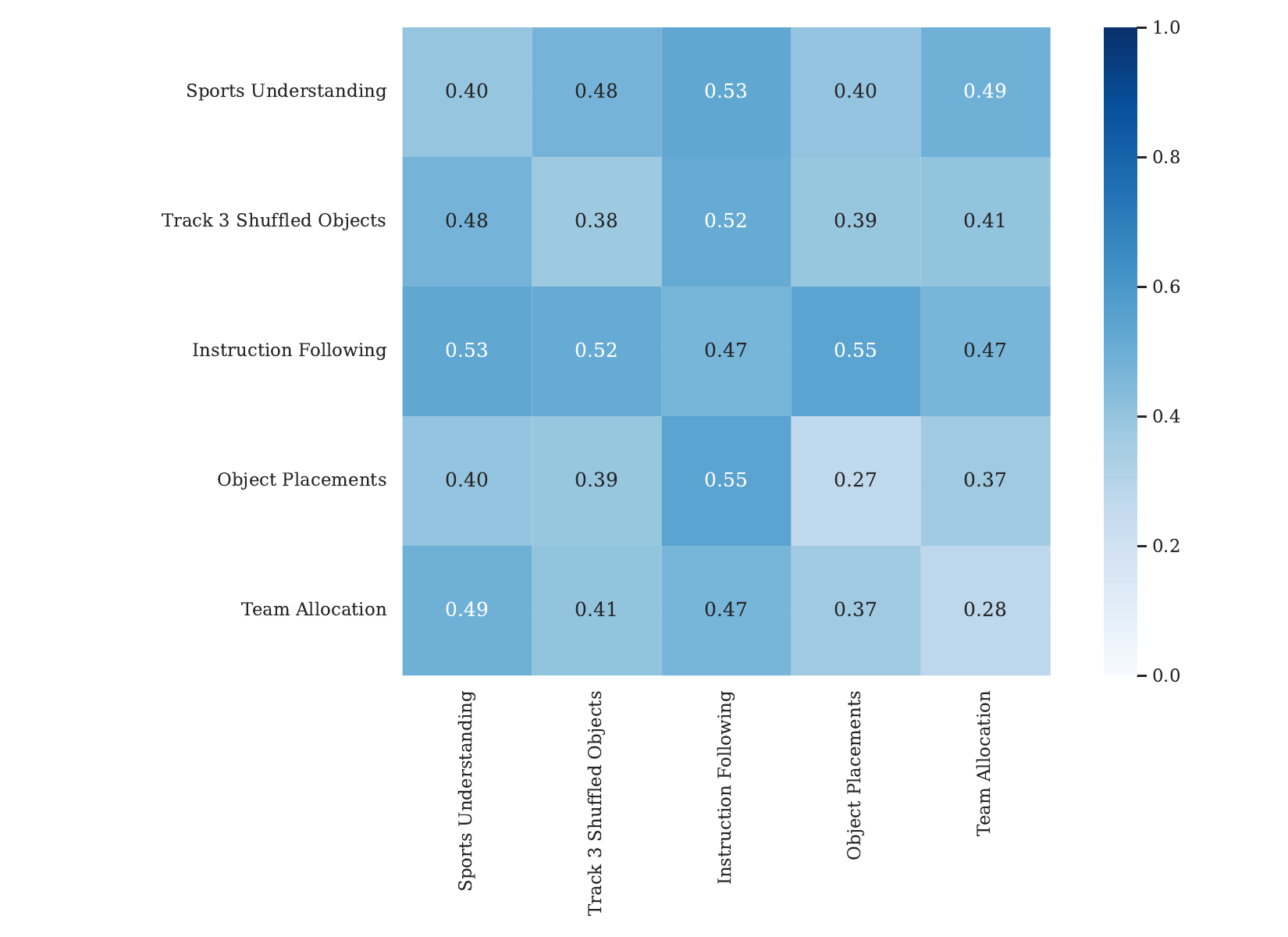}
        \vspace{2pt}
        {\small (b) After \textsc{BenchAlign}}
    \end{minipage}

    \vspace{4pt}

    \caption{Substantially Different Benchmarks. Target correlation between benchmark-induced model rankings and target ranking. The diagonal shows the correlation obtained when using each benchmark individually. Off-diagonal cells show the correlation obtained when using two joint benchmarks.}
    \label{fig:physics_case_study_subst_diff}
\end{figure*}

\begin{figure*}[t]
    \refstepcounter{figure}
    \centering
    \begin{minipage}[t]{0.48\textwidth}
        \centering
        \includegraphics[width=\linewidth]{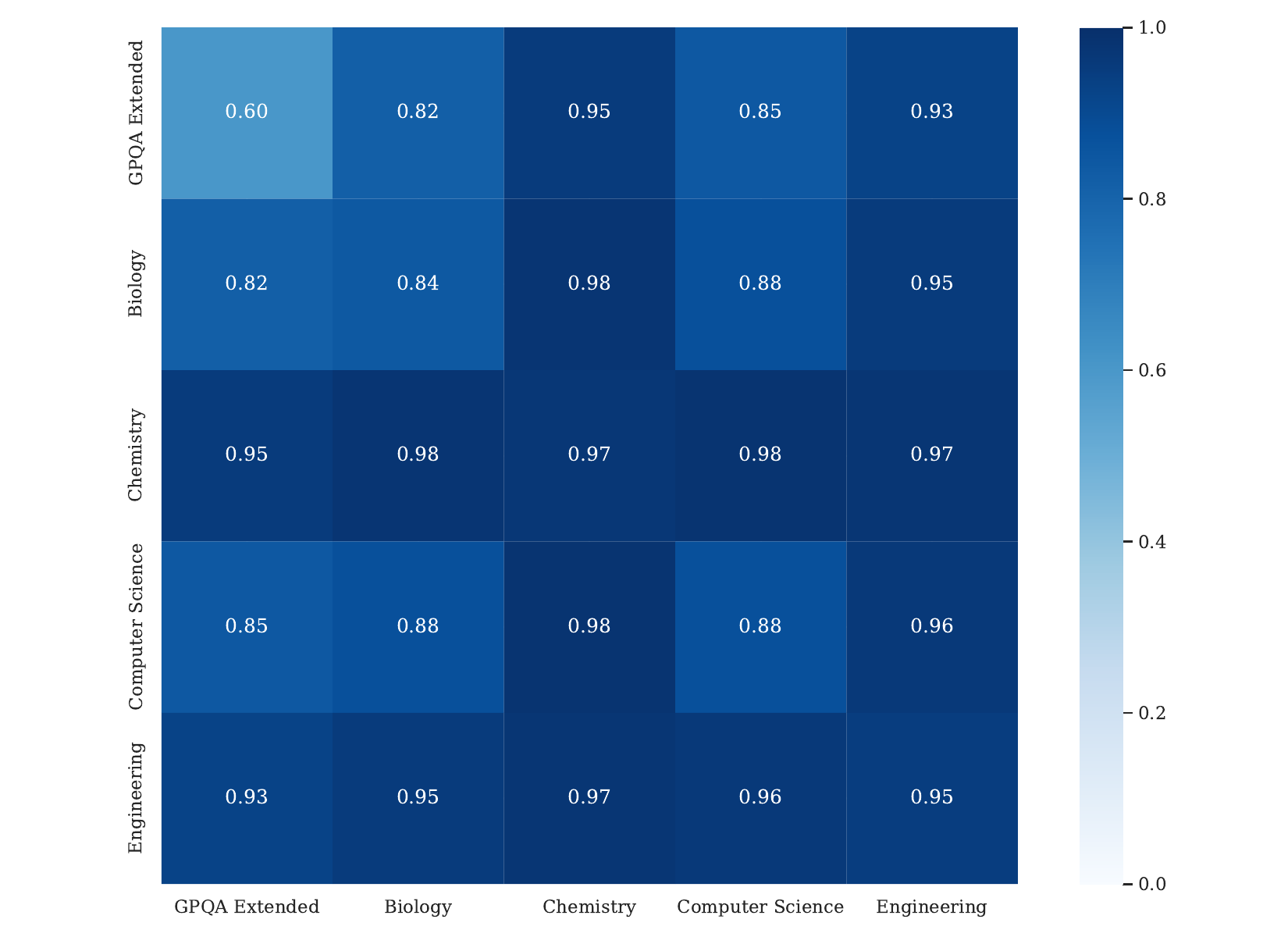}
        \vspace{2pt}
        {\small (a) Before \textsc{BenchAlign}}
    \end{minipage}\hfill
    \begin{minipage}[t]{0.48\textwidth}
        \centering
        \includegraphics[width=\linewidth]{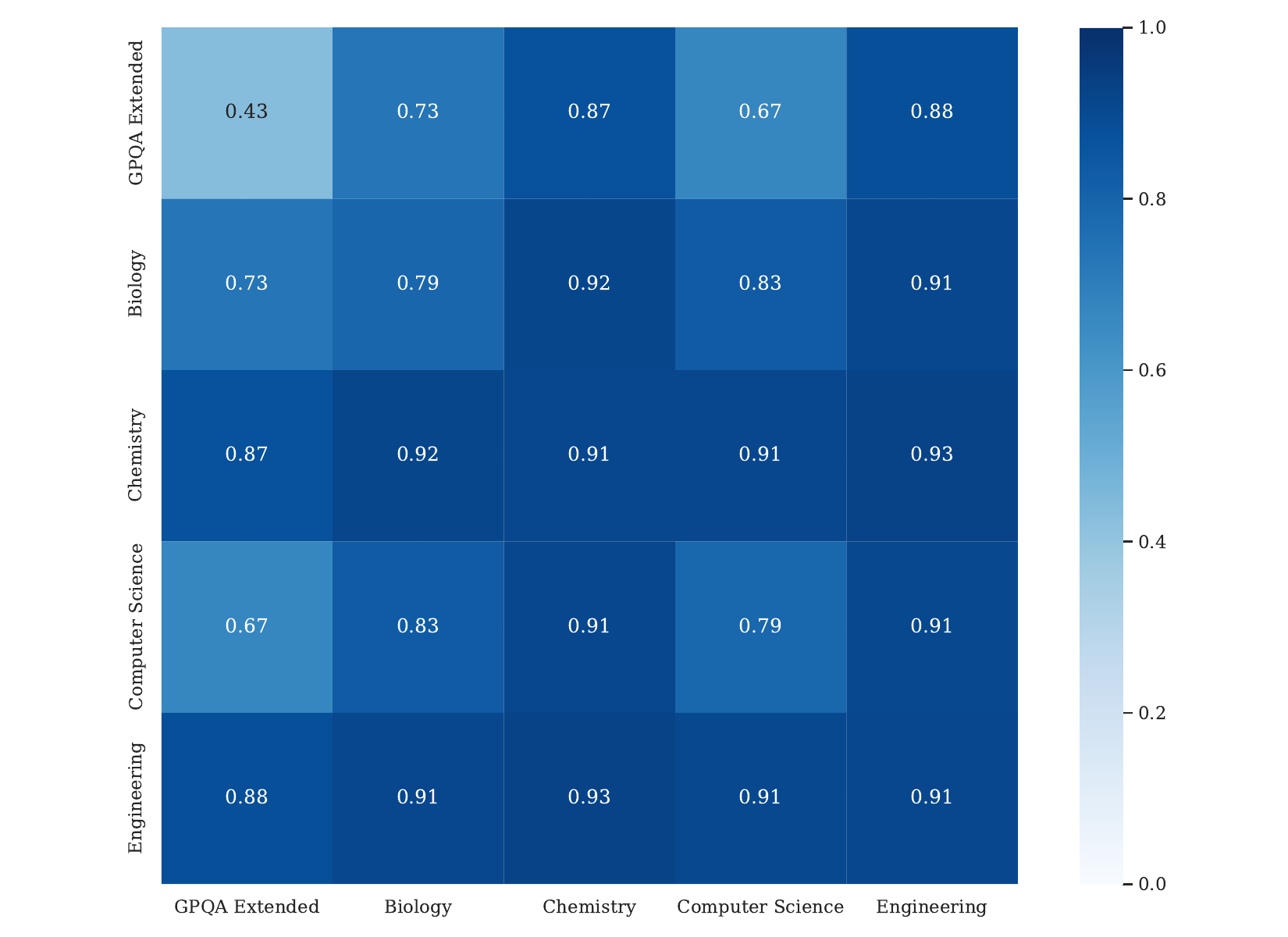}
        \vspace{2pt}
        {\small (b) After \textsc{BenchAlign}}
    \end{minipage}

    \vspace{4pt}

    \caption{Highly Relevant Benchmarks. Target correlation between benchmark-induced model rankings and target ranking. The diagonal shows the correlation obtained when using each benchmark individually. Off-diagonal cells show the correlation obtained when using two joint benchmarks.}
    \label{fig:physics_case_study_highly_sim}
\end{figure*}




We next analyze how benchmark alignment changes the task-level correlations with the target preferences. Figure \ref{fig:physics_case_study} provides a clearer picture on how our approach performs alignment.  The largest improvements occur for Sports Understanding and Tracking 3 Shuffled Objects, while Geometry Hard remains the most correlated benchmark overall, with its correlation increasing by only 0.01. At the task level, \textsc{BenchAlign} increases the target correlation of the lowest correlated tasks while keeping the correlation of the remaining ones around their same value. However, a more interesting finding is observed in two-task combinations as they show substantially larger improvements. Particularly, the correlation of Sports Understanding and Tracking 3 Shuffled Objects combined increases from approximately 0.15 to 0.48. In most combinations, correlation increases after applying \textsc{BenchAlign}, while the few combinations that decrease do so by at most 0.06. These results suggest that the weights learned by \textsc{BenchAlign} amplify complementary signals across benchmarks rather than simply reinforcing the most individually correlated benchmarks or questions. 

An important implication is that the learned weights do not merely reflect how strongly each benchmark task correlates with the target ranking in isolation. Instead, they encode how individual questions contribute to ranking models according to the latent capabilities they measure, both individually and in interaction with questions from other benchmarks. Benchmark alignment therefore identifies cross-task relationships that may not be evident from benchmark-level correlations alone.

In addition, we designed an experiment where the target is a MMLU Physics Pro and the candidate benchmarks are non-STEM and have the lowest correlations to the target (Figure \ref{fig:physics_case_study_subst_diff}). We measured the individual and joint correlations before and after alignment, ultimately finding that alignment recovers more meaningful rankings by finding latent relationships between benchmarks as shown by the correlation obtained when combining two tasks.

We replicated the same experiment as above but considering candidate benchmarks that were STEM and have the highest correlations to the target (Figure \ref{fig:physics_case_study_highly_sim}). The individual and joint correlations show that BenchAlign marginally reduces the relevance of these benchmarks. This is expected as performing benchmark alignment on evaluations already aligned with the target introduces noise in the signals captured by their test items.

Finally, we determine whether the question sets found by alignment are stable for a set of preferences across different seeds and not randomly determined by repeating the 5\% and 10\% experiments on $RM_{1}^{\text{Judge}}$ from Appendix \ref{ap:pref_heterog} (Preference Heterogeneity) using 5 different seeds to determine whether the high-weight questions from the initial results are stable. The target metric was the overlap (\%) of the top 10\% questions with the highest weights across seeds (Table \ref{tab:top_weighted_stability}). Our results show that BenchAlign finds a consistent set of questions for the same preferences regardless of and varying conditions. 

\begin{table}[t]
\centering
\caption{Stability of the top-weighted questions across different random seeds for 5\% and 10\% test sets.}
\label{tab:top_weighted_stability}
\begin{tabular}{lcccc}
\toprule
& \multicolumn{2}{c}{5\% Test Set} & \multicolumn{2}{c}{10\% Test Set} \\
\cmidrule(lr){2-3}\cmidrule(lr){4-5}
& Overlap (\%) & Count & Overlap (\%) & Count \\
\midrule
Seed 1 & 91.77 & 1349 / 1470 & 87.14 & 1294 / 1485 \\
Seed 2 & 91.50 & 1345 / 1470 & 85.99 & 1277 / 1485 \\
Seed 3 & 90.95 & 1337 / 1470 & 86.67 & 1287 / 1485 \\
Seed 4 & 91.09 & 1339 / 1470 & 86.53 & 1285 / 1485 \\
Seed 5 & 91.50 & 1345 / 1470 & 85.59 & 1271 / 1485 \\
\midrule
Mean $\pm$ Std & $\mathbf{91.36 \pm 0.30}$ & & $\mathbf{86.38 \pm 0.54}$ & \\
\bottomrule
\end{tabular}
\end{table}

\subsection{Compute Resources and Runtime Analysis}\label{ap:runtime_analysis}

We run our experiments on the HPC of the University of Virginia using the GPUs provided by the system based on availability. To assess the specific compute requirements and runtime analysis, we replicated our experiments from Section \ref{sec:perf_num_models} from Figure \ref{fig:bench_align_num_models} with different number of epochs, learning rates and number of models on a NVIDIA RTX A6000 with 48 GBs of VRAM. We show our results in Table \ref{tab:runtime_analysis}. As observed, the biggest factor influencing runtime is the number of epochs, while the number of models and learning rate have little effect on training time.

\begin{table}
    \centering
    \caption{Runtime Analysis of Benchmark Alignment in seconds}
    \begin{tabular}{cccccc}
    \toprule
 & & \multicolumn{4}{c}{Number of Epochs}\\ 
         Number of Models&  Learning Rate&  100&  1000&  5000& 10000\\ \midrule
         &  0.0001&  14.0&  154.0&  810.0& 941.0\\
         3700&  0.001&  17.0&  135.0&  459.0& 933.0\\
         &  1&  15.0&  162.0&  646.0& 969.0\\ \midrule
         &  0.0001&  12.0&  115.0&  731.0& 836.0\\
         2000&  0.001&  17.0&  95.0&  537.0& 966.0\\
         &  1&  11.0&  118.0&  602.0& 1112.0\\ \midrule
         &  0.0001&  7.0&  109.0&  474.0& 1199.0\\
         1000&  0.001&  13.0&  156.0&  581.0& 501.0\\
         &  1&  6.0&  108.0&  599.0& 888.0\\ \midrule
         &  0.0001&  16.0&  64.0&  553.0& 724.0\\
         500&  0.001&  9.0&  109.0&  443.0& 632.0\\
         &  1&  8.0&  104.0&  348.0& 995.0\\ \midrule
         &  0.0001&  8.0&  87.0&  456.0& 780.0\\
         100&  0.001&  7.0&  61.0&  309.0& 493.0\\
         &  1&  11.0&  99.0&  409.0& 607.0\\ \bottomrule
    \end{tabular}
    \label{tab:runtime_analysis}
\end{table}

\section{Baselines}\label{ap:baselines_implementation}

\subsection{No Alignment}
We aggregate all benchmarks into a single joint feature space. Then, we use the overall accuracy to rank models accordingly.

\subsection{Individual}
We first split each training benchmark into its constituent tasks. For example, MMLU Pro is subdivided into math, law, psychology, and other subject areas. We then append the benchmark name to each task (e.g., “MMLU Pro – Math”) to treat them as individual benchmarks and select the one with the highest Spearman Rank Correlation, which also happened to be the one with the highest Pairwise Ranking Accuracy in all settings.

By treating these tasks as individual benchmarks, we choose the best individual baseline from 52 options rather than the 5 available benchmarks. IFEval is not considered in this selection.

\subsection{Random Benchmark}

We use the splitted tasks and derive a numerical seed from the resulting string. This seed is used to randomly select a task from the training benchmarks, whose induced model ranking is used directly as the random baseline, while ensuring reproducibility. IFEval is not considered in this selection.

\subsection{MetaBench}

We implement the \textsc{MetaBench} baseline following the general pipeline proposed by \citet{Kipnis2025Metabench:Models} with one deliberate deviation: we do not apply the final information-filtering step based on Fisher information. Below, we describe our implementation in detail and motivate this design choice.

\subsubsection{Data preprocessing and benchmark construction}

We begin from question-level correctness data released as part of the OpenLLM Leaderboard. Following \textsc{MetaBench}, for each benchmark we compute item-level statistics on the training models only, including standard deviation, difficulty (scaled mean accuracy), and part-whole correlation between each item and the total score. Items are excluded if they (i) have near-zero variance, (ii) are excessively easy, or (iii) exhibit low discrimination. This mirrors the preprocessing steps described in Section 2.1 of \textsc{MetaBench}. To avoid information leakage, all item selection is performed strictly on the training portion of each split.

\subsubsection{Cross-validated subsampling}

After preprocessing, we follow the \textsc{MetaBench} subsampling procedure to select a subset of items. For each split, we repeatedly sample subsets of size $k$ (with $k$ ranging from $500$ to $1500$ in our experiments), compute subtest scores as mean correctness, and fit a generalized additive model (GAM) on the training models to predict the full benchmark score. Validation RMSE is estimated using five-fold cross-validation, and the subset minimizing this error is retained. This procedure closely follows Section 2.2 of \textsc{MetaBench}, with GAMs used to capture the non-linear relationship between subtest scores and full benchmark scores.

\subsubsection{IRT modeling and score prediction}

Using the selected item subsets, we fit one-dimensional IRT models (2PL/3PL/4PL depending on configuration) on the training models using the \verb|mirt| package. Latent abilities are estimated via maximum a posteriori (MAP) inference. As in \textsc{MetaBench}, we combine the estimated latent abilities with subtest scores and train a GAM to predict the target benchmark score. In our setting, the target score for each model is defined as its mean correctness across all raw benchmark items. Test performance is evaluated via RMSE between predicted and true scores on held-out models.

\subsubsection{Omission of information filtering}

\textsc{MetaBench} proposes a final information filtering stage, where items are selected based on Fisher information across quantiles of the estimated ability distribution, with hyperparameters tuned via Bayesian optimization (Section 2.4 in the original paper). We intentionally omit this step in our implementation.

This decision is empirically motivated. Across all splits and benchmarks we evaluated, applying information filtering did not further reduce test RMSE beyond what was already achieved after IRT fitting and GAM-based score prediction. While the IRT step alone typically reduced test RMSE to approximately 1, the subsequent information filtering stage consistently increased test RMSE, despite substantially shrinking the number of retained items (often to fewer than 200). 

Since we observed no improvements in the primary metric of interest (test RMSE), we excluded the information filtering step in our experiments. Importantly, this choice aligns with the stated objective of \textsc{MetaBench}: accurate score reconstruction and predictive performance on unseen models. Our results therefore suggest that, at least under our experimental conditions, \textsc{MetaBench}-style information filtering does not provide additional benefits beyond IRT-based ability estimation.

\subsection{TinyBenchmarks}
We adapt \textsc{TinyBenchmarks} \citep{Polo2024TinyBenchmarks:Examples}
to our setting by re-running their \texttt{anchor} selection routine on each of our
(benchmark, train/test split) pairs.  All steps below are repeated
independently for every pair; no information from held-out test
models is used either to choose items or to fit IRT.

\subsubsection{Anchor selection}
For each split we form the
$|\mathcal{M}_\text{train}|\times N$ binary response matrix on the
training models and fit weighted $k$-means in item space with
$K=100$ clusters and uniform item weights $1/N$, taking the best of
five random restarts.  The anchor for each cluster is the item
closest to the centroid (Euclidean), and its weight $w_i$ is the
fraction of items assigned to that cluster, so $\sum_i w_i = 1$.

\subsubsection{IRT fitting}
We fit a multidimensional 2PL IRT model on the same training-model
responses with \texttt{py-irt} using the \texttt{multidim\_2pl}
module of \citet{Polo2024TinyBenchmarks:Examples}, with $D=10$
latent dimensions, hierarchical priors, learning rate $0.1$, and
$2{,}000$ epochs.  This yields per-item parameters $(\mathbf{a}_i,
b_i)$ and per-train-model abilities
$\boldsymbol{\theta}_m$.  No items are filtered.  For each test
model we observe responses only on the 100 anchors and obtain
$\hat{\boldsymbol{\theta}}$ by maximum likelihood with item parameters held fixed.

\subsubsection{Reported metric}
Our score for test model $m$ is the na\"ive accuracy $\widehat{\mathrm{acc}}_{m}$ on the 100 selected items, and the per-split ranking is obtained by sorting test models in
descending order of $\widehat{\mathrm{acc}}_{m}$.  For completeness
we also compute the p-IRT and gp-IRT estimators of
\citet{Polo2024TinyBenchmarks:Examples} and log them as auxiliary
outputs, but they do not enter the reported results.

\section{Model Scale Experiments}
\label{ap:model_scale_experiments}

We obtain model parameter counts by querying the Hugging Face Hub API using the model identifiers provided by OpenLLMLeaderboard. When parameter information is unavailable, we approximate the number of parameters using string matching. For example, if ``7B'' appears in a model name, we assume the model has 7 billion parameters exactly.

Using this procedure, we recover parameter counts for only 3,994 of the 4,364 models for which complete benchmark data from OpenLLMLeaderboard was available; among these, 434 models require string matching.

From this set, we construct the following size-based source-target groups:
\begin{itemize}
    \item $>70$B: 142 target models and 3852 source models
    \item $>30$B: 255 target models and 3739 source models
    \item $>13$B: 845 target models and 3149 source models
\end{itemize}

\section{Ablation Studies}\label{ap:ablation}

\subsection{Learning-to-rank Algorithms}\label{ap:other_ltr}

We study the effect of learning-to-rank by comparing the performance of the five algorithms described in Table~\ref{tab:ltr_methods} against the ranking performance based on directly aggregating scores from all benchmarks, denoted by \textsc{No learning-to-rank}. All algorithms are drawn from \citet{Wang2018TheOptimization}, where they are systematically compared across multiple ranking settings, making them suitable for constructing alternative versions of \textsc{BenchAlign}. We refer readers to \citet{Wang2018TheOptimization} for implementation details. We perform our evaluation using the $\text{ArmoRM}$ (Helpful) and $\text{DPA}$ preferences on the $30$B+ target from Section~\ref{sec:perf_model_size_splits}.

\begin{table}[t]
\vspace{0.1in}
\caption{Comparison of learning-to-rank objectives.}
\vspace{0.1in}
\centering
\captionsetup{justification=centering}
\newcolumntype{C}{>{\centering\arraybackslash}X}   
\newcolumntype{J}{>{\justifying\arraybackslash}X}  
\begin{tabular}{p{0.2\columnwidth} p{0.65\columnwidth}}
\toprule
\multicolumn{1}{c}{\textbf{Method}} &
\multicolumn{1}{c}{\textbf{Description}} \\
\midrule
\textsc{RankNet} &
First Learning-to-Rank method using gradient descent. Assumes every pair of items has equal importance during training. \\
\textsc{LambdaRank} &
Reweights all pairwise orderings according to their impact on a target metric that measures the gains and losses of being assigned a position in the ranking. \\
\textsc{NDCG-Loss1} &  Reweights all pairings according to a loss function directly derived from the NDCG, rather than incorporating the metric value explicitly during optimization. \\ 
\textsc{NDCG-Loss2} &  Derived as an alternative implementation of \textsc{NDCG-Loss1} with more constrained loss values when the rank of one of the elements in the pair is too large. \\ 
\textsc{NDCG-Loss2++} & Hybrid method that linearly combines the \textsc{LambdaRank} optimization objective with \textsc{NDCG-Loss2}. \\
\bottomrule
\end{tabular}
\label{tab:ltr_methods}
\end{table}

\begin{table}[t]
\caption{Learning-to-rank Ablations}
\vspace{0.1in}
\label{tab:ltr_ablations}
\centering
\small
\setlength{\tabcolsep}{6pt}
\renewcommand{\arraystretch}{1.05}
\begin{tabular}{cc cc}
\toprule
 RM & Method & $Acc_{pair}$ &$\rho$\\ \midrule
\multirow{6}{*}{$\text{ArmoRM}$ (Helpful)}& \textsc{RankNet}& $0.765 \pm 0.005$&$0.710 \pm 0.056$\\
 & \textsc{LambdaRank}&$0.704 \pm 0.004$&$0.564 \pm 0.078$\\
 & \textsc{NDCG-Loss1}& $0.762 \pm 0.004$& $0.703 \pm 0.057$\\
 & \textsc{NDCG-Loss2}& $0.755 \pm 0.004$&$0.697 \pm 0.058$\\
 & \textsc{NDCG-Loss2++}& $0.727 \pm 0.005$&$0.621 \pm 0.070$\\
 & \textsc{No learning-to-rank}& $0.605 \pm 0.005$&$0.331 \pm 0.105$\\
\midrule
 \multirow{6}{*}{$\text{DPA}$}& \textsc{RankNet}& $0.763 \pm 0.005$&$0.713 \pm 0.055$\\
 & \textsc{LambdaRank}& $0.711 \pm 0.004$& $0.597 \pm 0.073$\\
 & \textsc{NDCG-Loss1}& $0.753 \pm 0.004$& $0.696 \pm 0.058$\\
 & \textsc{NDCG-Loss2}& $0.770 \pm 0.004$&$0.727 \pm 0.053$\\
 & \textsc{NDCG-Loss2++}& $0.723 \pm 0.005$&$0.630 \pm 0.069$\\
 & \textsc{No learning-to-rank}& $0.609 \pm 0.005$&$0.318 \pm 0.106$\\
\bottomrule
\end{tabular}
\end{table}

Table~\ref{tab:ltr_ablations} reports the ranking performance of the five learning-to-rank methods and the baseline without learning-to-rank. We find that \textsc{No learning-to-rank} yields a ranking that is positively correlated with our target preferences based on helpfulness and honesty. This suggests that there may be a potential overlap between models' capabilities and their perceived preferences on these benchmark tasks. The incorporation of learning-to-rank enables \textsc{BenchAlign} to learn benchmark weightings that better align with the target preference ranking, with all methods achieving $Acc_{pair}$ above 0.7. They also outperform around two times higher than \textsc{No learning-to-rank} baseline on the rank correlation $\rho$. These results demonstrate that learning-to-rank is effective for aligning benchmarks with target preferences.

\subsection{Generalization under Arbitrary Model Sets}\label{ap:random_splits}

In Sections \ref{sec:perf_model_size_splits}, we are curating the models that are part of our train-test splits according to the model size. As conditioning on a model's characteristics could impact our previous findings, we study the effect conditioning has on benchmark alignment by evaluating our baselines using three source-target splits formed at random, holding out $2\%$, $5\%$, $10\%$ of all models as target. 

\begin{table*}[t]
\vspace{0.1in}
\caption{Helpsteer and UltraFeedback results under arbitrary model sets}
\vspace{0.1in}
\label{tab:random_splits_helpsteer_ultrafeedback}
\centering
\small
\setlength{\tabcolsep}{5pt}
\renewcommand{\arraystretch}{1.05}

\resizebox{\textwidth}{!}{
\begin{tabular}{c c c c c c  |c c c c}
\toprule
 & & \multicolumn{4}{c |}{Helpsteer} & \multicolumn{4}{c}{UltraFeedback} \\
\cline{3-10}  

Target & Method
& \multicolumn{2}{c}{$Acc_{pair}$}
& \multicolumn{2}{c}{$\rho$}
& \multicolumn{2}{c}{$Acc_{pair}$}
& \multicolumn{2}{c}{$\rho$} \\
\cline{3-10}  

& & $ArmoRM$& $GPT2$& $ArmoRM$& $GPT2$& $ArmoRM$& $DPA$& $ArmoRM$& $DPA$\\
\midrule

\multirow{6}{*}{2\%}& \textsc{No Alignment}& $0.720\pm0.014$& $0.633\pm0.015$& $0.614\pm0.151$& $0.383\pm0.196$& $0.736\pm0.014$& $0.724\pm0.014$& $0.652\pm0.140$&$0.627\pm0.147$ \\ 
 & \textsc{Random}& $0.675 \pm 0.015$& $0.587 \pm 0.016$& $0.385 \pm 0.181$& $0.252 \pm 0.198$&  $0.739 \pm 0.014$&  $0.733 \pm 0.014$&  $0.666 \pm 0.120$&  $0.655 \pm 0.123$ \\ 
 & \textsc{Individual}& $0.733\pm0.014$& $0.675\pm0.015$& $0.637\pm0.145$& $0.464\pm0.183$& $0.750\pm0.014$& $0.733\pm0.014$& $0.680\pm0.136$&$0.648\pm0.142$ \\ 
& \textsc{Metabench}& $0.727 \pm 0.010$& $0.632 \pm 0.011$& $0.624 \pm 0.131$& $0.386 \pm 0.181$& $0.746 \pm 0.010$& $0.748 \pm 0.010$& $0.698 \pm 0.111$& $0.689 \pm 0.113$\\
& \textsc{TinyBenchmarks}& $0.713 \pm 0.014$ & $0.626 \pm 0.016$ & $0.598 \pm 0.137$ & $0.375 \pm 0.182$ & $0.729 \pm 0.014$ & $0.716 \pm 0.014$ & $0.638 \pm 0.127$ & $0.614 \pm 0.134$ \\
& \textsc{BenchAlign}& $0.866 \pm 0.011$ & $0.807 \pm 0.013$ & $\mathbf{0.895 \pm 0.035}$&$\mathbf{0.800 \pm 0.078}$&  $\mathbf{0.891 \pm 0.010}$&  $\mathbf{0.891 \pm 0.010}$&  $\mathbf{0.933 \pm 0.028}$&  $\mathbf{0.929 \pm 0.030}$\\
\midrule

\multirow{6}{*}{5\%}& \textsc{No Alignment}& $0.708\pm0.006$& $0.605\pm0.006$& $0.580\pm0.095$& $0.314\pm0.125$& $0.727\pm0.006$& $0.722\pm0.006$& $0.630\pm0.088$&$0.616\pm0.090$ \\ 
 & \textsc{Random}& $0.670 \pm 0.006$& $0.589 \pm 0.006$& $0.384 \pm 0.114$& $0.257 \pm 0.124$&  $0.715 \pm 0.006$&  $0.706 \pm 0.006$&  $0.614 \pm 0.084$&  $0.594 \pm 0.087$ \\ 
 & \textsc{Individual}& $0.722\pm0.006$& $0.639\pm0.006$& $0.611\pm0.091$& $0.379\pm0.120$& $0.747\pm0.006$& $0.737\pm0.006$& $0.669\pm0.081$&$0.648\pm0.084$ \\ 
& \textsc{Metabench}& $0.708 \pm 0.004$& $0.604 \pm 0.004$& $0.581 \pm 0.089$& $0.312 \pm 0.120$& $0.718 \pm 0.004$& $0.715 \pm 0.004$& $0.627 \pm 0.081$& $0.611 \pm 0.084$\\
& \textsc{TinyBenchmarks}& $0.703 \pm 0.006$& $0.602 \pm 0.006$ & $0.570 \pm 0.090$ & $0.312 \pm 0.120$ & $0.722 \pm 0.006$ & $0.715 \pm 0.006$ & $0.620 \pm 0.082$ & $0.605 \pm 0.085$ \\
& \textsc{BenchAlign}& $0.849 \pm 0.005$& $\mathbf{0.792 \pm 0.005}$& $0.866 \pm 0.034$& $0.762 \pm 0.056$&  $\mathbf{0.870 \pm 0.004}$&  $\mathbf{0.867 \pm 0.004}$&  $\mathbf{0.905 \pm 0.024}$&  $\mathbf{0.899 \pm 0.026}$\\
\midrule

\multirow{6}{*}{10\%}& \textsc{No Alignment}& $0.678\pm0.003$& $0.590\pm0.003$& $0.502\pm0.074$& $0.264\pm0.090$& $0.697\pm0.003$& $0.695\pm0.003$& $0.558\pm0.068$&$0.551\pm0.069$ \\ 
 & \textsc{Random}& $0.652 \pm 0.003$& $0.580 \pm 0.003$& $0.345 \pm 0.083$& $0.223 \pm 0.089$&  $0.704 \pm 0.003$&  $0.696 \pm 0.003$&  $0.588 \pm 0.062$&  $0.571 \pm 0.064$ \\ 
 & \textsc{Individual}& $0.701\pm0.003$& $0.632\pm0.003$& $0.545\pm0.070$& $0.356\pm0.085$& $0.713\pm0.003$& $0.724\pm0.003$& $0.589\pm0.065$&$0.598\pm0.064$ \\ 
& \textsc{Metabench}& $0.674 \pm 0.002$& $0.585 \pm 0.002$& $0.493 \pm 0.071$& $0.252 \pm 0.088$& $0.702 \pm 0.002$& $0.698 \pm 0.002$& $0.580 \pm 0.063$& $0.566 \pm 0.064$\\
& \textsc{TinyBenchmarks}& $0.675 \pm 0.003$ & $0.589 \pm 0.003$ & $0.494 \pm 0.071$ & $0.264 \pm 0.087$ & $0.693 \pm 0.003$ & $0.690 \pm 0.003$ & $0.550 \pm 0.066$ & $0.539 \pm 0.067$ \\
& \textsc{BenchAlign}& $\mathbf{0.855 \pm 0.002}$& $\mathbf{0.814 \pm 0.002}$& $\mathbf{0.876 \pm 0.022}$& $\mathbf{0.804 \pm 0.033}$&  $0.863 \pm 0.002$&  $0.862 \pm 0.002$ & $0.894 \pm 0.019$ &  $0.891 \pm 0.019$\\
\bottomrule
\end{tabular}
}
\end{table*}

Table \ref{tab:random_splits_helpsteer_ultrafeedback} reports the ranking performance on all target sets. We observe that \textsc{BenchAlign} produces better aligned benchmarks under random splits than when conditioning on model scale, achieving an increase of around $0.1$ points in $Acc_{pair}$ and $\rho$ in most cases. This comparison suggests that alignment benefits from the greater variability introduced by random sampling into the model pool. The metrics achieved by the remaining baselines provide further evidence of the gains obtained from an unconditioned sampling, as all of them also output rankings of unseen models with higher $Acc_{pair}$ and $\rho$.

However, the results in $GPT2$ provide further insights on generalization. $GPT2$ is the worst performing setting from all experiments and also one of the settings where the correlation obtained by \textsc{BenchAlign} increases the most. \textsc{BenchAlign} correlations in Section \ref{sec:perf_model_size_splits} remain between $0.3$ and $0.5$, while \textsc{TinyBenchmarks} and \textsc{MetaBench} struggle to output significant correlations. In arbitrary model sets, these correlations increase to values between $0.76$ and $0.8$ for \textsc{BenchAlign}, and between $0.2$ and $0.3$ for both \textsc{TinyBenchmarks} and \textsc{MetaBench}. 
As the results of \textsc{TinyBenchmarks} and \textsc{MetaBench} are the equivalent of using the overall accuracy in all benchmarks as a proxy for preferences, these results suggest that benchmark data with small or no correlation with the target preferences might still provide useful information for alignment as long as there is enough variability in the data available during training. 

\subsection{Preference Heterogeneity}\label{ap:pref_heterog}

We evaluate the robustness of our approach to preference heterogeneity by doing an additional set of experiments considering preferences regarding whether a human will judge the responses of a model better than other. This preferences were obtained by training reward models on WebGPT\footnote{\url{https://huggingface.co/datasets/openai/webgpt_comparisons}}, Summarize from Feedback\footnote{\url{https://huggingface.co/datasets/openai/summarize_from_feedback}} and Synthethic Instruct GPT\footnote{\url{Dahoas/synthetic-instruct-gptj-pairwise}} and the base architectures correspond to \cite{He2021DeBERTaV3:Sharing}. All reward models below were trained differently on these 3 datasets, making them suitable to evaluate heterogeneity. Furthermore, we evaluate the responses of all models in Math Precalculus Hard rather than IFEval as finding positive results in Math would show that the results for heterogeneity are not exclusive to the benchmark we used in previous settings. We perform our evaluation on 3 randomly splitted target sets: 2\%, 5\% and 10\%. Specifically, we consider four reward models\footnote{The implementations used are from OpenAssistant and can be found here: \url{https://huggingface.co/OpenAssistant}}:
\begin{itemize} 
\item $RM_{1}^{judge}$: Deberta-v3-base \item $RM_{2}^{judge}$: Deberta-v3-large \item $RM_{3}^{judge}$: Deberta-v3-large-v2 \item $RM_{4}^{judge}$: Electra-Large-Discriminator 
\end{itemize}

\newpage
We observe in Tables \ref{tab:model_size_splits_heterog} and \ref{tab:eval_heterogeneous} that our simple alignment approach outperforms all cases, including the \textsc{Individual} baseline. In the 10\% target, this difference is even stronger, as the difference between accuracies of these two baselines is of $\sim18\%$ in $RM_{1}^{Judge}$, while the difference in the correlations is much larger, reaching approximately 42\%. The \textsc{No Alignment} baseline shows further limitations of benchmarks under these conditions as the correlations are low and inconsistent across preferences. Overall, our results suggest that the alignment can be performed across heterogeneous sets of preferences, even when the benchmarks used don't have a stable correlation with the target preferences.

\begin{table*}[t]
\vspace{0.1in}
\caption{Results under Math Precalculus Hard as evaluation dataset and heterogenous preference sources}
\vspace{0.1in}
\label{tab:eval_heterogeneous}
\centering
\small
\setlength{\tabcolsep}{5pt}
\renewcommand{\arraystretch}{1.05}

\resizebox{\textwidth}{!}{
\begin{tabular}{c c c c c c  |c c c c}
\toprule
\multirow{2}{*}{Target}& \multirow{2}{*}{Method}& \multicolumn{4}{c}{$Acc_{pair}$}& \multicolumn{4}{c}{$\rho$}\\
\cline{3-10}& & $RM_{1}^{Judge}$& $RM_{2}^{Judge}$& $RM_{3}^{Judge}$& $RM_{4}^{Judge}$& $RM_{1}^{Judge}$& $RM_{2}^{Judge}$& $RM_{3}^{Judge}$& $RM_{4}^{Judge}$\\
\midrule

\multirow{7}{*}{2\%}
& \textsc{No Alignment}& $0.578\pm0.001$ & $0.743\pm0.007$& $0.638\pm0.015$& $0.684\pm0.015$ & $0.272\pm0.207$& $0.700\pm0.122$& $0.390\pm0.191$ &$0.511\pm0.174$\\
& \textsc{Random}& $0.606\pm0.016$& $0.760\pm0.014$& $0.650\pm0.020$& $0.681\pm0.015$&  $0.299\pm0.209$&  $0.717\pm0.121$&  $0.431\pm0.199$&  $0.506\pm0.175$\\
& \textsc{Individual}& $0.649\pm0.015$ & $0.773\pm0.031$ & $0.709\pm0.015$& $0.729\pm0.014$ & $0.395\pm0.194$ & $0.742\pm0.112$& $0.548\pm0.167$ & $0.598\pm0.155$\\
& \textsc{Metabench}& $0.581\pm0.011$ & $0.748\pm0.010$ & $0.631\pm0.011$& $0.678\pm0.011$ & $0.244\pm0.199$ & $0.701\pm0.110$ & $0.383\pm0.181$ & $0.503\pm0.159$\\
& \textsc{TinyBenchmarks}& $0.604 \pm 0.016$ & $0.745 \pm 0.014$ & $0.647 \pm 0.015$ & $0.679 \pm 0.015$ & $0.300 \pm 0.193$ & $0.695 \pm 0.111$ & $0.418 \pm 0.175$ & $0.507 \pm 0.158$ \\
& \textsc{BenchAlign}& $\mathbf{0.768\pm0.014}$ & $\mathbf{0.842\pm0.012}$ & $\mathbf{0.820\pm0.012}$ & $\mathbf{0.773\pm0.013}$ & $\mathbf{0.698\pm0.127}$ & $\mathbf{0.855\pm0.065}$ & $\mathbf{0.824\pm0.082}$ & $\mathbf{0.693\pm0.128}$\\
\midrule

\multirow{7}{*}{5\%}
& \textsc{No Alignment} & $0.559\pm0.001$& $0.713\pm0.006$& $0.610\pm0.006$& $0.616\pm0.006$& $0.193\pm0.131$& $0.606\pm0.091$& $0.327\pm0.124$&$0.336\pm0.123$\\
& \textsc{Random} & $0.574\pm0.006$& $0.713\pm0.006$& $0.623\pm0.006$& $0.609\pm0.006$&  $0.209\pm0.131$&  $0.591\pm0.091$&  $0.358\pm0.122$&  $0.309\pm0.125$\\
& \textsc{Individual}& $0.607\pm0.006$& $0.724\pm0.007$& $0.670\pm0.006$& $0.671\pm0.006$& $0.277\pm0.127$& $0.643\pm0.085$& $0.453\pm0.112$&$0.460\pm0.112$\\
& \textsc{Metabench}& $0.552\pm0.004$ & $0.713\pm0.004$ & $0.601\pm0.004$ & $0.612\pm0.004$ & $0.159\pm0.130$& $0.603\pm0.085$& $0.305\pm0.121$& $0.325\pm0.119$\\
& \textsc{TinyBenchmarks}& $0.569 \pm 0.006$ & $0.701 \pm 0.006$ & $0.623 \pm 0.006$ & $0.608 \pm 0.006$ & $0.209 \pm 0.127$ & $0.577 \pm 0.089$ & $0.363 \pm 0.116$ & $0.312 \pm 0.120$ \\
& \textsc{BenchAlign}& $\mathbf{0.770\pm0.005}$ & $\mathbf{0.850\pm0.005}$ & $\mathbf{0.811\pm0.005}$ & $\mathbf{0.803\pm0.005}$ & $\mathbf{0.705\pm0.074}$ & $\mathbf{0.873\pm0.036}$ & $\mathbf{0.806\pm0.052}$ & $\mathbf{0.770\pm0.060}$\\
\midrule

\multirow{7}{*}{10\%} 
& \textsc{No Alignment}& $0.563\pm0.003$& $0.713\pm0.003$& $0.605\pm0.003$& $0.626\pm0.003$& $0.184\pm0.094$& $0.601\pm0.000$& $0.314\pm0.087$&$0.363\pm0.084$\\
& \textsc{Random}& $0.565\pm0.003$& $0.706\pm0.003$& $0.612\pm0.003$& $0.619\pm0.003$&  $0.181\pm0.092$&  $0.576\pm0.066$&  $0.323\pm0.087$&  $0.337\pm0.086$\\
& \textsc{Individual}& $0.622\pm0.003$& $0.728\pm0.003$& $0.682\pm0.003$& $0.664\pm0.003$& $0.327\pm0.087$& $0.622\pm0.061$& $0.492\pm0.075$&$0.448\pm0.078$\\
& \textsc{Metabench}& $0.557\pm0.002$& $0.715\pm0.002$ & $0.598\pm0.002$& $0.625\pm0.002$ & $0.168\pm0.091$ & $0.606\pm0.060$& $0.296\pm0.086$& $0.363\pm0.082$ \\
& \textsc{TinyBenchmarks}& $0.566 \pm 0.003$ & $0.701 \pm 0.003$ & $0.613 \pm 0.003$ & $0.619 \pm 0.003$ & $0.195 \pm 0.090$ & $0.576 \pm 0.063$ & $0.335 \pm 0.083$ & $0.348 \pm 0.083$ \\
& \textsc{BenchAlign}& $\mathbf{0.782\pm0.003}$ & $\mathbf{0.850\pm0.002}$ & $\mathbf{0.814\pm0.002}$ & $\mathbf{0.808\pm0.003}$ & $\mathbf{0.746\pm0.045}$ & $\mathbf{0.875\pm0.024}$ & $\mathbf{0.814\pm0.034}$ & $\mathbf{0.796\pm0.037}$\\
\bottomrule
\end{tabular}
}
\end{table*}

\begin{table*}[t]
\centering
\small
\setlength{\tabcolsep}{5pt}
\renewcommand{\arraystretch}{1.05}

\resizebox{\textwidth}{!}{
\begin{tabular}{c c c c c c  |c c c c}
\toprule
\multirow[c]{2}{*}{\makecell{\textbf{Target}\\\textbf{Models}}} & \multirow[c]{2}{*}{\textbf{Method}} & \multicolumn{4}{c}{$Acc_{pair}$} & \multicolumn{4}{c}{$\rho$} \\
\cline{3-10}
& & $RM_{1}^{Judge}$ & $RM_{2}^{Judge}$ & $RM_{3}^{Judge}$ & $RM_{4}^{Judge}$ & $RM_{1}^{Judge}$ & $RM_{2}^{Judge}$ & $RM_{3}^{Judge}$ & $RM_{4}^{Judge}$ \\
\midrule

\multirow{4}{*}{70B+} & \textsc{No Alignment} & $0.482\pm0.010$ & $0.607\pm0.010$ & $0.509\pm0.010$ & $0.541\pm0.010$ & $-0.051\pm0.166$ & $0.318\pm0.141$ & $0.030\pm0.164$ & $0.163\pm0.156$ \\
 & \textsc{Random} & $0.555\pm0.010$ & $0.583\pm0.010$ & $\mathbf{0.632\pm0.009}$ & $0.541\pm0.010$ & $0.123\pm0.159$ & $0.236\pm0.150$ & $0.322\pm0.140$ & $0.123\pm0.159$ \\
 & \textsc{Individual} & $0.583\pm0.010$ & $0.668\pm0.009$ & $\mathbf{0.632\pm0.009}$ & $0.647\pm0.009$ & $0.198\pm0.153$ & $0.465\pm0.120$ & $0.322\pm0.140$ & $0.372\pm0.134$ \\
 & \textsc{Metabench} & $0.481\pm0.007$ & $0.611\pm0.007$ & $0.498\pm0.007$ & $0.546\pm0.007$ & $-0.059\pm0.164$ & $0.331\pm0.147$ & $0.000\pm0.165$ & $0.177\pm0.160$ \\
 & \textsc{TinyBenchmarks} & $0.457\pm0.007$ & $0.585\pm0.007$ & $0.511\pm0.001$ & $0.542\pm0.007$ & $-0.113\pm0.163$ & $0.267\pm0.153$ & $0.032\pm0.067$ & $0.142\pm0.162$ \\
 & \textsc{BenchAlign} & $\mathbf{0.620\pm0.010}$ & $\mathbf{0.735\pm0.009}$ & $0.617\pm0.010$ & $\mathbf{0.712\pm0.009}$ & $\mathbf{0.340\pm0.138}$ & $\mathbf{0.647\pm0.087}$ & $\mathbf{0.338\pm0.138}$ & $\mathbf{0.603\pm0.095}$ \\
\midrule
\multirow{4}{*}{30B+} & \textsc{No Alignment} & $0.433\pm0.005$ & $0.619\pm0.005$ & $0.482\pm0.005$ & $0.569\pm0.005$ & $-0.186\pm0.121$ & $0.351\pm0.103$ & $-0.048\pm0.123$ & $0.229\pm0.113$ \\
 & \textsc{Random} & $0.467\pm0.005$ & $0.605\pm0.005$ & $0.630\pm0.005$ & $0.544\pm0.005$ & $-0.112\pm0.123$ & $0.302\pm0.108$ & $0.335\pm0.105$ & $0.133\pm0.119$ \\
 & \textsc{Individual} & $0.549\pm0.005$ & $0.666\pm0.005$ & $0.630\pm0.005$ & $0.622\pm0.005$ & $0.114\pm0.120$ & $0.466\pm0.091$ & $0.335\pm0.105$ & $0.340\pm0.104$ \\
 & \textsc{Metabench} & $0.406\pm0.004$ & $0.601\pm0.004$ & $0.449\pm0.004$ & $0.554\pm0.004$ & $-0.268\pm0.114$ & $0.294\pm0.112$ & $-0.141\pm0.120$ & $0.177\pm0.119$ \\
 & \textsc{TinyBenchmarks} & $0.432\pm0.004$ & $0.606\pm0.004$ & $0.491\pm0.004$ & $0.561\pm0.004$ & $-0.179\pm0.119$ & $0.315\pm0.111$ & $-0.028\pm0.123$ & $0.194\pm0.118$ \\
 & \textsc{BenchAlign} & $\mathbf{0.621\pm0.005}$ & $\mathbf{0.737\pm0.005}$ & $\mathbf{0.677\pm0.005}$ & $\mathbf{0.721\pm0.005}$ & $\mathbf{0.348\pm0.104}$ & $\mathbf{0.645\pm0.067}$ & $\mathbf{0.500\pm0.087}$ & $\mathbf{0.606\pm0.072}$ \\
\midrule
\multirow{4}{*}{13B+} & \textsc{No Alignment} & $0.426\pm0.002$ & $0.603\pm0.002$ & $0.495\pm0.002$ & $0.529\pm0.002$ & $-0.215\pm0.065$ & $0.298\pm0.060$ & $-0.016\pm0.068$ & $0.081\pm0.067$ \\
 & \textsc{Random} & $0.454\pm0.002$ & $0.599\pm0.002$ & $0.581\pm0.002$ & $0.509\pm0.002$ & $-0.156\pm0.067$ & $0.277\pm0.061$ & $0.224\pm0.063$ & $0.030\pm0.067$ \\
 & \textsc{Individual} & $0.587\pm0.002$ & $0.660\pm0.002$ & $\mathbf{0.652\pm0.002}$ & $0.612\pm0.002$ & $0.223\pm0.063$ & $0.451\pm0.052$ & $0.408\pm0.055$ & $0.319\pm0.059$ \\
 & \textsc{Metabench} & $0.432\pm0.001$ & $0.621\pm0.001$ & $0.492\pm0.001$ & $0.540\pm0.001$ & $-0.194\pm0.065$ & $0.343\pm0.060$ & $-0.025\pm0.067$ & $0.114\pm0.067$ \\
 & \textsc{TinyBenchmarks} & $0.440\pm0.001$ & $0.579\pm0.001$ & $0.508\pm0.001$ & $0.520\pm0.001$ & $-0.175\pm0.065$ & $0.232\pm0.064$ & $0.025\pm0.067$ & $0.060\pm0.067$ \\
 & \textsc{BenchAlign} & $\mathbf{0.591\pm0.002}$ & $\mathbf{0.709\pm0.001}$ & $0.646\pm0.002$ & $\mathbf{0.690\pm0.002}$ & $\mathbf{0.261\pm0.062}$ & $\mathbf{0.571\pm0.044}$ & $\mathbf{0.417\pm0.054}$ & $\mathbf{0.520\pm0.048}$ \\
\bottomrule
\end{tabular}
}
\caption{Alignment with heterogeneous preferences under model scale splits}
\label{tab:model_size_splits_heterog}
\end{table*}

\section{Minimizing Data Requirements across Multiple Preferences and Targets}\label{ap:opt_data_reqs}

\subsection{Number of Models}\label{ap:opt_num_models}

We consider the best smallest subsets that enable alignment from the ones with a correlation or accuracy whose confidence bands overlap those of using all training models. Table \ref{tab:best_model_subsets_summary} shows a summary of these subsets for all experiments corresponding to Section \ref{sec:perf_model_size_splits}. For all cases but $\text{GPT2}$, the windows are slightly above the average and median number of parameters, but none of the cases are close to the maximum number of parameters from their corresponding training set (13B, 30B or 70B). Surprisingly, these results indicate that the model size is relevant, but not crucial for determining the best subset of models. In the following subsections, we show the figures corresponding to each set of experiments.

\begin{table}
    \centering
    \vspace{0.1in}
    \caption{Number of Parameters for Best Subset of Models}
    \vspace{0.1in}
    \resizebox{\columnwidth}{!}{%
    \begin{tabular}{cccccc}\toprule
         RM&  Target&  Number of Models&  Average Size of Best Subset&  Average Size - All Subsets&Median Size - All Subsets\\\midrule
         $\text{ArmoRM}$ (Helpful)&  13B+&  20&  8.030&  6.361&7.616\\
         $\text{ArmoRM}$ (Helpful)&  30B+&  10&  14.770&  7.918&8.000\\
         $\text{ArmoRM}$ (Helpful)&  70B+&  10&  14.770&  8.320&8.008\\
         $\text{GPT2}$&  13B+&  400&  7.823&  6.503&7.626\\
         $\text{GPT2}$&  30B+&  10&  0.385&  7.918&8.000\\
         $\text{GPT2}$&  70B+&  10&  1.257&  8.320&8.008\\
         $\text{ArmoRM}$ (Honest)&  13B+&  20&  8.030&  6.361&7.616\\
         $\text{ArmoRM}$ (Honest)&  30B+&  10&  14.766&  7.918&8.000\\
         $\text{ArmoRM}$ (Honest)&  70B+&  10&  12.100&  8.320&8.008\\
 $\text{DPA}$& 13B+& 20& 8.030& 6.361&7.616\\
 $\text{DPA}$& 30B+& 10& 14.766& 7.918&8.000\\
 $\text{DPA}$& 70B+& 10& 7.616& 8.320&8.008\\ \bottomrule
    \end{tabular}
    }
    \label{tab:best_model_subsets_summary}
\end{table}

\subsubsection{$\text{ArmoRM}$ (Helpful)}

\xhdr{Target: 13B+} Figures \ref{fig:bench_align_opt_num_models_13b_helpful_rm1} and \ref{fig:bench_align_opt_num_models_13b_helpful_rm1} represent the results of all subset of models and the best subset by number of models, respectively.

\begin{figure*}[t]
    \centering
    \begin{subfigure}[b]{0.49\linewidth}
        \centering
        \includegraphics[width=\linewidth]{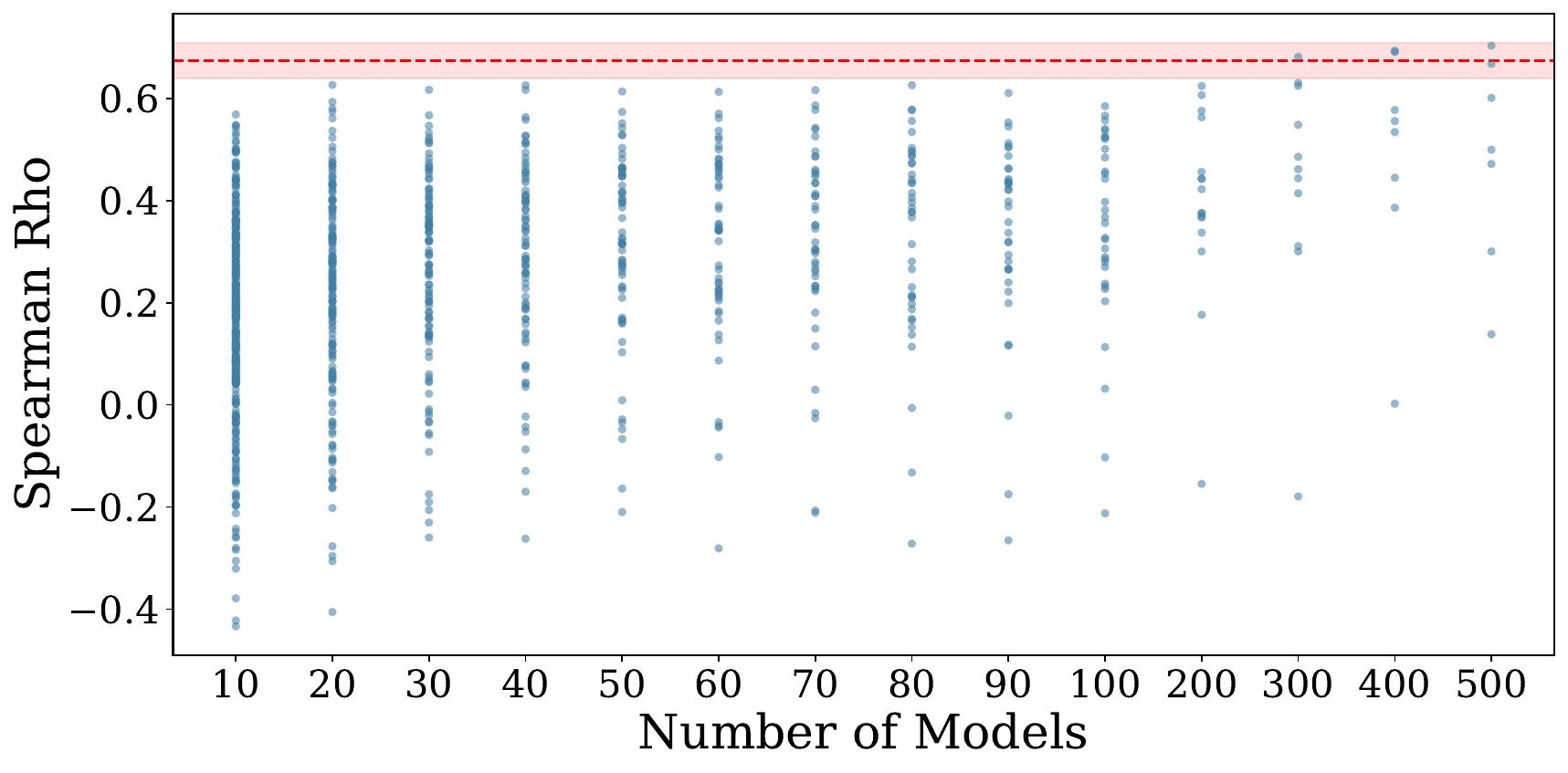}
        \caption{Spearman Rho ($\rho$)}
    \end{subfigure}
    \hfill
    \begin{subfigure}[b]{0.49\linewidth}
        \centering
        \includegraphics[width=\linewidth]{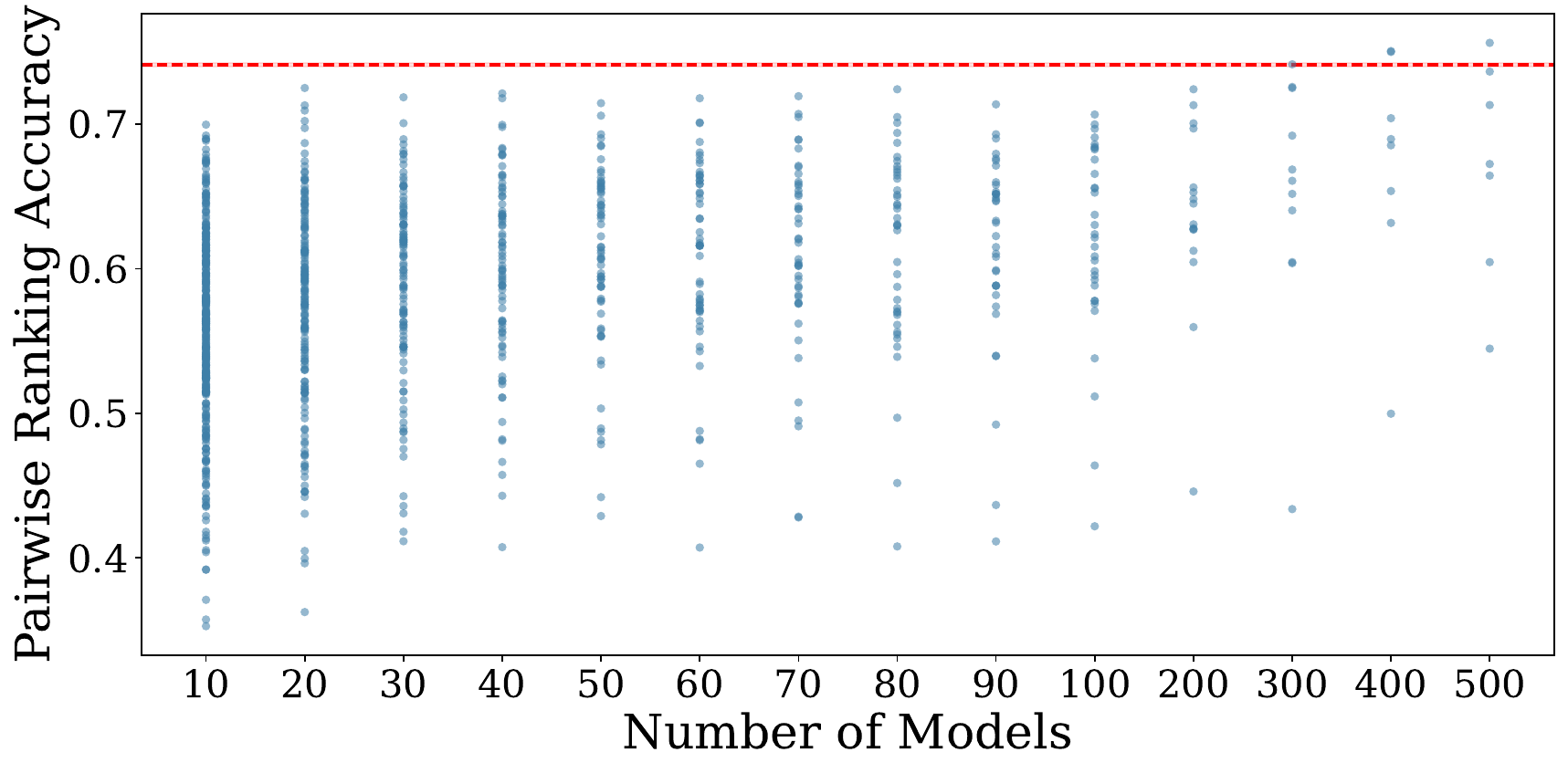}
        \caption{Pairwise Ranking Accuracy ($Acc_{pair}$)}
    \end{subfigure}
    \caption{$\text{ArmoRM}$ (Helpful) - Target 13B+. Aligned benchmarks created with different numbers of models. In (a), each point shows the Spearman Rho for one subset of models. In (b), each point shows the Pairwise Ranking Accuracy ($Acc_{pair}$).  Error bars obtained using the sample size from the target set. The dotted line represents the results of alignment using all models.}
    \label{fig:bench_align_opt_num_models_13b_helpful_rm1}
\end{figure*}

\begin{figure*}[t]
    \centering
    \begin{subfigure}[b]{0.49\linewidth}
        \centering
        \includegraphics[width=\linewidth]{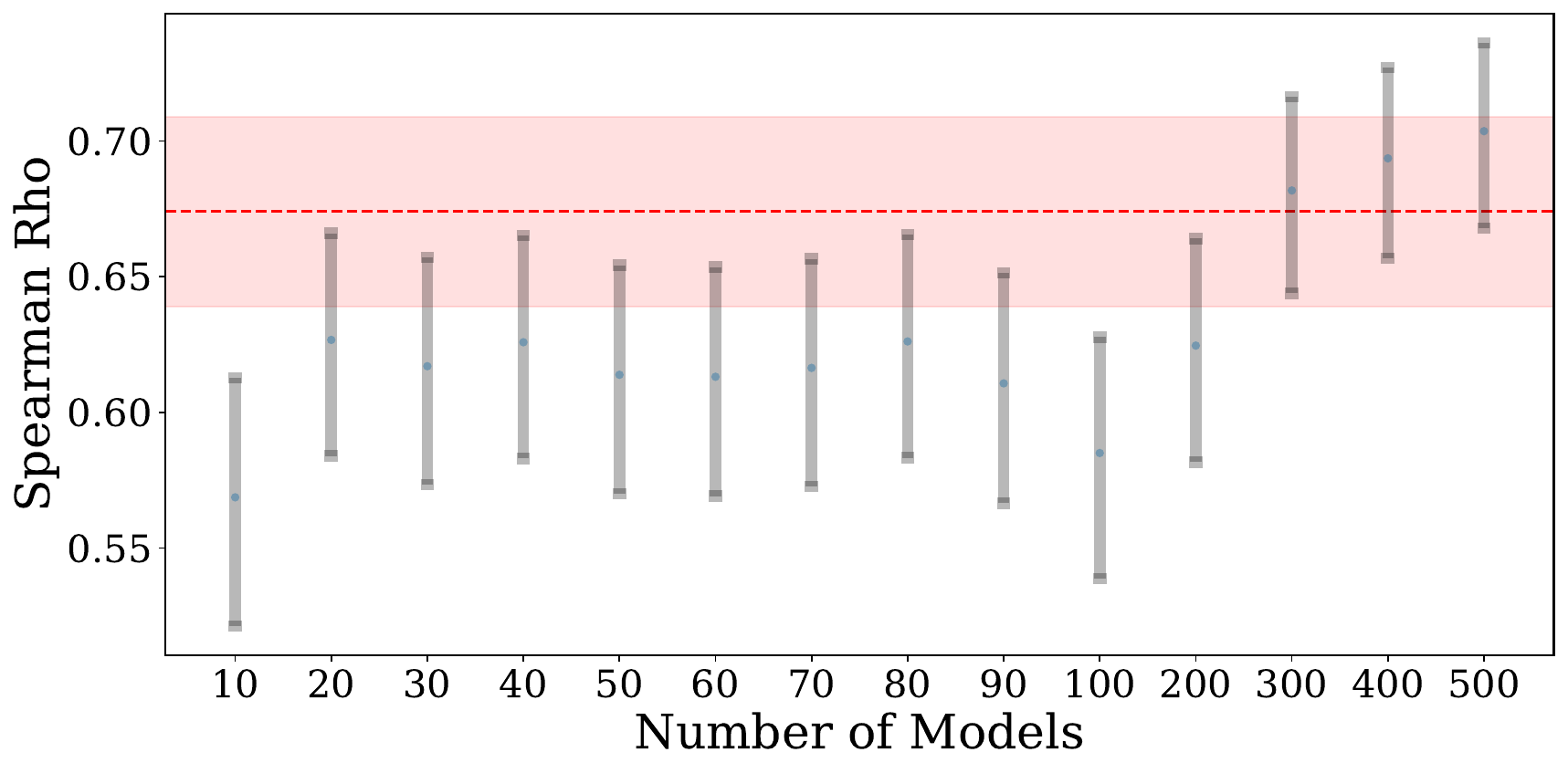}
        \caption{Spearman Rho ($\rho$)}
    \end{subfigure}
    \hfill
    \begin{subfigure}[b]{0.49\linewidth}
        \centering
        \includegraphics[width=\linewidth]{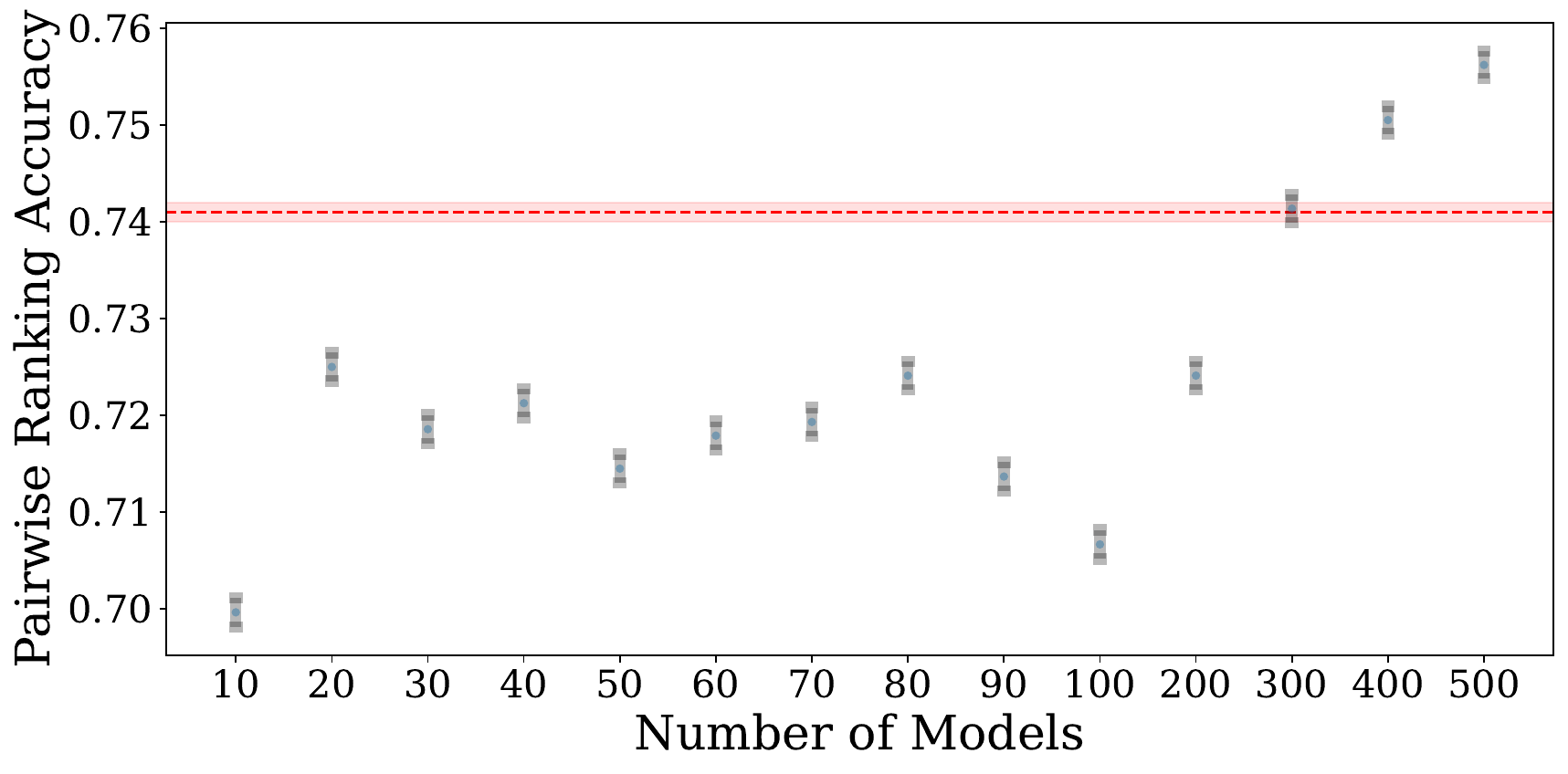}
        \caption{Pairwise Ranking Accuracy ($Acc_{pair}$)}
    \end{subfigure}
    \caption{$\text{ArmoRM}$ (Helpful) - Target 13B+. Aligned benchmarks created with different numbers of models. In (a), each point reports the Spearman Rho ($\rho$) of the best subset of models. In (b), each point shows the Pairwise Ranking Accuracy ($Acc_{pair}$). Error bars obtained using the sample size from the target set. The dotted line represents the results of alignment using all models.}
    \label{fig:bench_align_opt_num_models_13b_helpful_rm1}
\end{figure*}

\xhdr{Target: 30B+} Figures \ref{fig:bench_align_opt_num_models_30b_helpful_rm1} and \ref{fig:bench_align_opt_num_models_30b_helpful_rm1} represent the results of all subset of models and the best subset by number of models, respectively.

\begin{figure*}[t]
    \centering
    \begin{subfigure}[b]{0.49\linewidth}
        \centering
        \includegraphics[width=\linewidth]{images/model_window_exps/armo-helpsteer-helpfulness/30b_70b/pdfs/armo-helpsteer-helpfulness_30b_70b_rho_vs_window.pdf}
        \caption{Spearman Rho ($\rho$)}
    \end{subfigure}
    \hfill
    \begin{subfigure}[b]{0.49\linewidth}
        \centering
        \includegraphics[width=\linewidth]{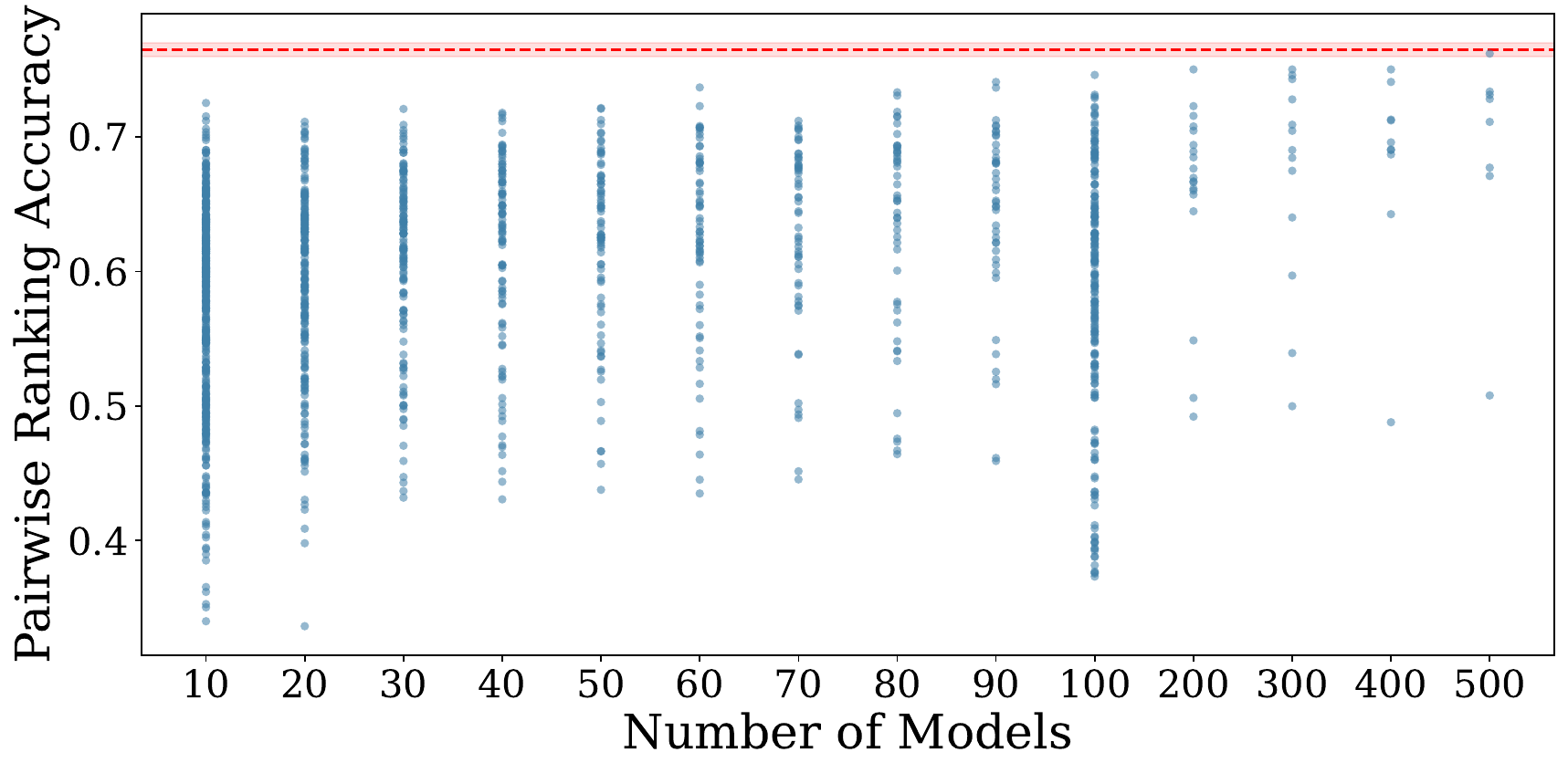}
        \caption{Pairwise Ranking Accuracy ($Acc_{pair}$)}
    \end{subfigure}
    \caption{$\text{ArmoRM}$ (Helpful) - Target 30B+. Aligned benchmarks created with different numbers of models. In (a), each point shows the Spearman Rho for one subset of models. In (b), each point shows the Pairwise Ranking Accuracy ($Acc_{pair}$).  Error bars obtained using the sample size from the target set. The dotted line represents the results of alignment using all models.}
    \label{fig:bench_align_opt_num_models_30b_helpful_rm1}
\end{figure*}

\begin{figure*}[t]
    \centering
    \begin{subfigure}[b]{0.49\linewidth}
        \centering
        \includegraphics[width=\linewidth]{images/model_window_best_exps/armo-helpsteer-helpfulness/30b_70b/pdfs/armo-helpsteer-helpfulness_30b_70b_rho_best_vs_window.pdf}
        \caption{Spearman Rho ($\rho$)}
    \end{subfigure}
    \hfill
    \begin{subfigure}[b]{0.49\linewidth}
        \centering
        \includegraphics[width=\linewidth]{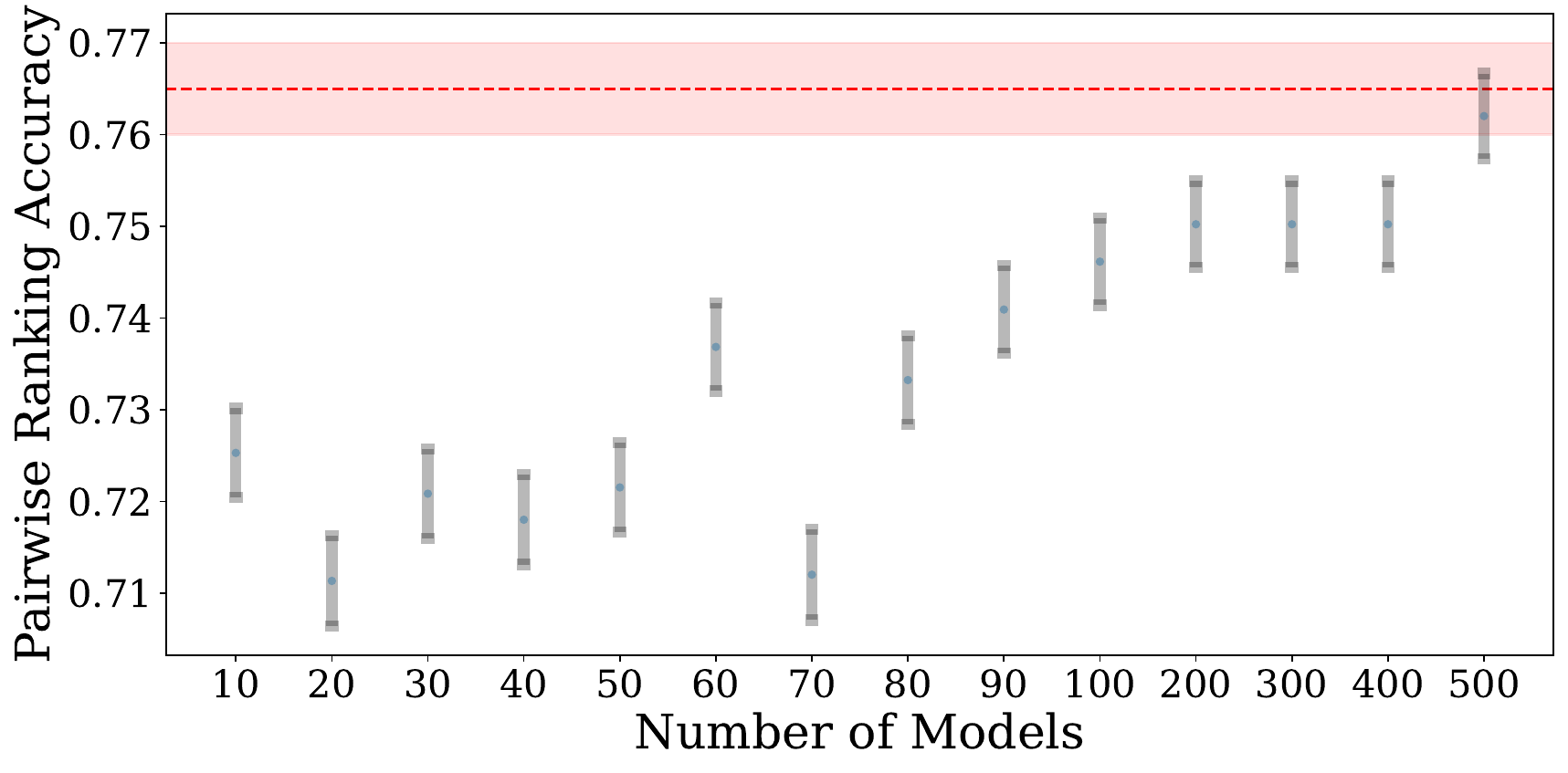}
        \caption{Pairwise Ranking Accuracy ($Acc_{pair}$)}
    \end{subfigure}
    \caption{$\text{ArmoRM}$ (Helpful) - Target 30B+. Aligned benchmarks created with different numbers of models. In (a), each point reports the Spearman Rho ($\rho$) of the best subset of models. In (b), each point shows the Pairwise Ranking Accuracy ($Acc_{pair}$). Error bars obtained using the sample size from the target set. The dotted line represents the results of alignment using all models.}
    \label{fig:bench_align_opt_num_models_30b_helpful_rm1}
\end{figure*}

\xhdr{Target: 70B+} Figures \ref{fig:bench_align_opt_num_models_70b_helpful_rm1} and \ref{fig:bench_align_opt_num_models_70b_helpful_rm1} represent the results of all subset of models and the best subset by number of models, respectively.

\begin{figure*}[t]
    \centering
    \begin{subfigure}[b]{0.49\linewidth}
        \centering
        \includegraphics[width=\linewidth]{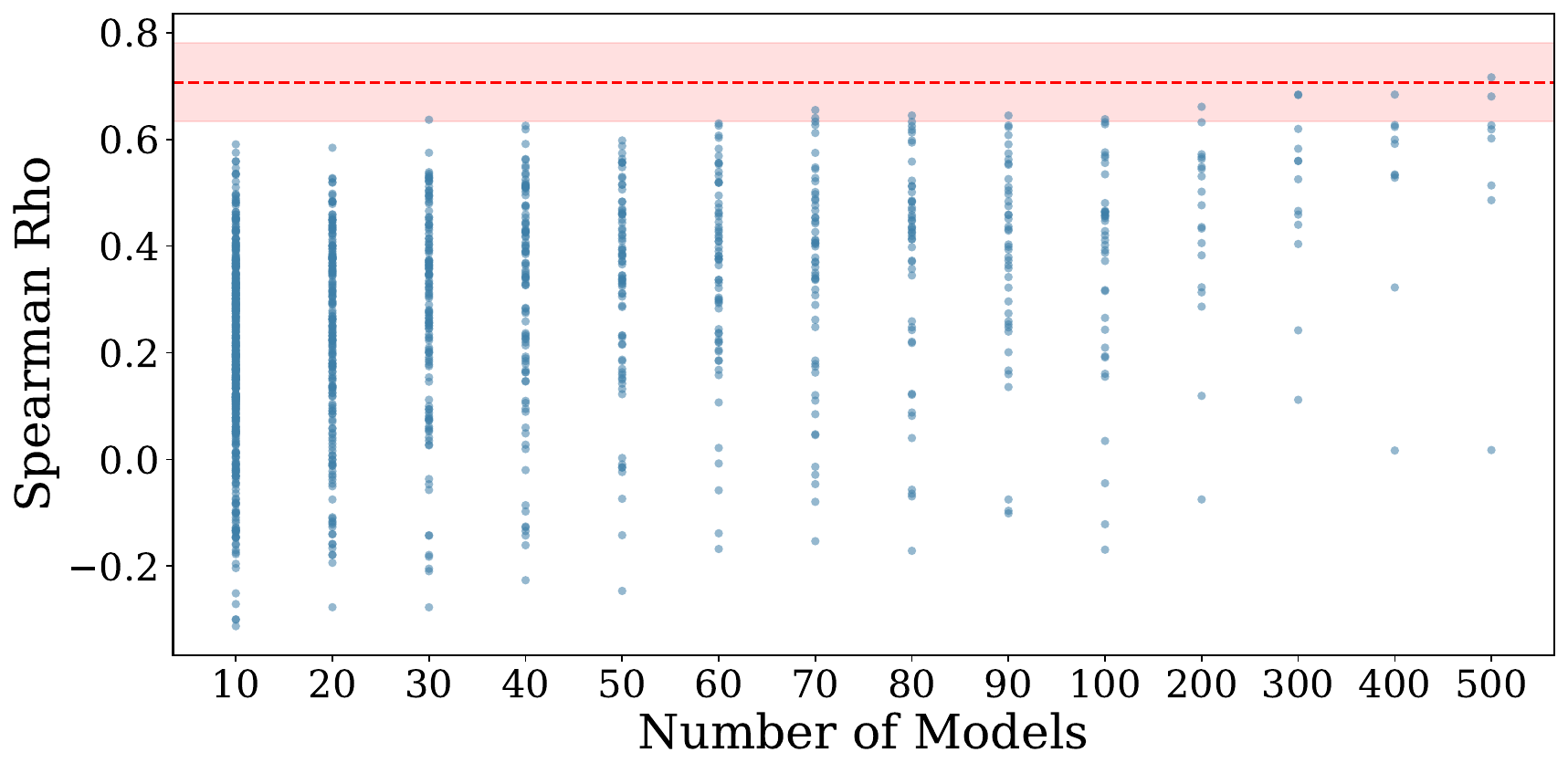}
        \caption{Spearman Rho ($\rho$)}
    \end{subfigure}
    \hfill
    \begin{subfigure}[b]{0.49\linewidth}
        \centering
        \includegraphics[width=\linewidth]{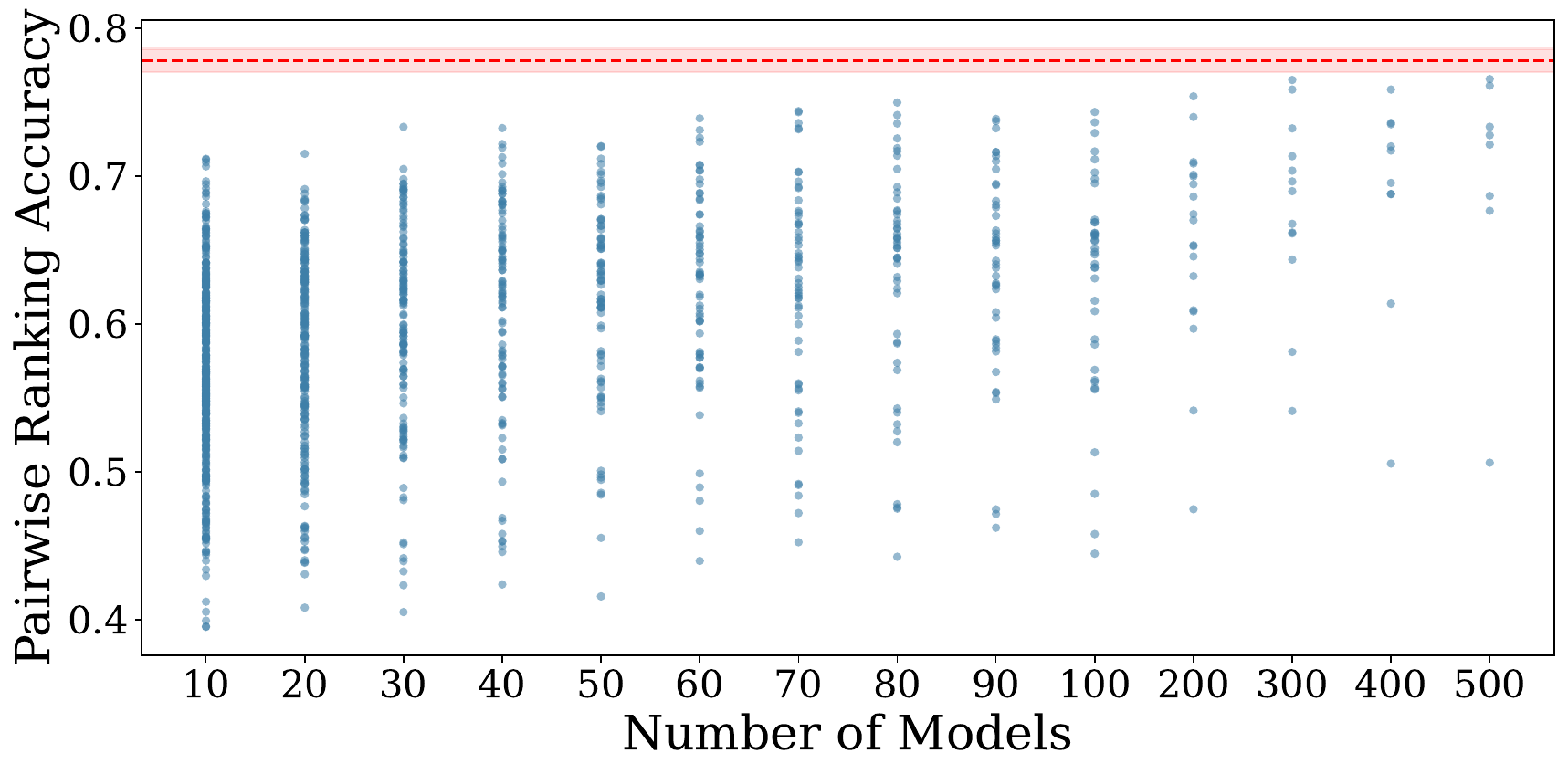}
        \caption{Pairwise Ranking Accuracy ($Acc_{pair}$)}
    \end{subfigure}
    \caption{$\text{ArmoRM}$ (Helpful) - Target 70B+. Aligned benchmarks created with different numbers of models. In (a), each point shows the Spearman Rho for one subset of models. In (b), each point shows the Pairwise Ranking Accuracy ($Acc_{pair}$).  Error bars obtained using the sample size from the target set. The dotted line represents the results of alignment using all models.}
    \label{fig:bench_align_opt_num_models_70b_helpful_rm1}
\end{figure*}

\begin{figure*}[t]
    \centering
    \begin{subfigure}[b]{0.49\linewidth}
        \centering
        \includegraphics[width=\linewidth]{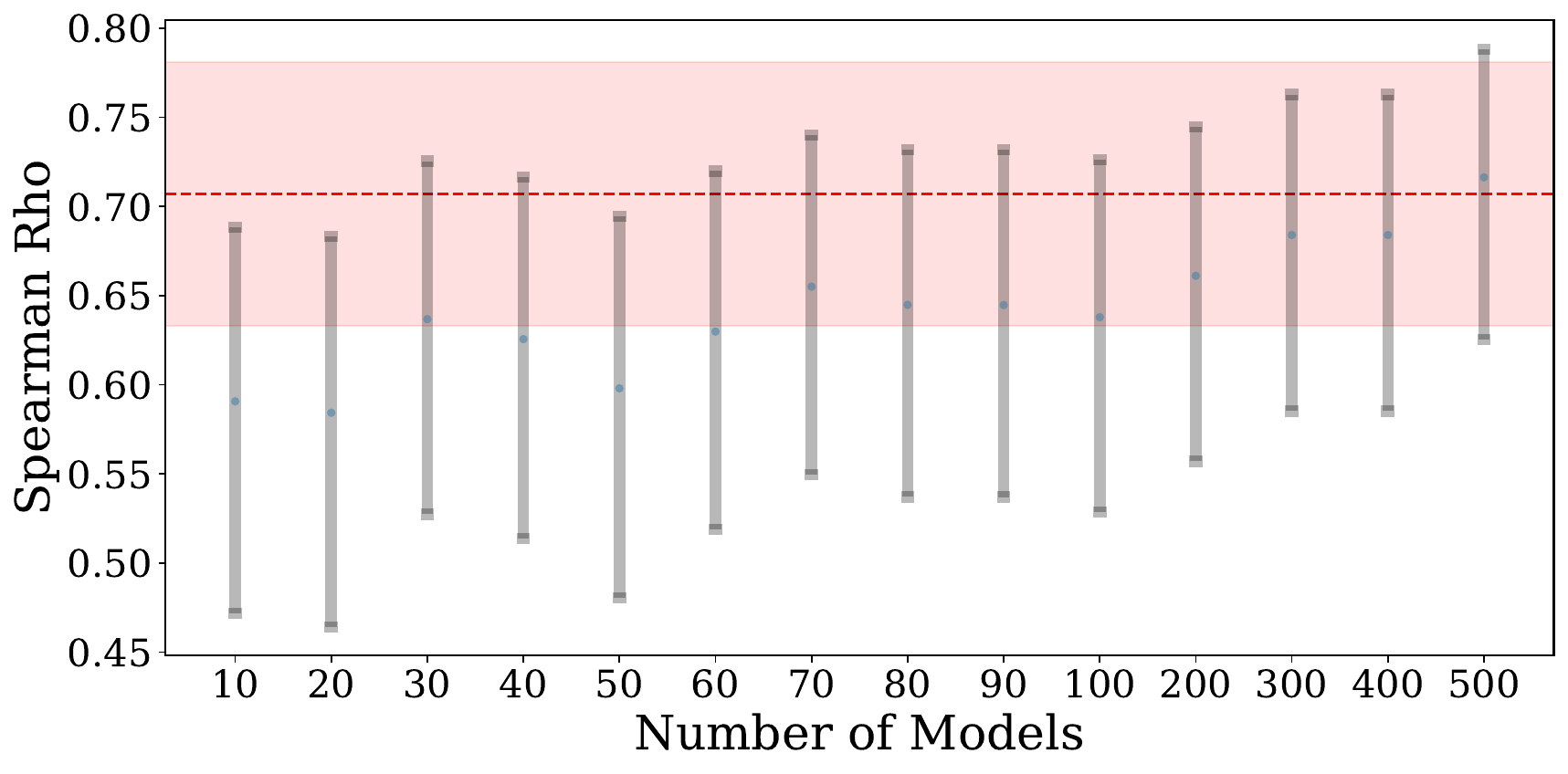}
        \caption{Spearman Rho ($\rho$)}
    \end{subfigure}
    \hfill
    \begin{subfigure}[b]{0.49\linewidth}
        \centering
        \includegraphics[width=\linewidth]{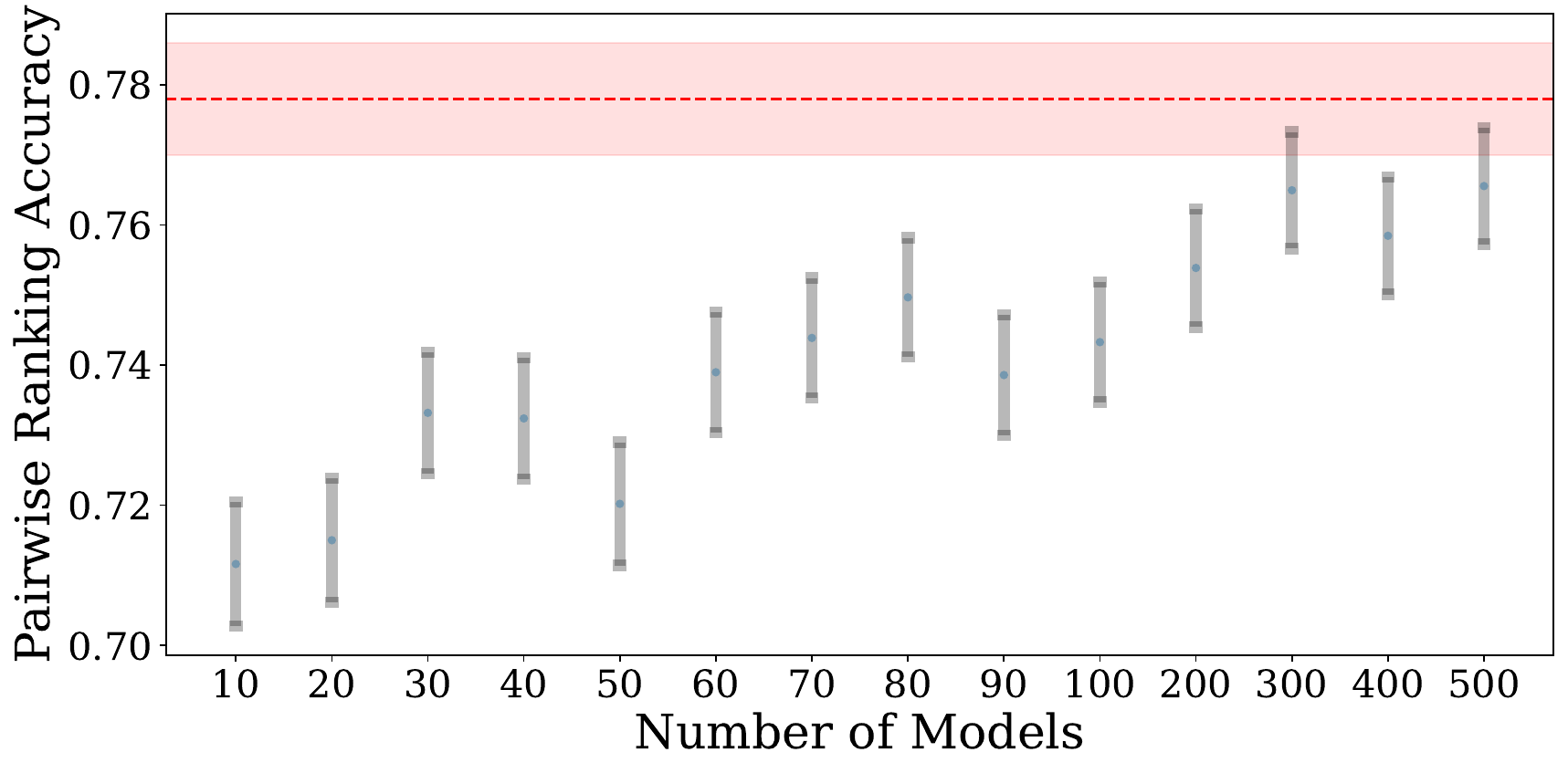}
        \caption{Pairwise Ranking Accuracy ($Acc_{pair}$)}
    \end{subfigure}
    \caption{$\text{ArmoRM}$ (Helpful) - Target 70B+. Aligned benchmarks created with different numbers of models. In (a), each point reports the Spearman Rho ($\rho$) of the best subset of models. In (b), each point shows the Pairwise Ranking Accuracy ($Acc_{pair}$). Error bars obtained using the sample size from the target set. The dotted line represents the results of alignment using all models.}
    \label{fig:bench_align_opt_num_models_70b_helpful_rm1}
\end{figure*}

\clearpage
\subsubsection{$\text{GPT2}$}

\xhdr{Target: 13B+}. Figures \ref{fig:bench_align_opt_num_models_13b_helpful_rm2} and \ref{fig:bench_align_opt_num_models_13b_helpful_rm2} represent the results of all subset of models and the best subset by number of models, respectively.

\begin{figure*}[t]
    \centering
    \begin{subfigure}[b]{0.49\linewidth}
        \centering
        \includegraphics[width=\linewidth]{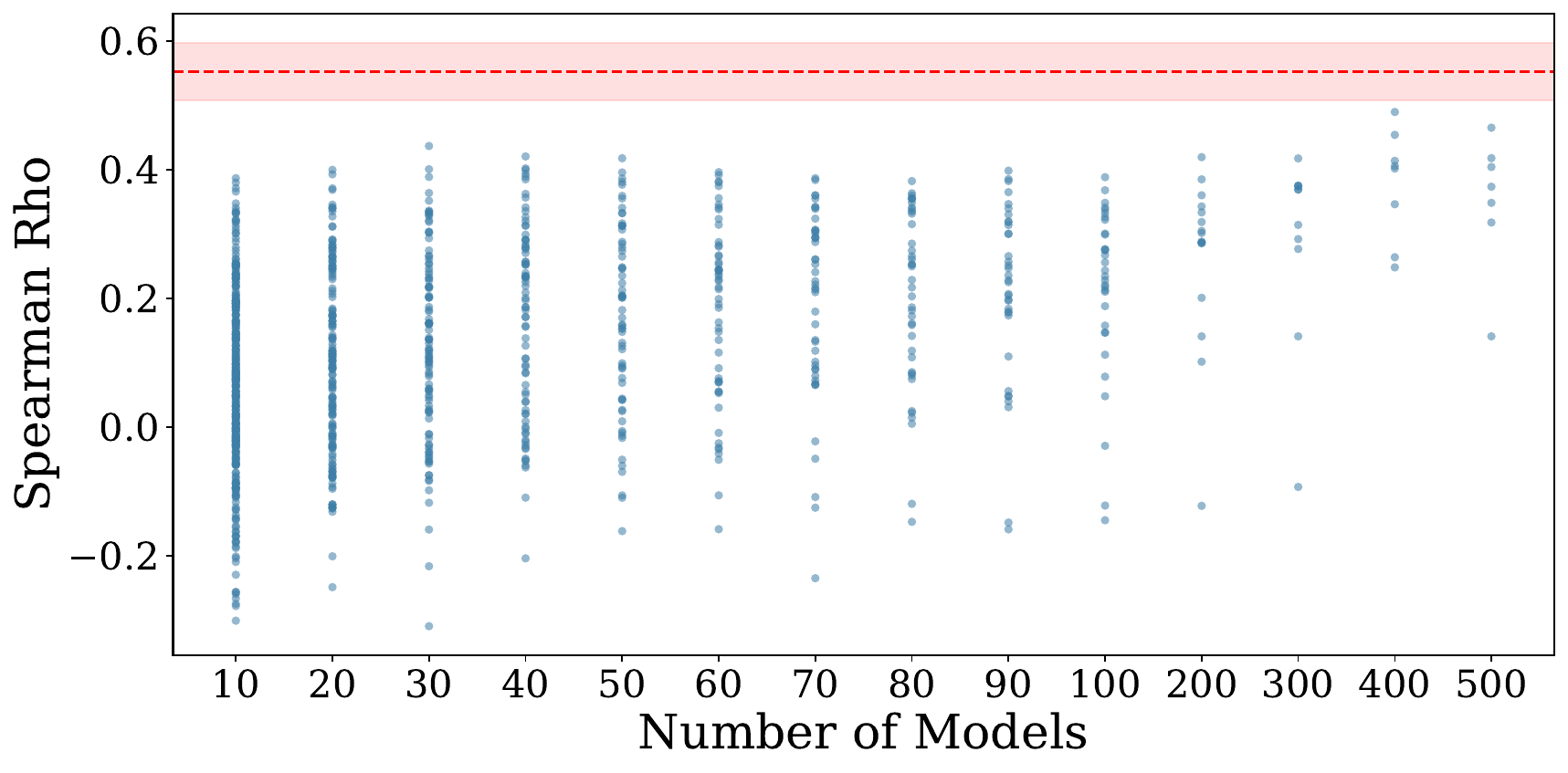}
        \caption{Spearman Rho ($\rho$)}
    \end{subfigure}
    \hfill
    \begin{subfigure}[b]{0.49\linewidth}
        \centering
        \includegraphics[width=\linewidth]{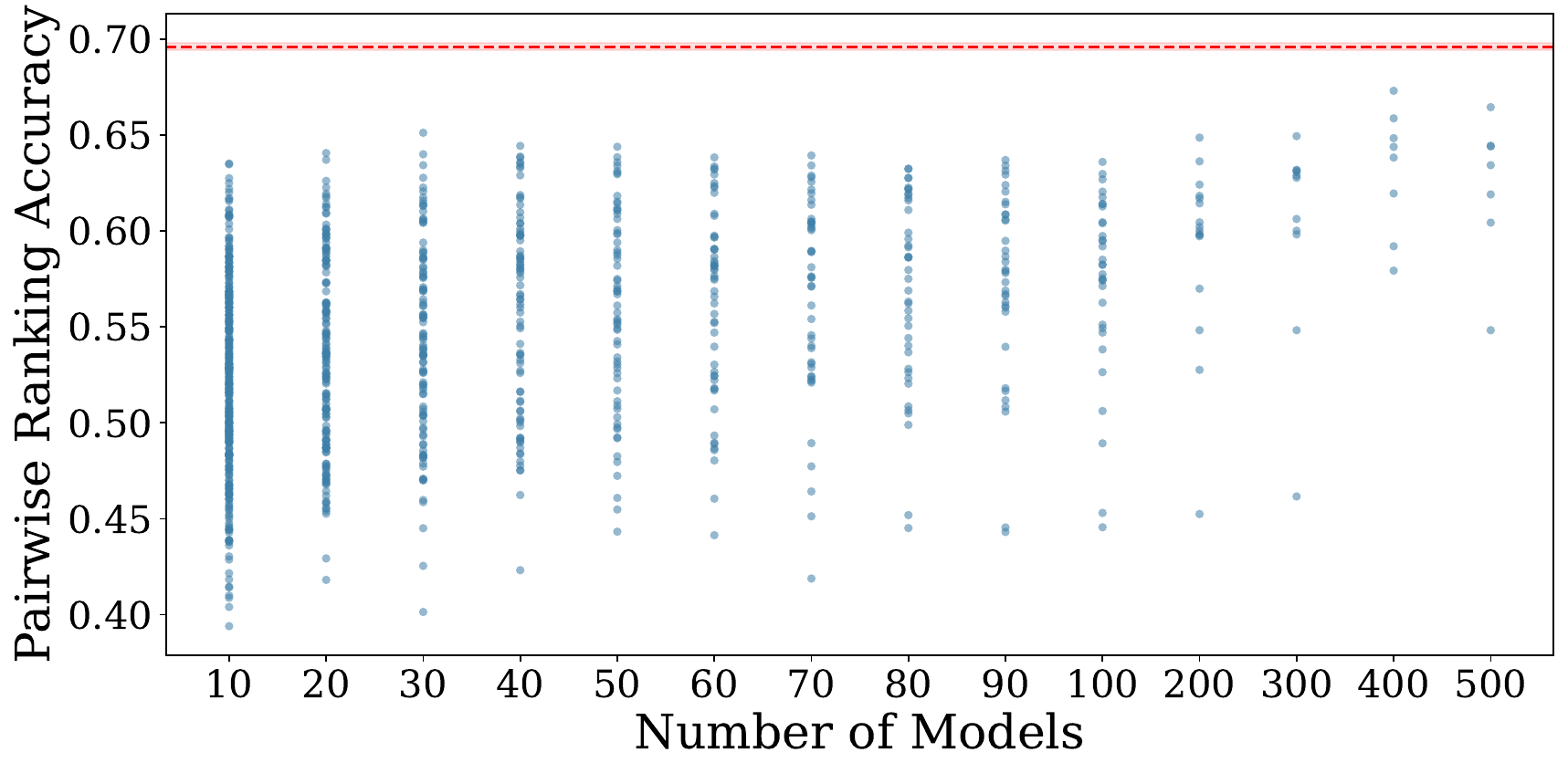}
        \caption{Pairwise Ranking Accuracy ($Acc_{pair}$)}
    \end{subfigure}
    \caption{$\text{GPT2}$ - Target 13B+. Aligned benchmarks created with different numbers of models. In (a), each point shows the Spearman Rho for one subset of models. In (b), each point shows the Pairwise Ranking Accuracy ($Acc_{pair}$).  Error bars obtained using the sample size from the target set. The dotted line represents the results of alignment using all models.}
    \label{fig:bench_align_opt_num_models_13b_helpful_rm2}
\end{figure*}

\begin{figure*}[t]
    \centering
    \begin{subfigure}[b]{0.49\linewidth}
        \centering
        \includegraphics[width=\linewidth]{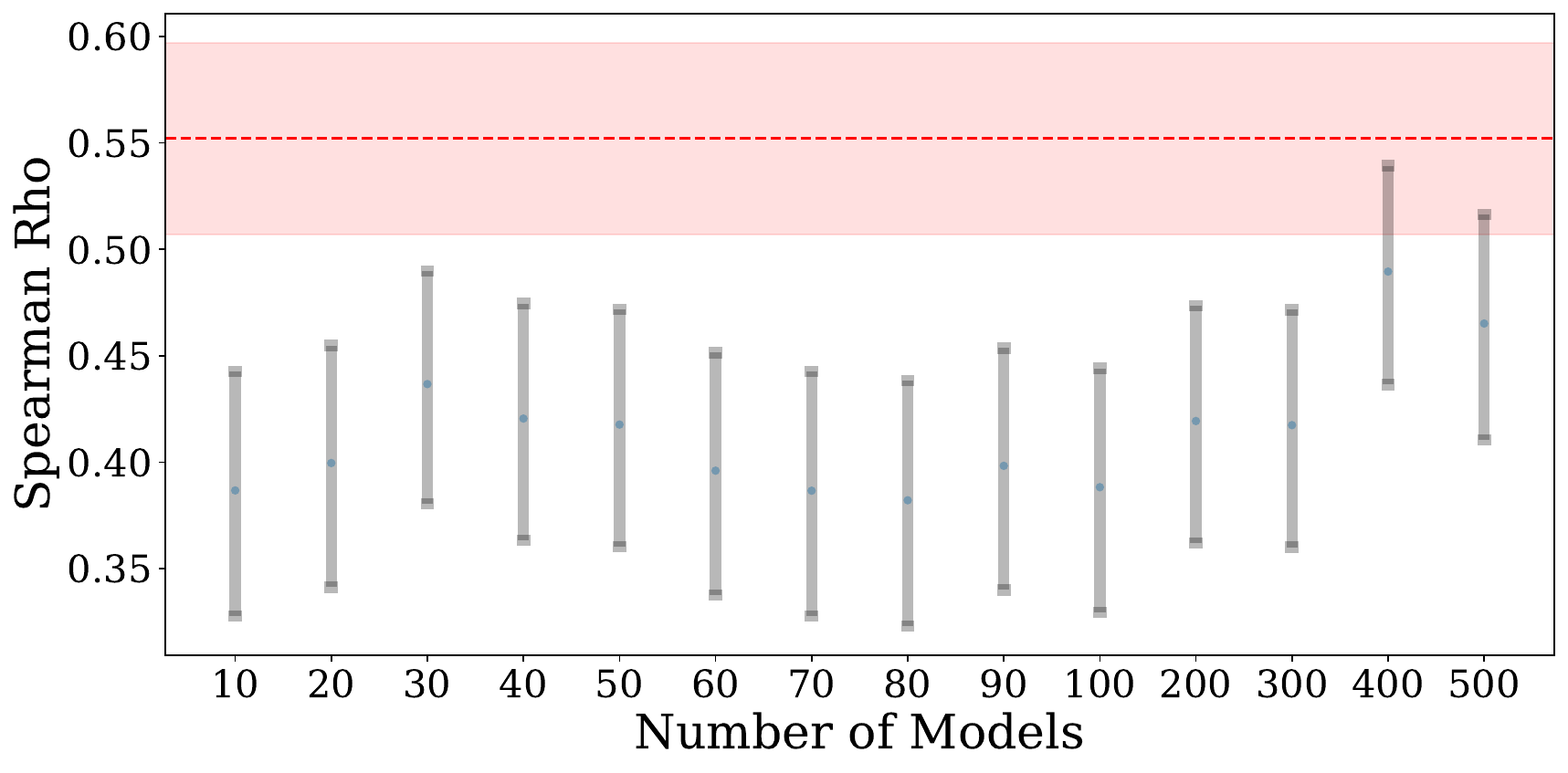}
        \caption{Spearman Rho ($\rho$)}
    \end{subfigure}
    \hfill
    \begin{subfigure}[b]{0.49\linewidth}
        \centering
        \includegraphics[width=\linewidth]{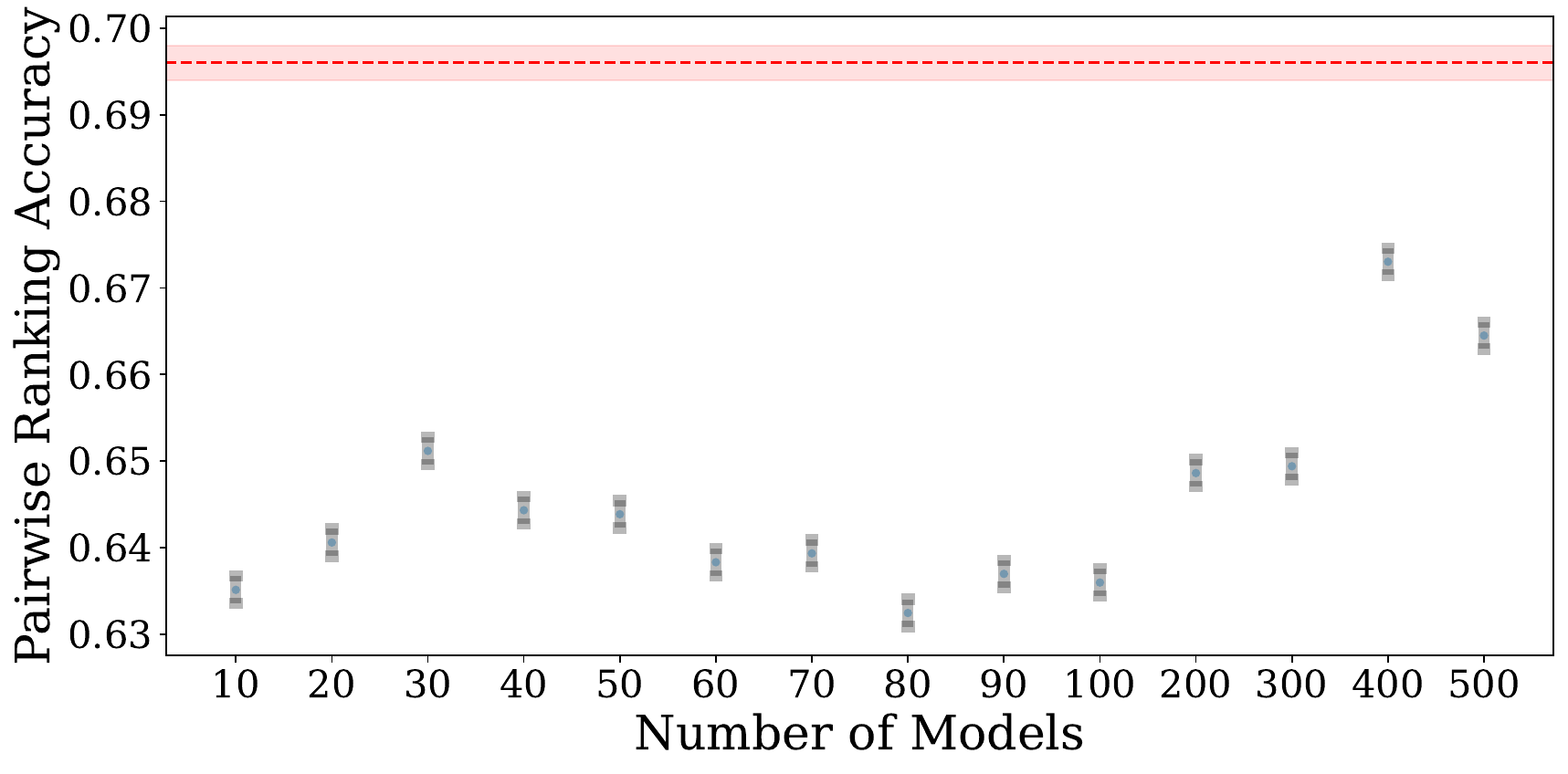}
        \caption{Pairwise Ranking Accuracy ($Acc_{pair}$)}
    \end{subfigure}
    \caption{$\text{GPT2}$ - Target 13B+. Aligned benchmarks created with different numbers of models. In (a), each point reports the Spearman Rho ($\rho$) of the best subset of models. In (b), each point shows the Pairwise Ranking Accuracy ($Acc_{pair}$). Error bars obtained using the sample size from the target set. The dotted line represents the results of alignment using all models.}
    \label{fig:bench_align_opt_num_models_13b_helpful_rm2}
\end{figure*}

\xhdr{Target: 30B+}. Figures \ref{fig:bench_align_opt_num_models_30b_helpful_rm2} and \ref{fig:bench_align_opt_num_models_30b_helpful_rm2} represent the results of all subset of models and the best subset by number of models, respectively.

\begin{figure*}[t]
    \centering
    \begin{subfigure}[b]{0.49\linewidth}
        \centering
        \includegraphics[width=\linewidth]{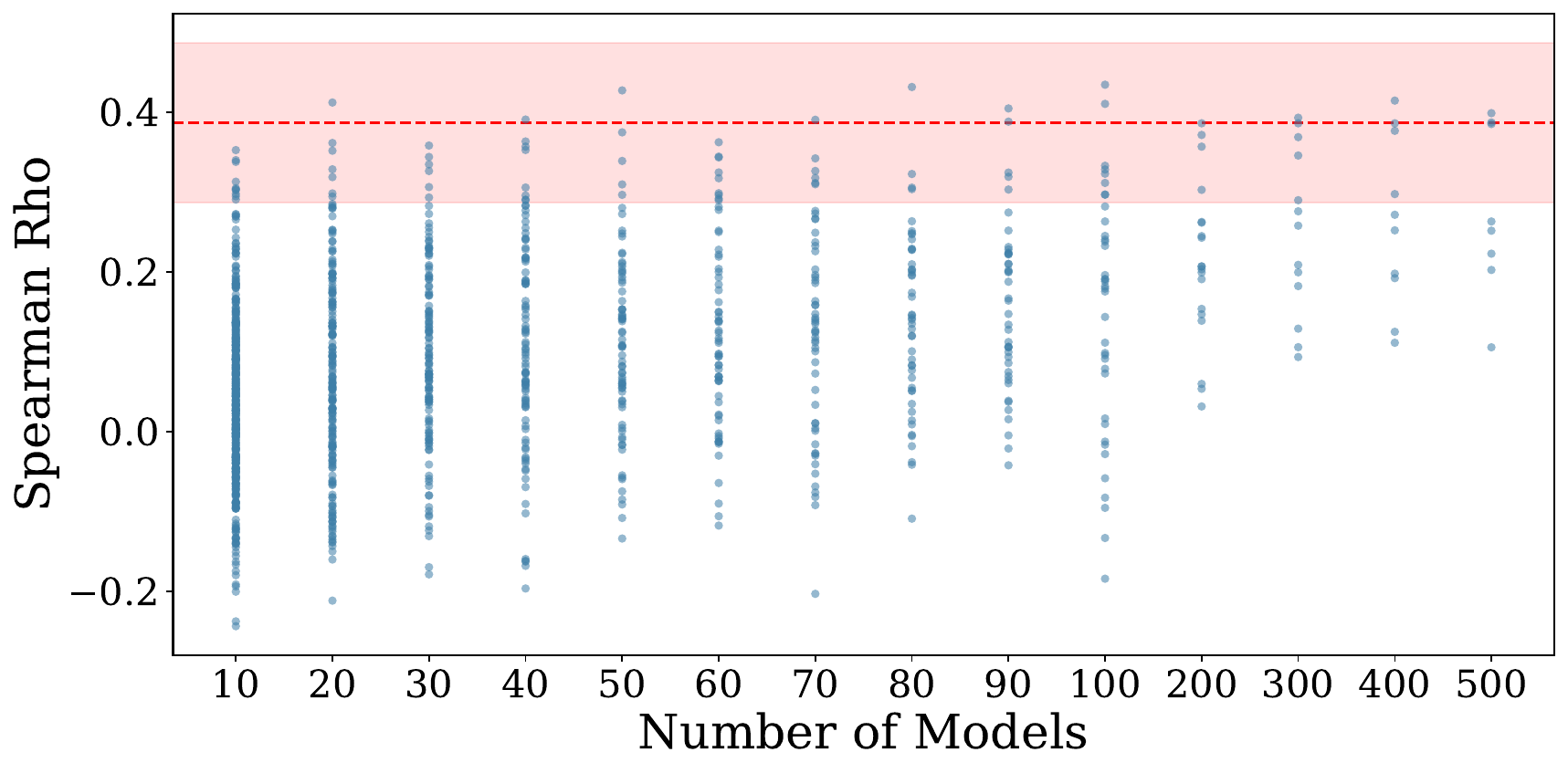}
        \caption{Spearman Rho ($\rho$)}
    \end{subfigure}
    \hfill
    \begin{subfigure}[b]{0.49\linewidth}
        \centering
        \includegraphics[width=\linewidth]{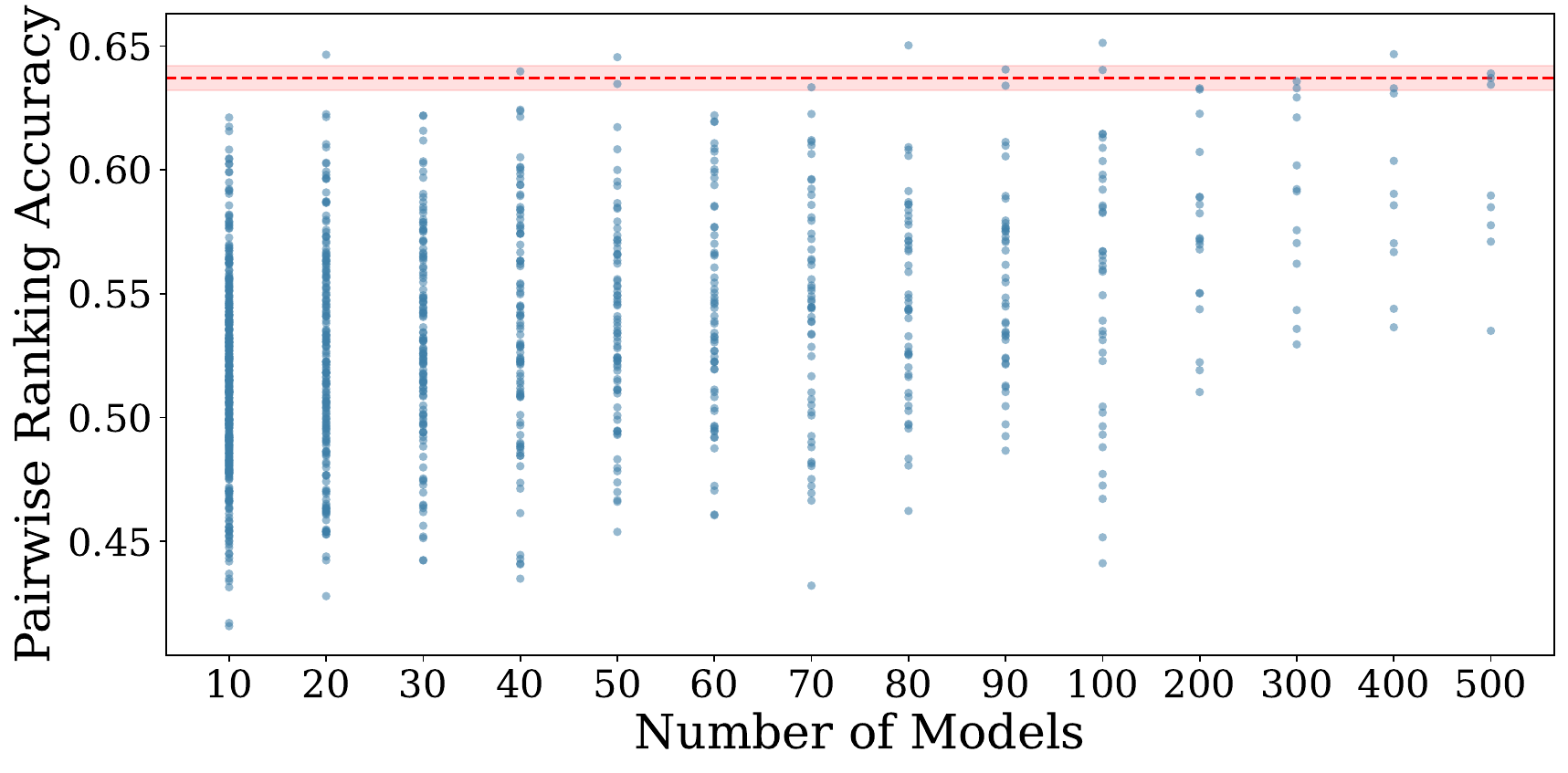}
        \caption{Pairwise Ranking Accuracy ($Acc_{pair}$)}
    \end{subfigure}
    \caption{$\text{GPT2}$ - Target 30B+. Aligned benchmarks created with different numbers of models. In (a), each point shows the Spearman Rho for one subset of models. In (b), each point shows the Pairwise Ranking Accuracy ($Acc_{pair}$).  Error bars obtained using the sample size from the target set. The dotted line represents the results of alignment using all models.}
    \label{fig:bench_align_opt_num_models_30b_helpful_rm2}
\end{figure*}

\begin{figure*}[t]
    \centering
    \begin{subfigure}[b]{0.49\linewidth}
        \centering
        \includegraphics[width=\linewidth]{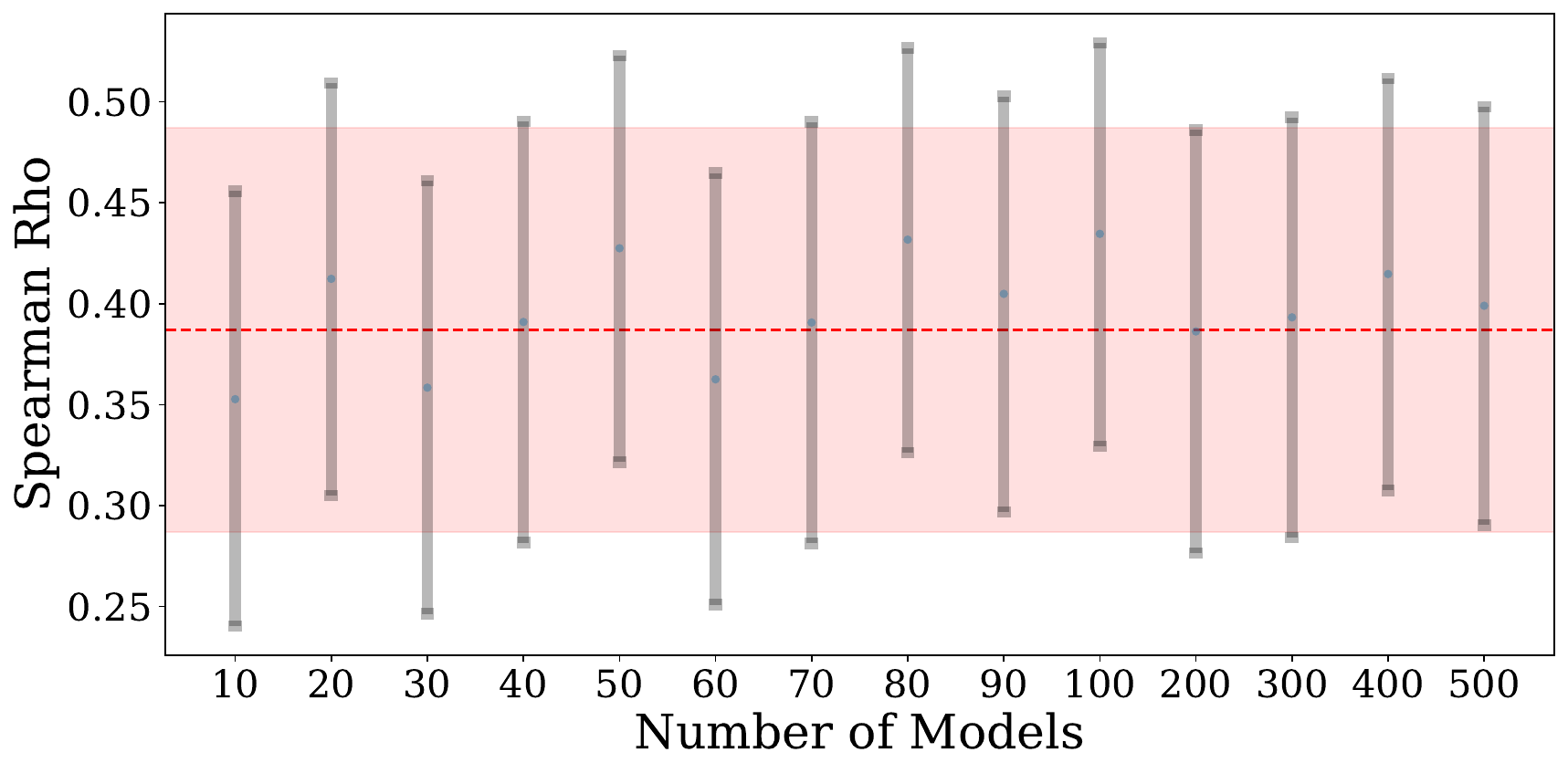}
        \caption{Spearman Rho ($\rho$)}
    \end{subfigure}
    \hfill
    \begin{subfigure}[b]{0.49\linewidth}
        \centering
        \includegraphics[width=\linewidth]{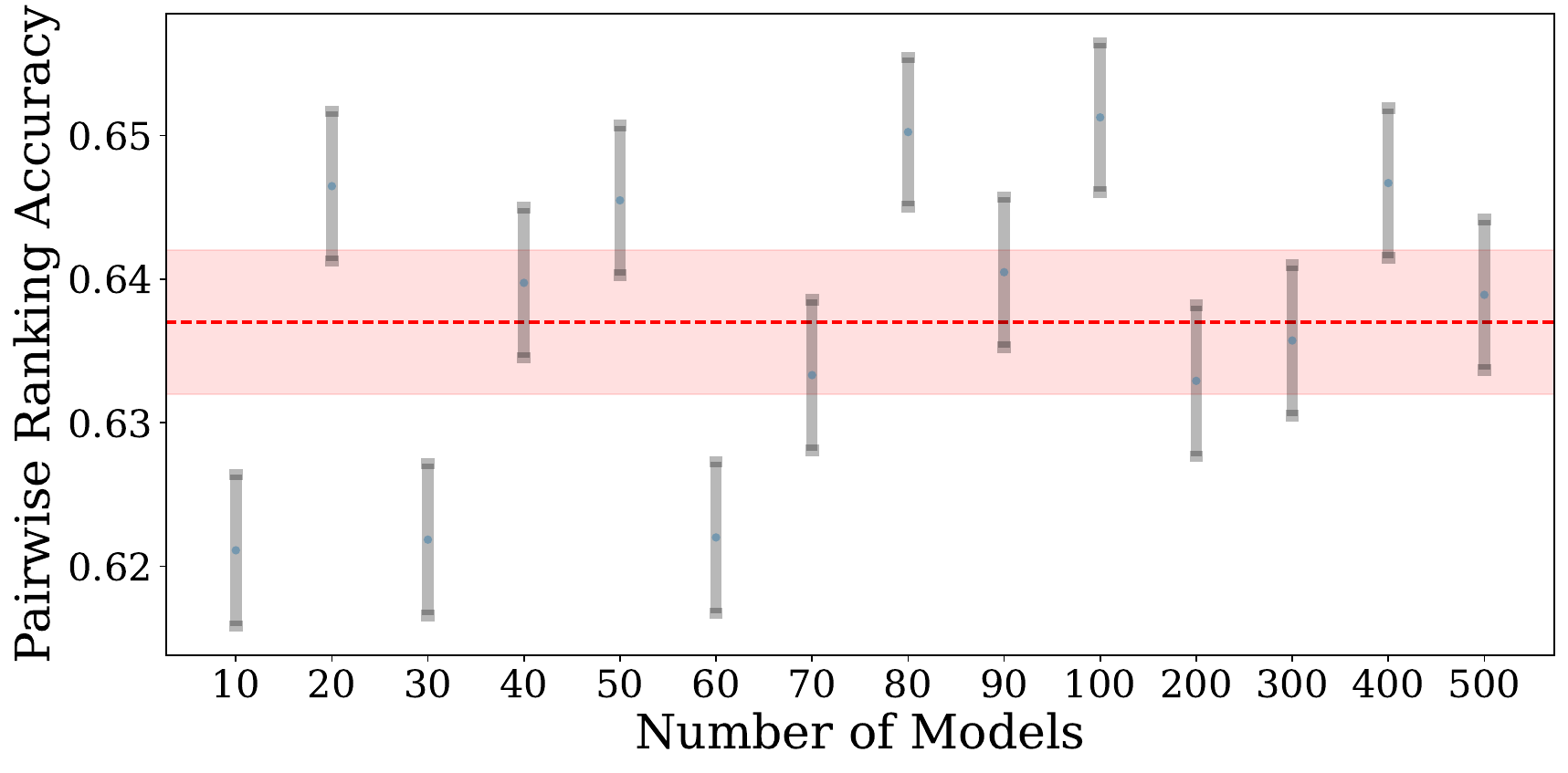}
        \caption{Pairwise Ranking Accuracy ($Acc_{pair}$)}
    \end{subfigure}
    \caption{$\text{GPT2}$ - Target 30B+. Aligned benchmarks created with different numbers of models. In (a), each point reports the Spearman Rho ($\rho$) of the best subset of models. In (b), each point shows the Pairwise Ranking Accuracy ($Acc_{pair}$). Error bars obtained using the sample size from the target set. The dotted line represents the results of alignment using all models.}
    \label{fig:bench_align_opt_num_models_30b_helpful_rm2}
\end{figure*}

\xhdr{Target: 70B+}. Figures \ref{fig:bench_align_opt_num_models_70b_helpful_rm2} and \ref{fig:bench_align_opt_num_models_70b_helpful_rm2} represent the results of all subset of models and the best subset by number of models, respectively.

\begin{figure*}[t]
    \centering
    \begin{subfigure}[b]{0.49\linewidth}
        \centering
        \includegraphics[width=\linewidth]{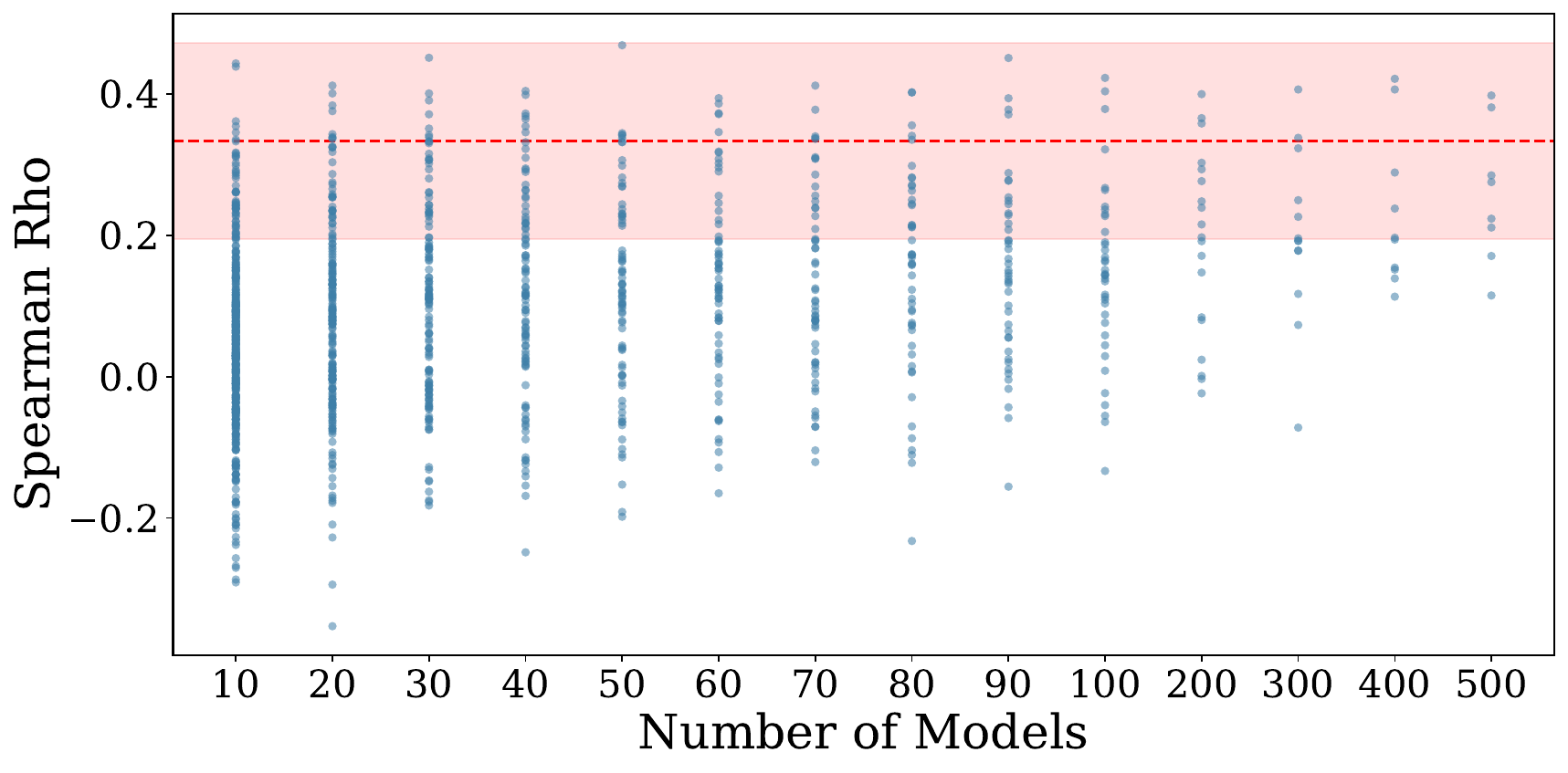}
        \caption{Spearman Rho ($\rho$)}
    \end{subfigure}
    \hfill
    \begin{subfigure}[b]{0.49\linewidth}
        \centering
        \includegraphics[width=\linewidth]{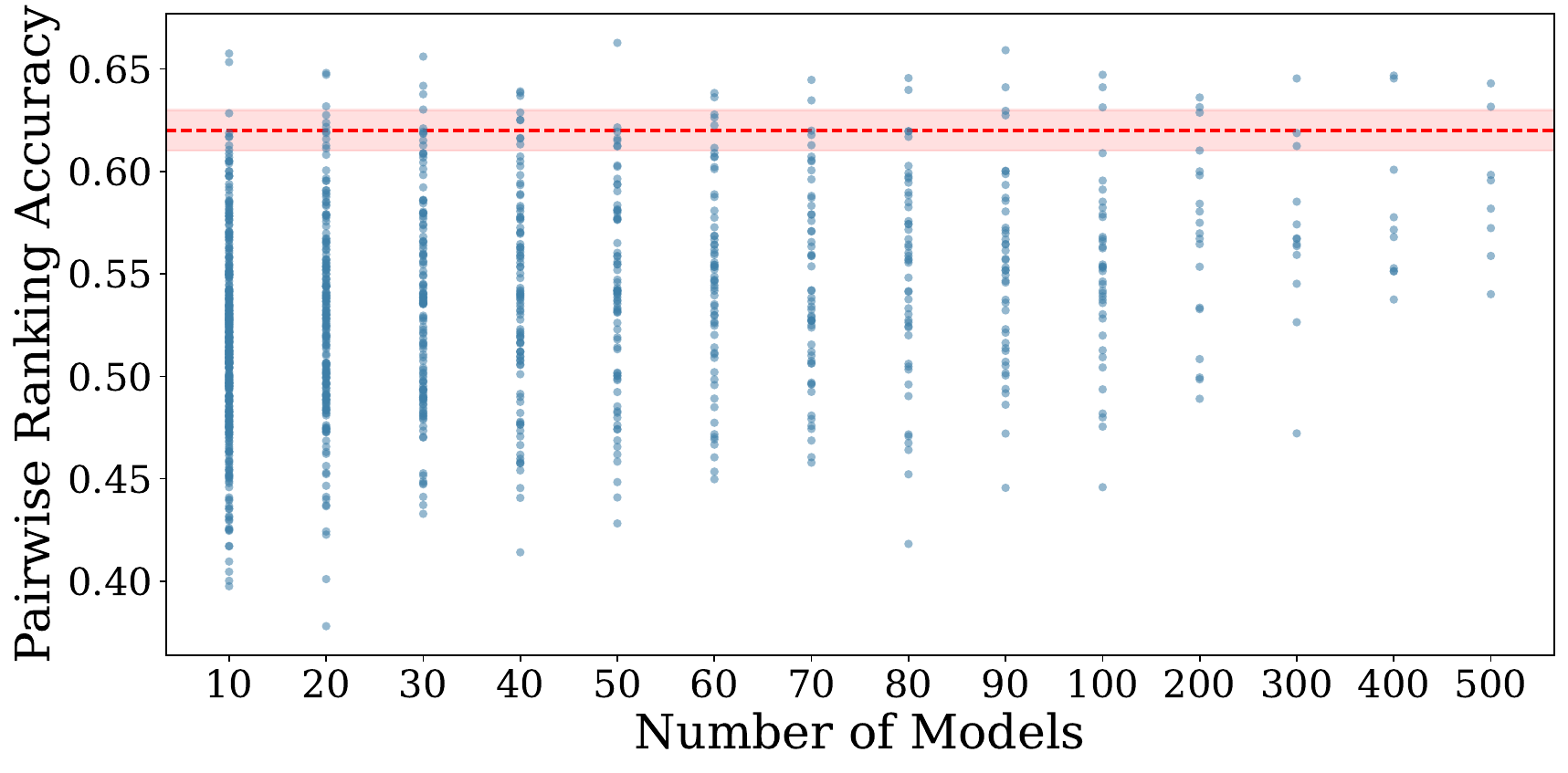}
        \caption{Pairwise Ranking Accuracy ($Acc_{pair}$)}
    \end{subfigure}
    \caption{$\text{GPT2}$ - Target 70B+. Aligned benchmarks created with different numbers of models. In (a), each point shows the Spearman Rho for one subset of models. In (b), each point shows the Pairwise Ranking Accuracy ($Acc_{pair}$).  Error bars obtained using the sample size from the target set. The dotted line represents the results of alignment using all models.}
    \label{fig:bench_align_opt_num_models_70b_helpful_rm2}
\end{figure*}

\begin{figure*}[t]
    \centering
    \begin{subfigure}[b]{0.49\linewidth}
        \centering
        \includegraphics[width=\linewidth]{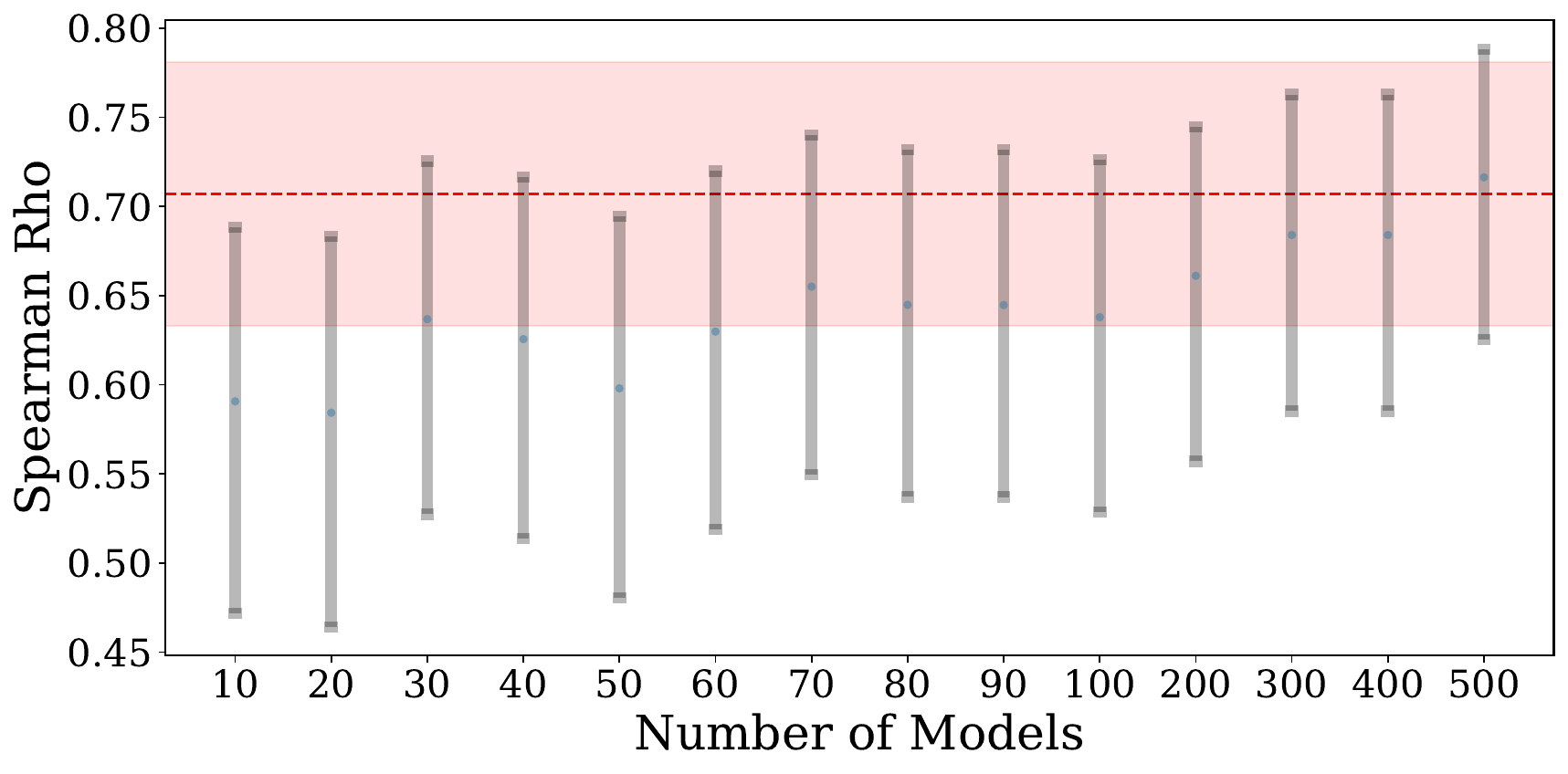}
        \caption{Spearman Rho ($\rho$)}
    \end{subfigure}
    \hfill
    \begin{subfigure}[b]{0.49\linewidth}
        \centering
        \includegraphics[width=\linewidth]{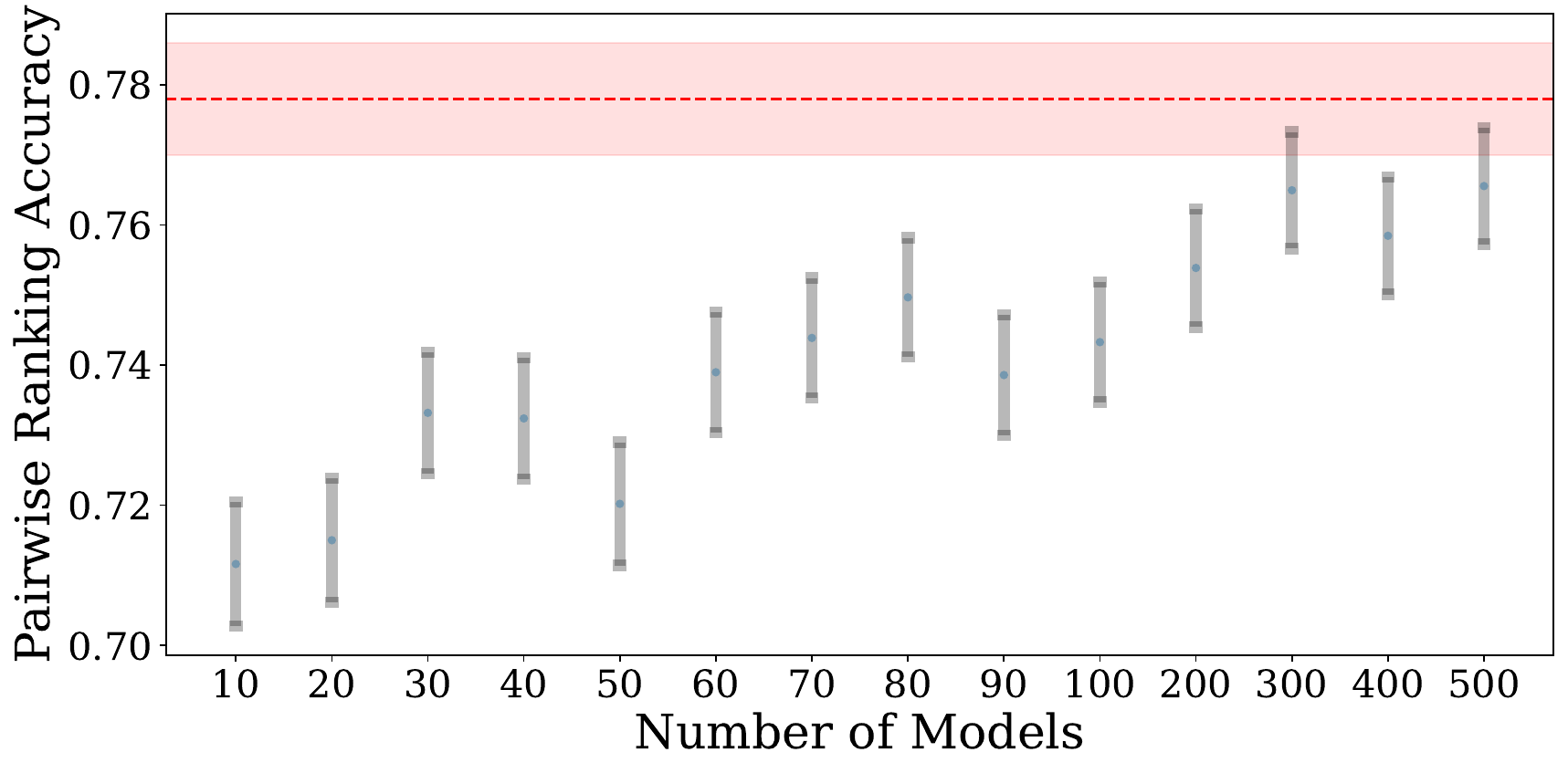}
        \caption{Pairwise Ranking Accuracy ($Acc_{pair}$)}
    \end{subfigure}
    \caption{$\text{GPT2}$ - Target 70B+. Aligned benchmarks created with different numbers of models. In (a), each point reports the Spearman Rho ($\rho$) of the best subset of models. In (b), each point shows the Pairwise Ranking Accuracy ($Acc_{pair}$). Error bars obtained using the sample size from the target set. The dotted line represents the results of alignment using all models.}
    \label{fig:bench_align_opt_num_models_70b_helpful_rm2}
\end{figure*}

\clearpage
\subsubsection{$\text{ArmoRM}$ (Honest)}

\xhdr{Target: 13B+} Figures \ref{fig:bench_align_opt_num_models_13b_honesty_rm1} and \ref{fig:bench_align_opt_num_models_13b_honesty_rm1} represent the results of all subset of models and the best subset by number of models, respectively.

\begin{figure*}[t]
    \centering
    \begin{subfigure}[b]{0.49\linewidth}
        \centering
        \includegraphics[width=\linewidth]{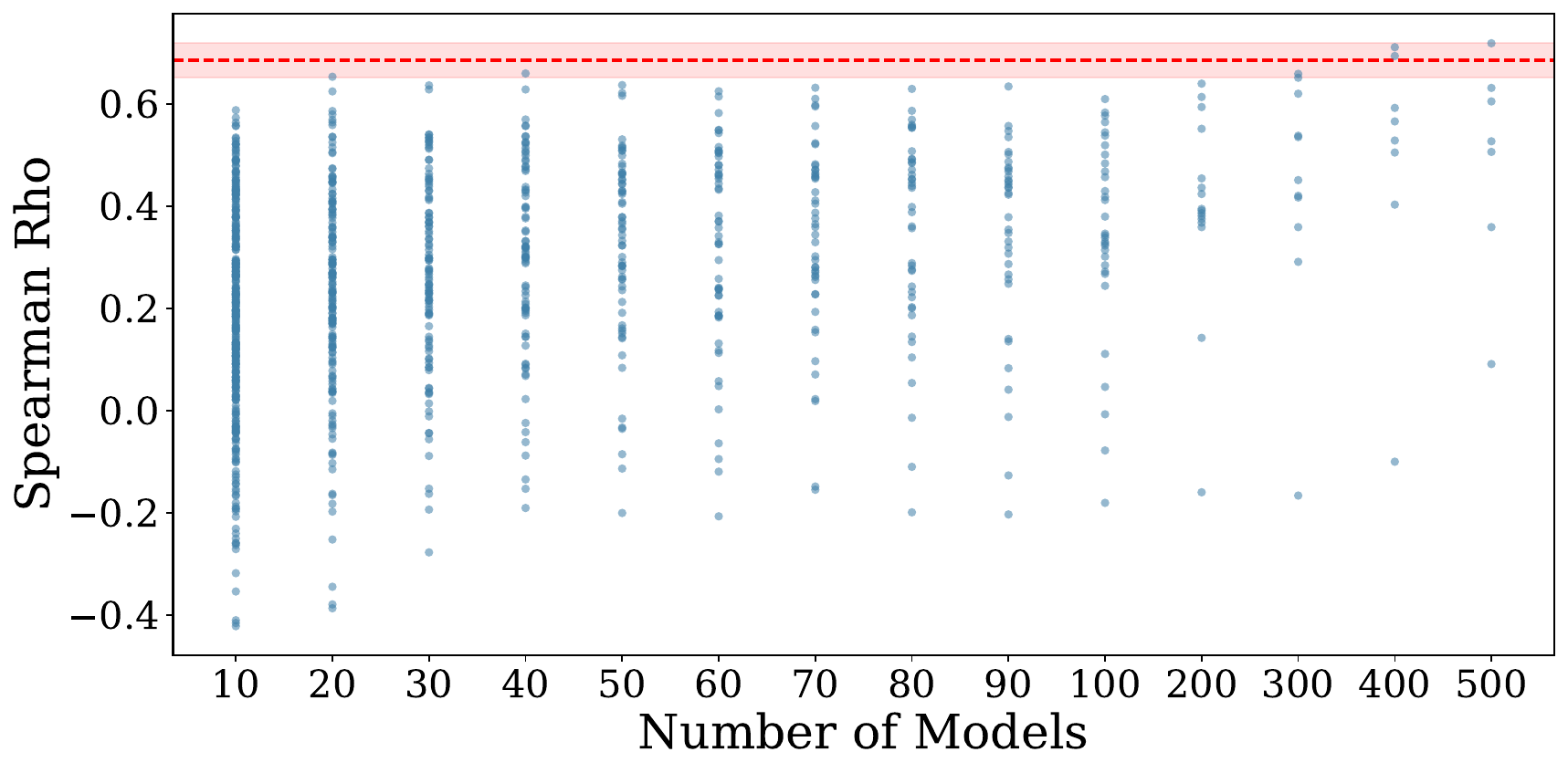}
        \caption{Spearman Rho ($\rho$)}
    \end{subfigure}
    \hfill
    \begin{subfigure}[b]{0.49\linewidth}
        \centering
        \includegraphics[width=\linewidth]{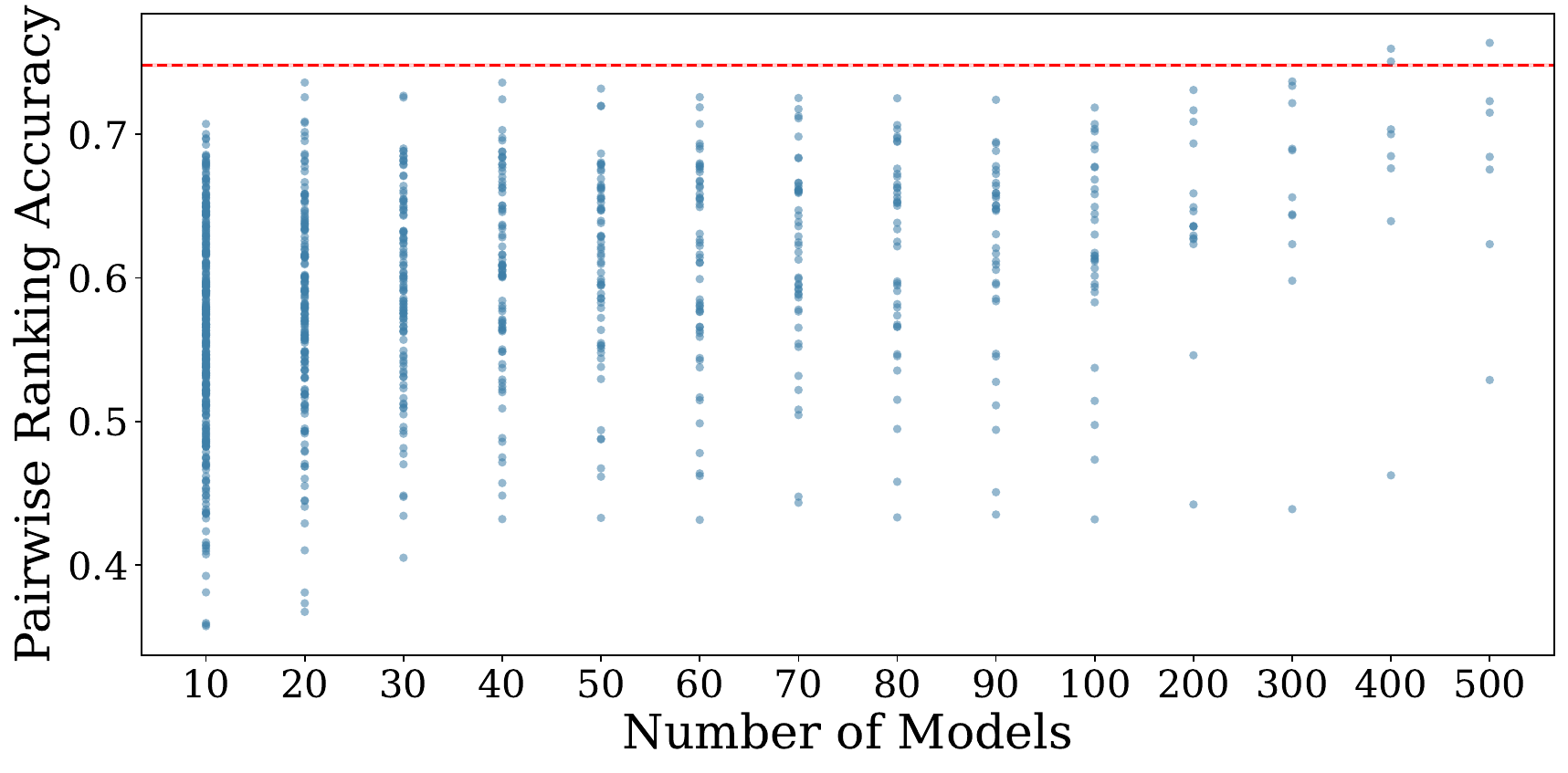}
        \caption{Pairwise Ranking Accuracy ($Acc_{pair}$)}
    \end{subfigure}
    \caption{$\text{ArmoRM}$ (Honest) - Target 13B+. Aligned benchmarks created with different numbers of models. In (a), each point shows the Spearman Rho for one subset of models. In (b), each point shows the Pairwise Ranking Accuracy ($Acc_{pair}$).  Error bars obtained using the sample size from the target set. The dotted line represents the results of alignment using all models.}
    \label{fig:bench_align_opt_num_models_13b_honesty_rm1}
\end{figure*}

\begin{figure*}[t]
    \centering
    \begin{subfigure}[b]{0.49\linewidth}
        \centering
        \includegraphics[width=\linewidth]{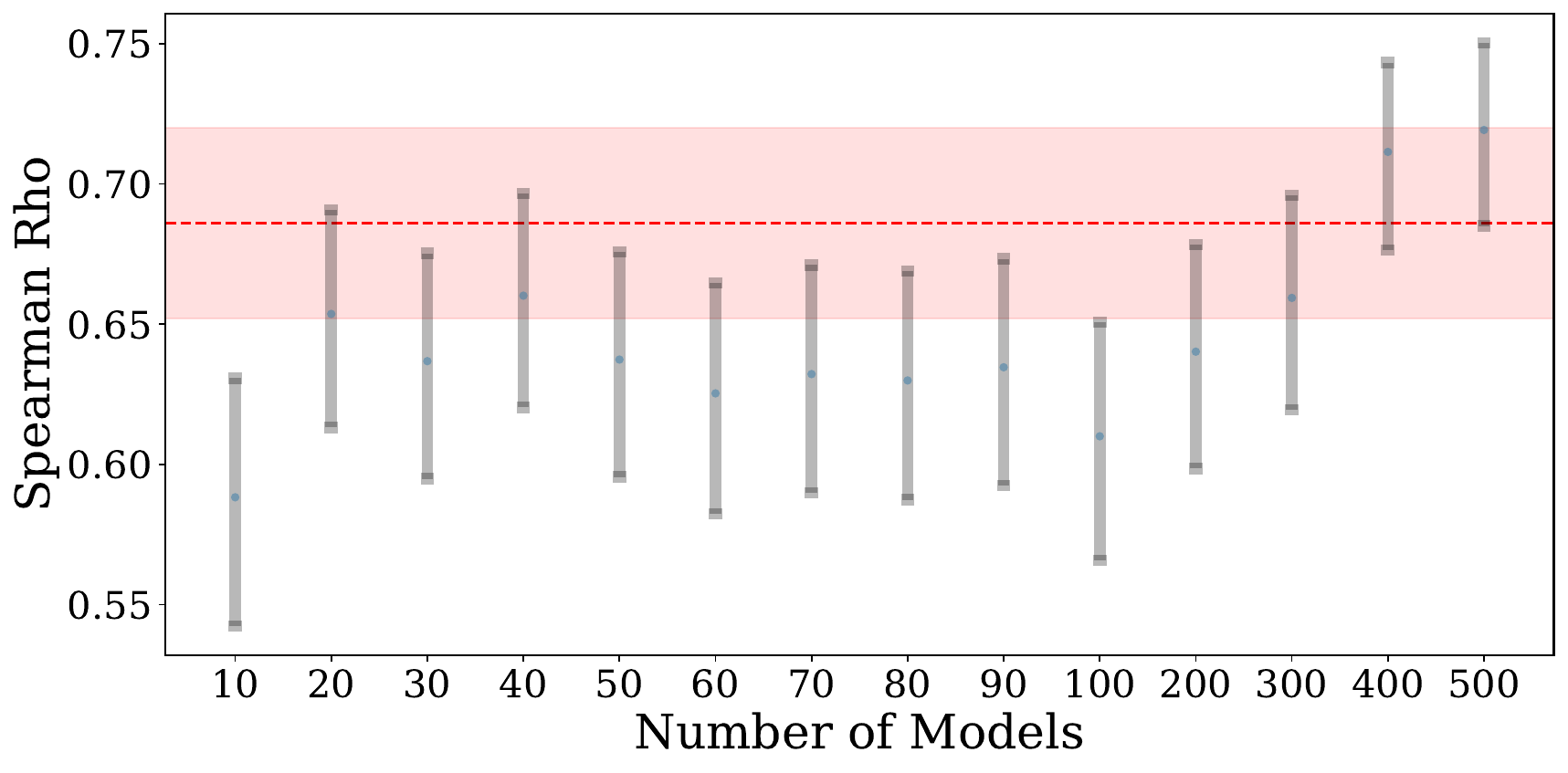}
        \caption{Spearman Rho ($\rho$)}
    \end{subfigure}
    \hfill
    \begin{subfigure}[b]{0.49\linewidth}
        \centering
        \includegraphics[width=\linewidth]{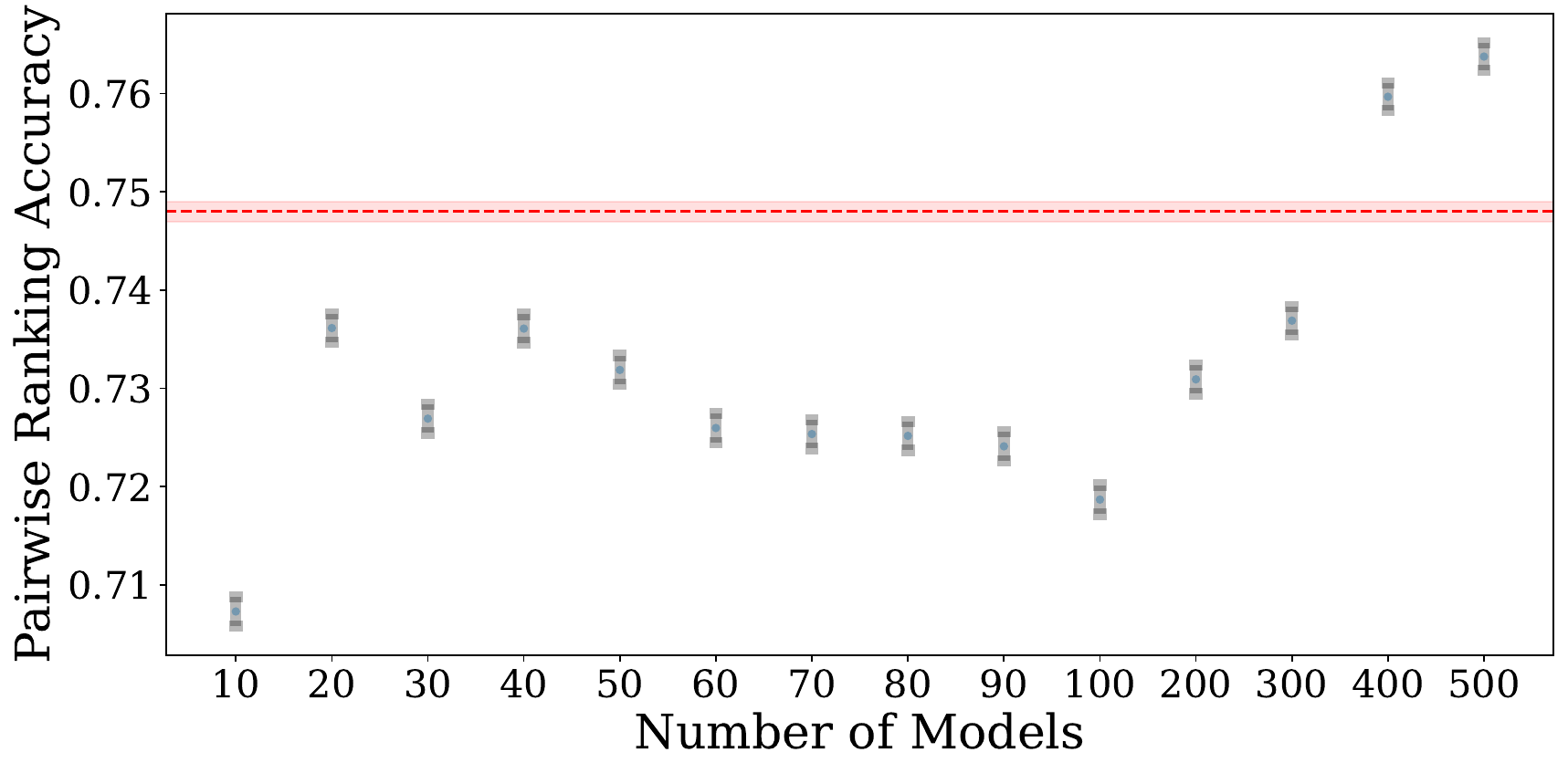}
        \caption{Pairwise Ranking Accuracy ($Acc_{pair}$)}
    \end{subfigure}
    \caption{$\text{ArmoRM}$ (Honest) - Target 13B+. Aligned benchmarks created with different numbers of models. In (a), each point reports the Spearman Rho ($\rho$) of the best subset of models. In (b), each point shows the Pairwise Ranking Accuracy ($Acc_{pair}$). Error bars obtained using the sample size from the target set. The dotted line represents the results of alignment using all models.}
    \label{fig:bench_align_opt_num_models_13b_honesty_rm1}
\end{figure*}

\xhdr{Target: 30B+} Figures \ref{fig:bench_align_opt_num_models_30b_honesty_rm1} and \ref{fig:bench_align_opt_num_models_30b_honesty_rm1} represent the results of all subset of models and the best subset by number of models, respectively.

\begin{figure*}[t]
    \centering
    \begin{subfigure}[b]{0.49\linewidth}
        \centering
        \includegraphics[width=\linewidth]{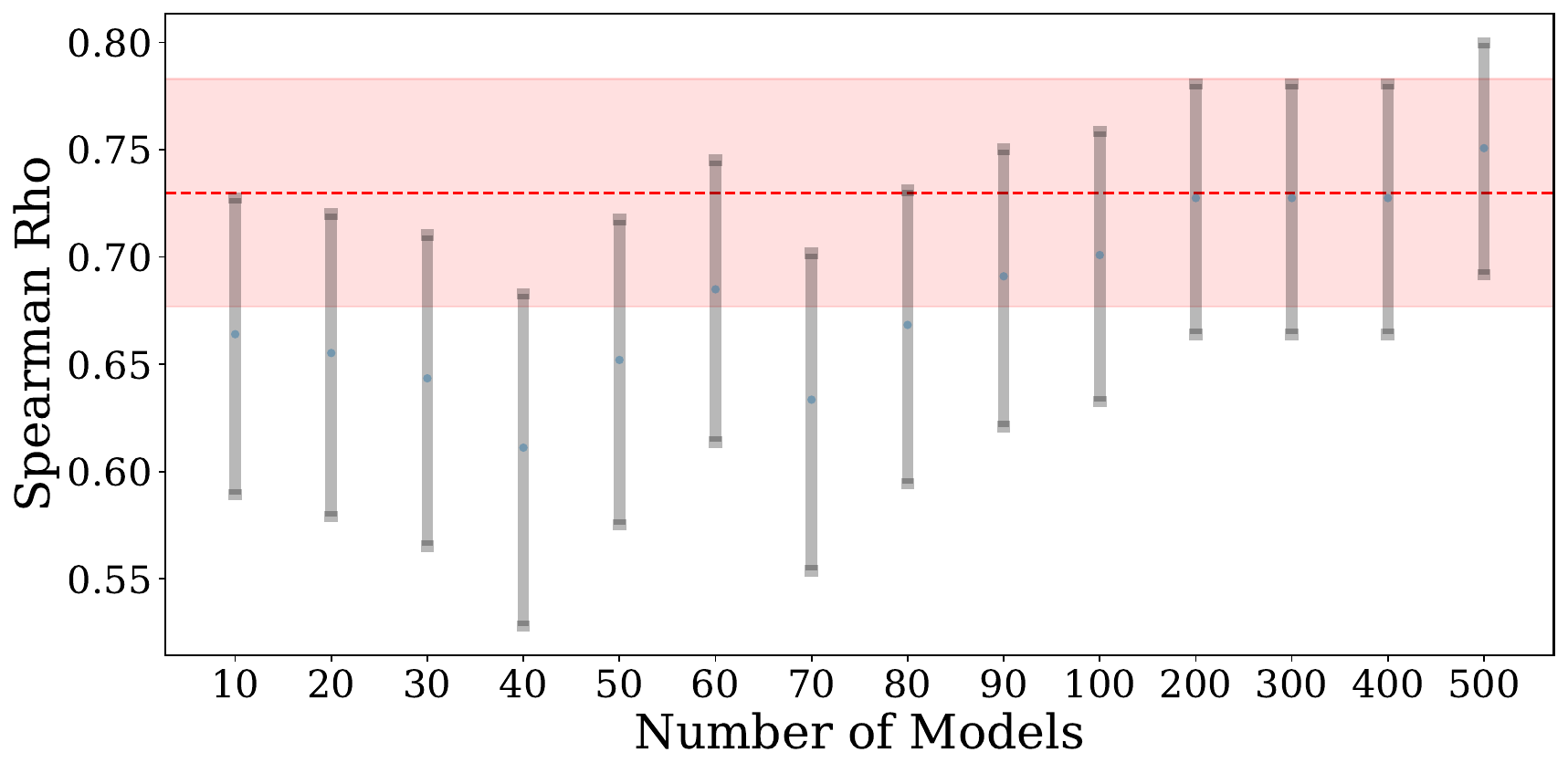}
        \caption{Spearman Rho ($\rho$)}
    \end{subfigure}
    \hfill
    \begin{subfigure}[b]{0.49\linewidth}
        \centering
        \includegraphics[width=\linewidth]{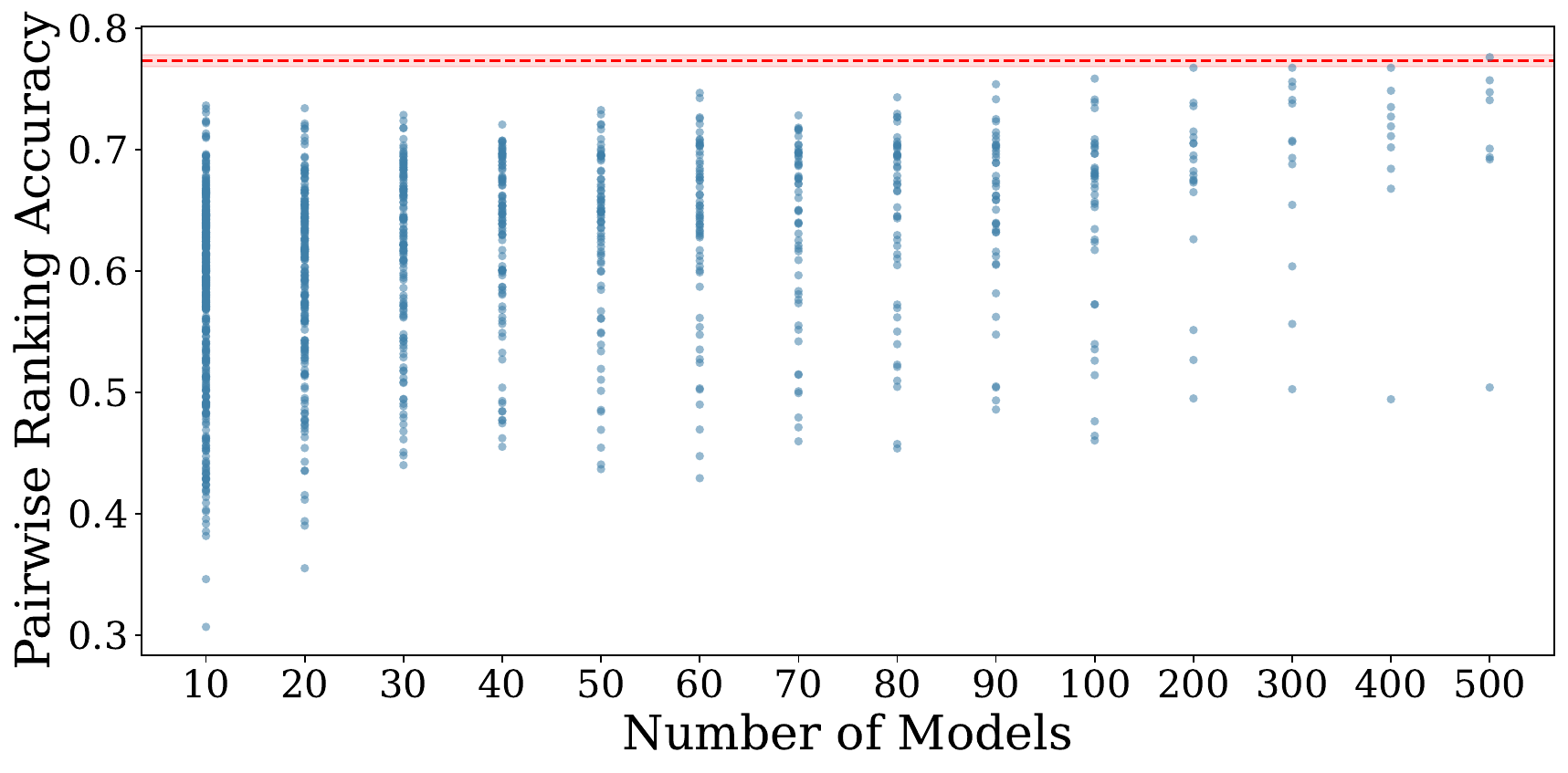}
        \caption{Pairwise Ranking Accuracy ($Acc_{pair}$)}
    \end{subfigure}
    \caption{$\text{ArmoRM}$ (Honest) - Target 30B+. Aligned benchmarks created with different numbers of models. In (a), each point shows the Spearman Rho for one subset of models. In (b), each point shows the Pairwise Ranking Accuracy ($Acc_{pair}$). Error bars obtained using the sample size from the target set. The dotted line represents the results of alignment using all models.}
    \label{fig:bench_align_opt_num_models_30b_honesty_rm1}
\end{figure*}

\begin{figure*}[t]
    \centering
    \begin{subfigure}[b]{0.49\linewidth}
        \centering
        \includegraphics[width=\linewidth]{images/model_window_best_exps/armo-ultrafeedback-honesty/30b_70b/pdfs/armo-ultrafeedback-honesty_30b_70b_rho_best_vs_window.pdf}
        \caption{Spearman Rho ($\rho$)}
    \end{subfigure}
    \hfill
    \begin{subfigure}[b]{0.49\linewidth}
        \centering
        \includegraphics[width=\linewidth]{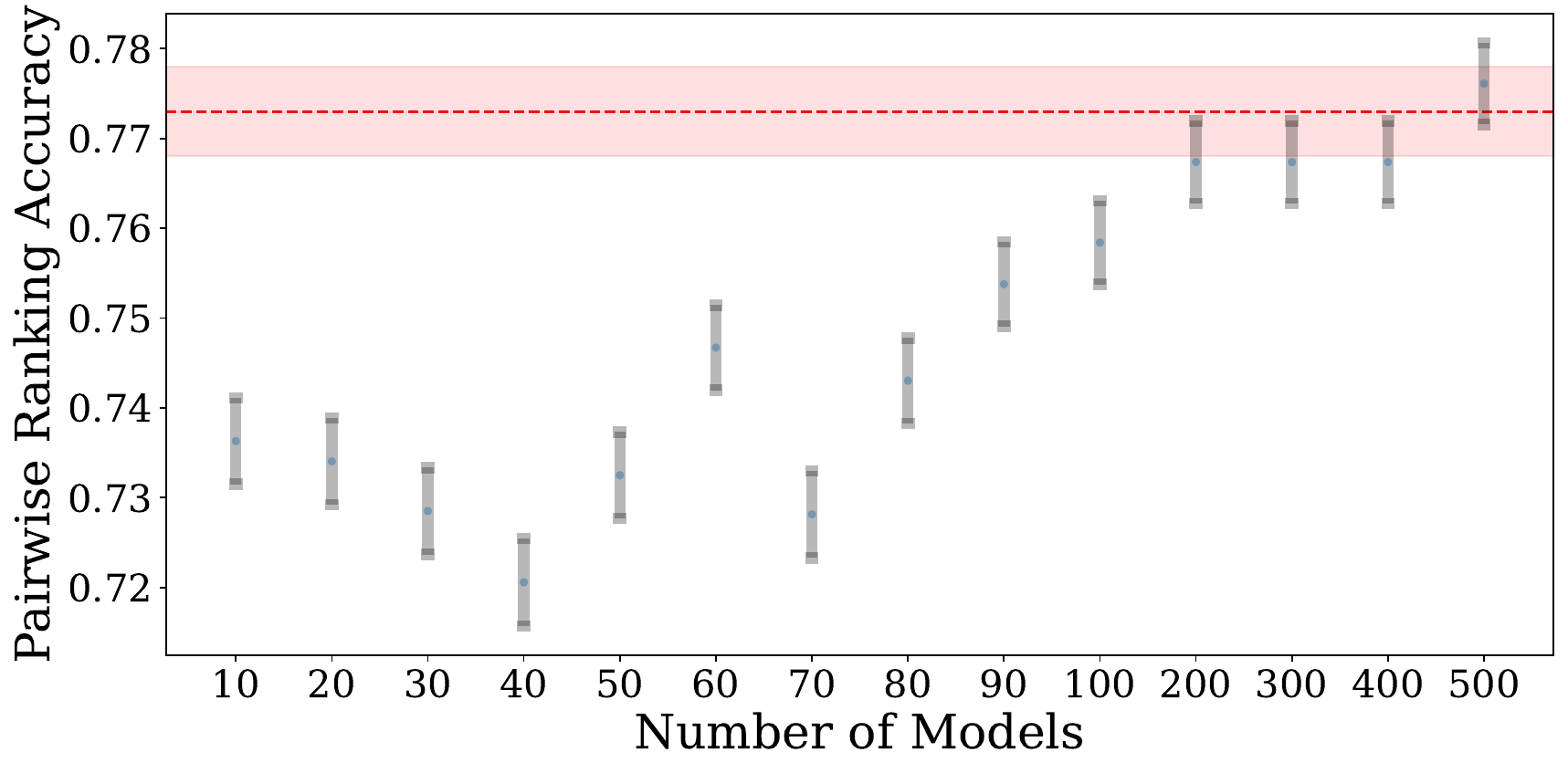}
        \caption{Pairwise Ranking Accuracy ($Acc_{pair}$)}
    \end{subfigure}
    \caption{$\text{ArmoRM}$ (Honest) - Target 30B+. Aligned benchmarks created with different numbers of models. In (a), each point reports the Spearman Rho ($\rho$) of the best subset of models. In (b), each point shows the Pairwise Ranking Accuracy ($Acc_{pair}$). Error bars obtained using the sample size from the target set. The dotted line represents the results of alignment using all models.}
    \label{fig:bench_align_opt_num_models_30b_honesty_rm1}
\end{figure*}

\xhdr{Target: 70B+} Figures \ref{fig:bench_align_opt_num_models_70b_honesty_rm1} and \ref{fig:bench_align_opt_num_models_70b_honesty_rm1} represent the results of all subset of models and the best subset by number of models, respectively.

\begin{figure*}[t]
    \centering
    \begin{subfigure}[b]{0.49\linewidth}
        \centering
        \includegraphics[width=\linewidth]{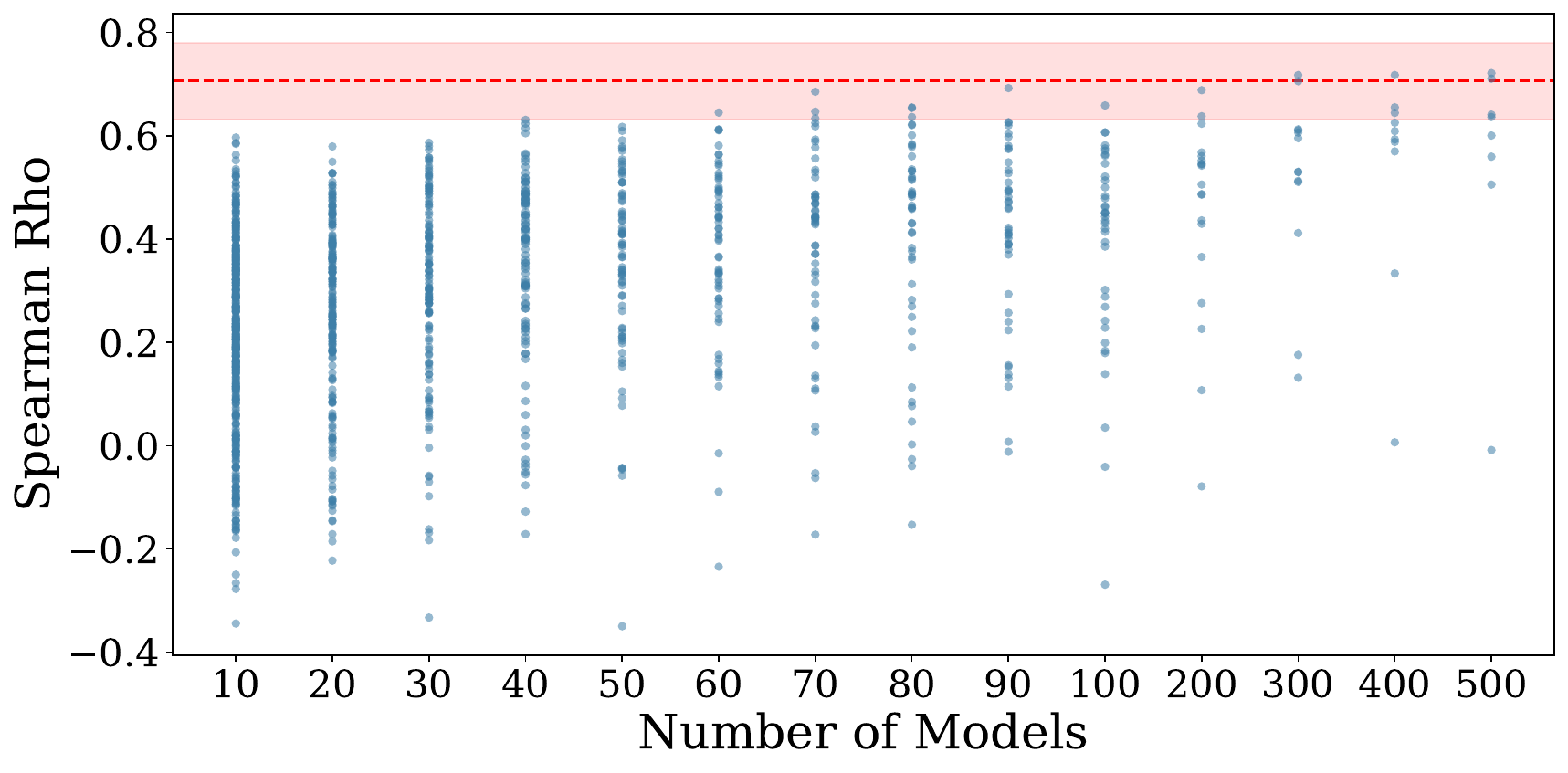}
        \caption{Spearman Rho ($\rho$)}
    \end{subfigure}
    \hfill
    \begin{subfigure}[b]{0.49\linewidth}
        \centering
        \includegraphics[width=\linewidth]{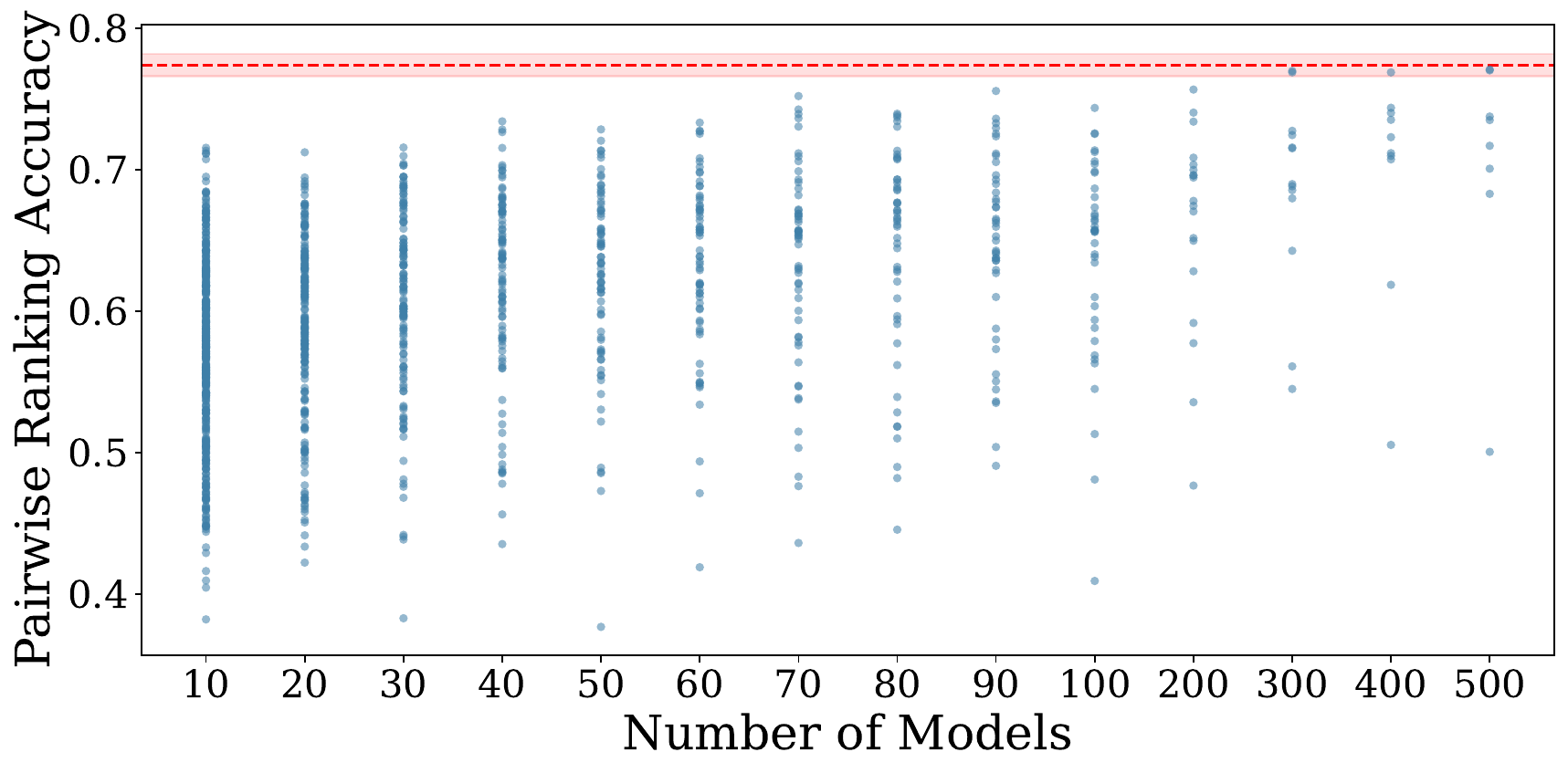}
        \caption{Pairwise Ranking Accuracy ($Acc_{pair}$)}
    \end{subfigure}
    \caption{$\text{ArmoRM}$ (Honest) - Target 70B+. Aligned benchmarks created with different numbers of models. In (a), each point shows the Spearman Rho for one subset of models. In (b), each point shows the Pairwise Ranking Accuracy ($Acc_{pair}$).  Error bars obtained using the sample size from the target set. The dotted line represents the results of alignment using all models.}
    \label{fig:bench_align_opt_num_models_70b_honesty_rm1}
\end{figure*}

\begin{figure*}[t]
    \centering
    \begin{subfigure}[b]{0.49\linewidth}
        \centering
        \includegraphics[width=\linewidth]{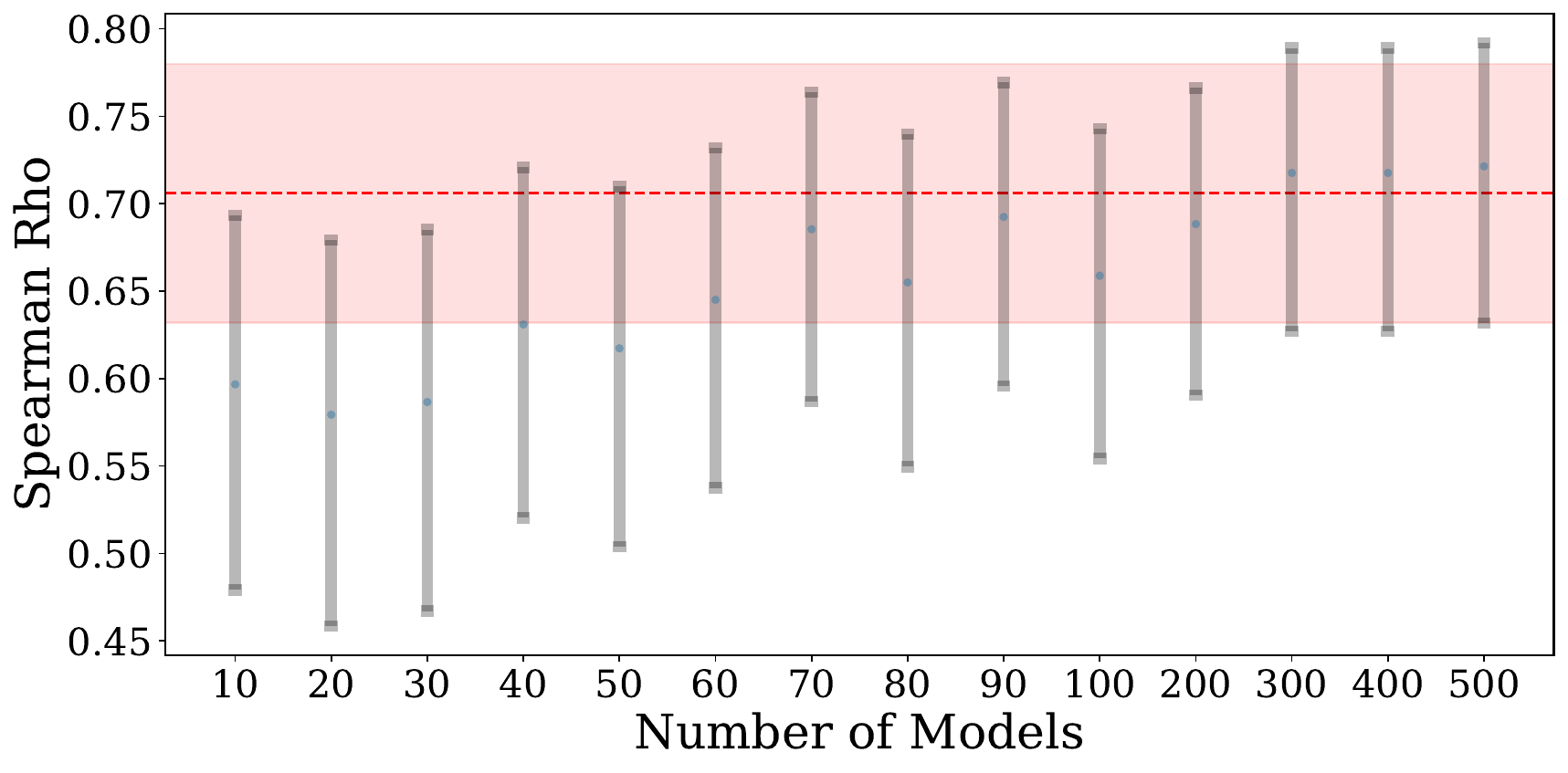}.
        \caption{Spearman Rho ($\rho$)}
    \end{subfigure}
    \hfill
    \begin{subfigure}[b]{0.49\linewidth}
        \centering
        \includegraphics[width=\linewidth]{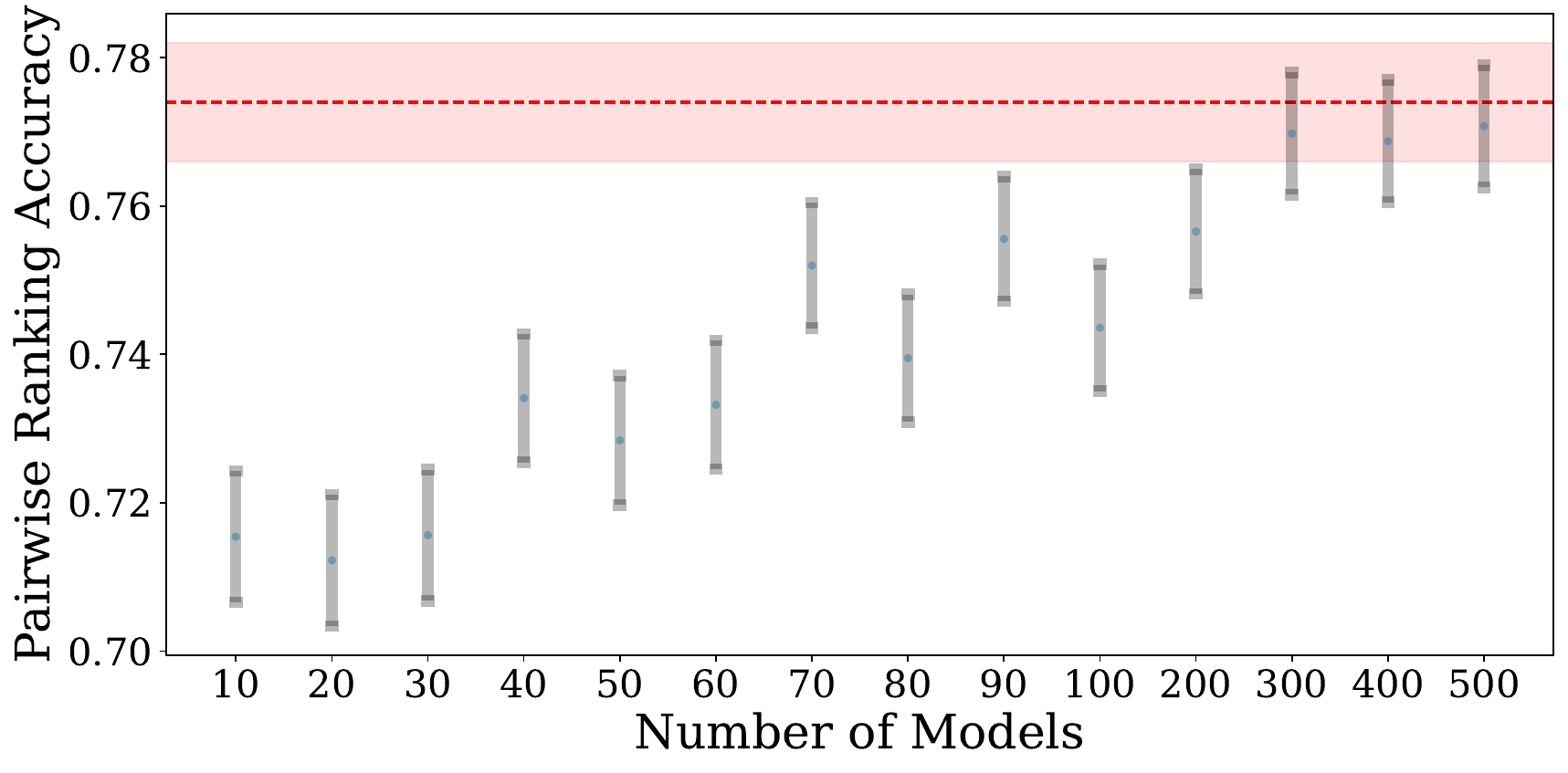}
        \caption{Pairwise Ranking Accuracy ($Acc_{pair}$)}
    \end{subfigure}
    \caption{$\text{ArmoRM}$ (Honest) - Target 70B+. Aligned benchmarks created with different numbers of models. In (a), each point reports the Spearman Rho ($\rho$) of the best subset of models. In (b), each point shows the Pairwise Ranking Accuracy ($Acc_{pair}$). Error bars obtained using the sample size from the target set. The dotted line represents the results of alignment using all models.}
    \label{fig:bench_align_opt_num_models_70b_honesty_rm1}
\end{figure*}

\clearpage
\subsubsection{$\text{DPA}$}

\xhdr{Target: 13B+} Figures \ref{fig:bench_align_opt_num_models_13b_honesty_rm2} and \ref{fig:bench_align_opt_num_models_13b_honesty_rm2} represent the results of all subset of models and the best subset by number of models, respectively.

\begin{figure*}[t]
    \centering
    \begin{subfigure}[b]{0.49\linewidth}
        \centering
        \includegraphics[width=\linewidth]{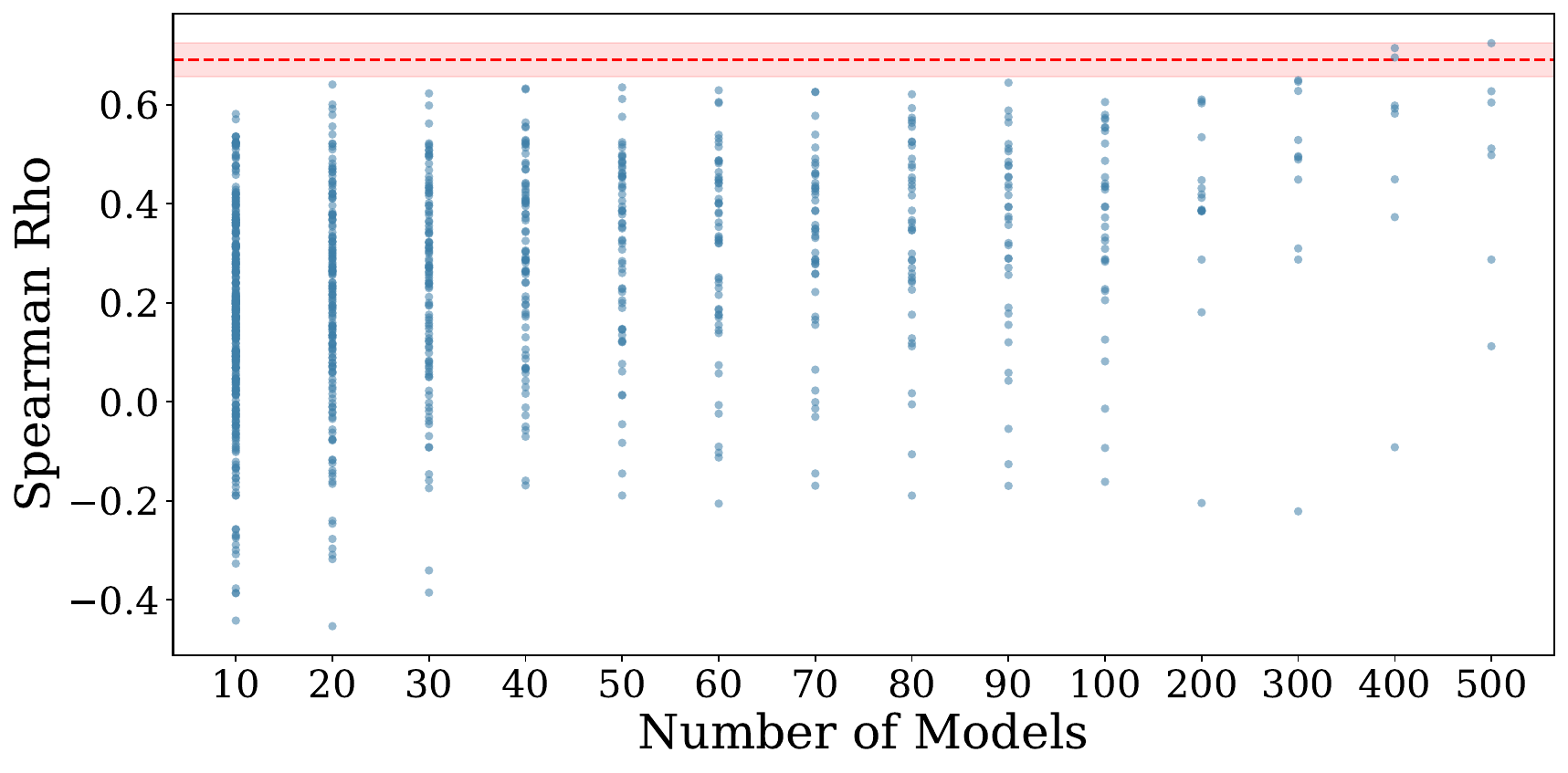}
        \caption{Spearman Rho ($\rho$)}
    \end{subfigure}
    \hfill
    \begin{subfigure}[b]{0.49\linewidth}
        \centering
        \includegraphics[width=\linewidth]{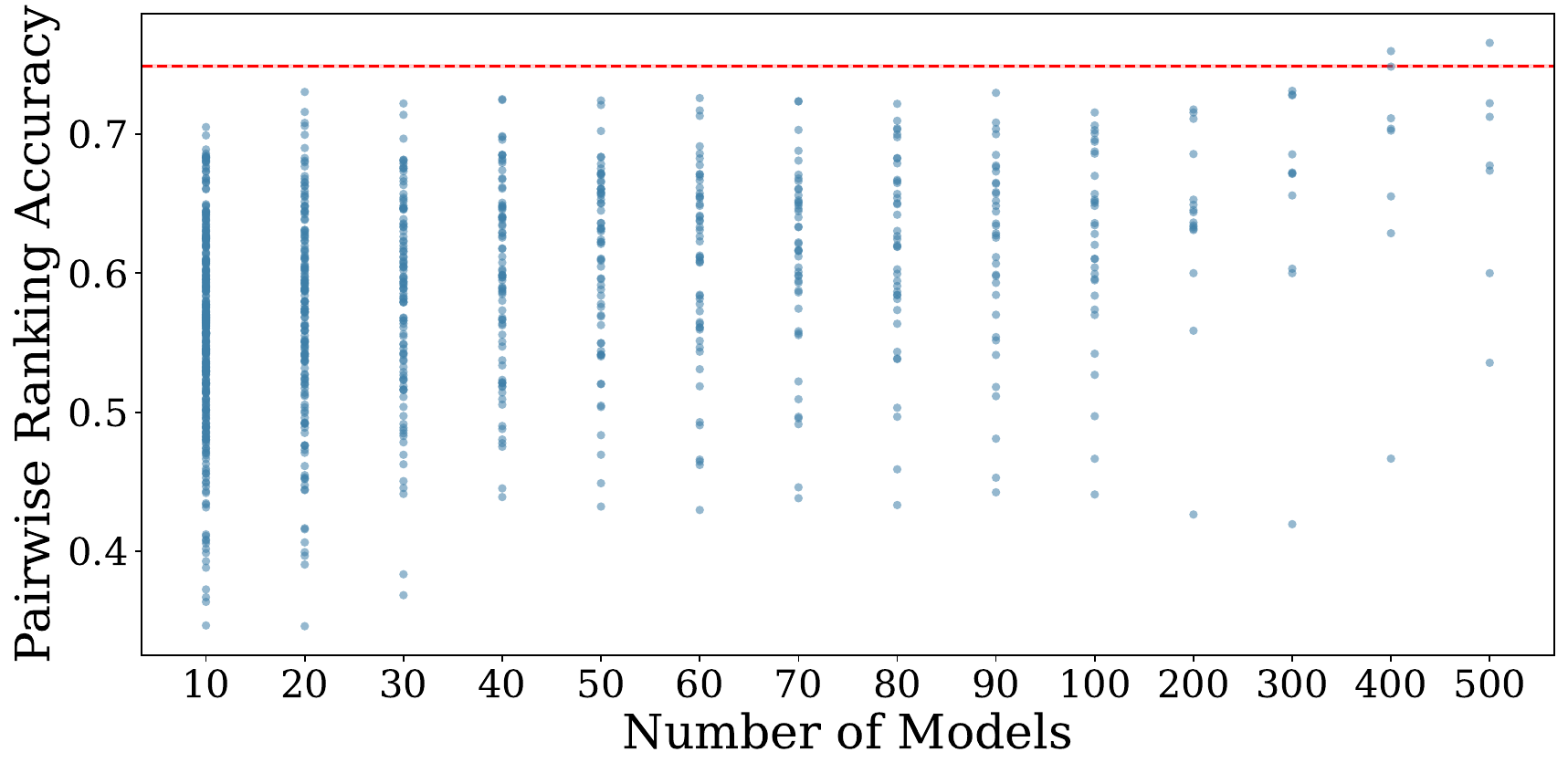}
        \caption{Pairwise Ranking Accuracy ($Acc_{pair}$)}
    \end{subfigure}
    \caption{$\text{DPA}$ - Target 13B+. Aligned benchmarks created with different numbers of models. In (a), each point shows the Spearman Rho for one subset of models. In (b), each point shows the Pairwise Ranking Accuracy ($Acc_{pair}$).  Error bars obtained using the sample size from the target set. The dotted line represents the results of alignment using all models.}
    \label{fig:bench_align_opt_num_models_13b_honesty_rm2}
\end{figure*}

\begin{figure*}[t]
    \centering
    \begin{subfigure}[b]{0.49\linewidth}
        \centering
        \includegraphics[width=\linewidth]{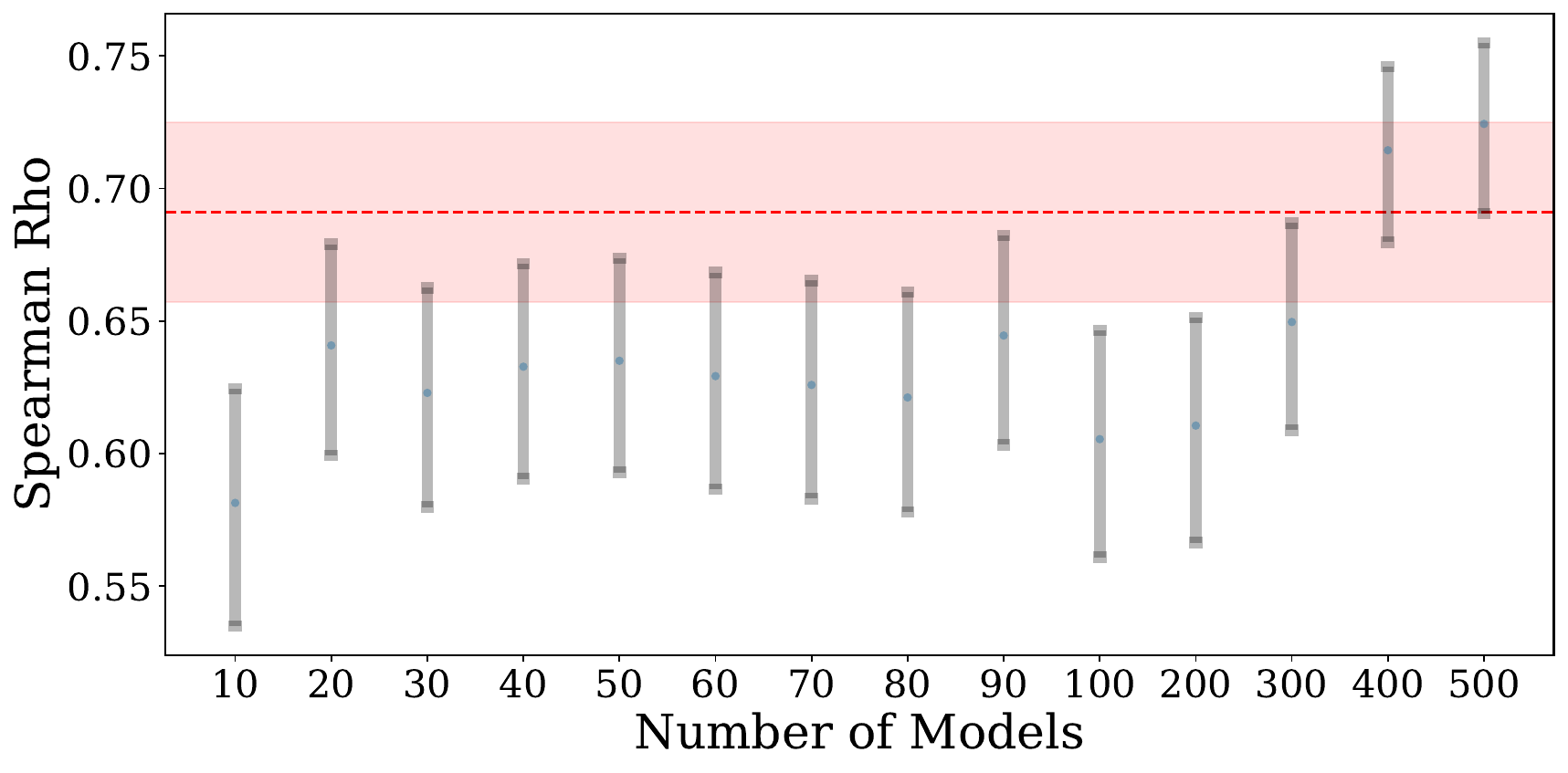}
        \caption{Spearman Rho ($\rho$)}
    \end{subfigure}
    \hfill
    \begin{subfigure}[b]{0.49\linewidth}
        \centering
        \includegraphics[width=\linewidth]{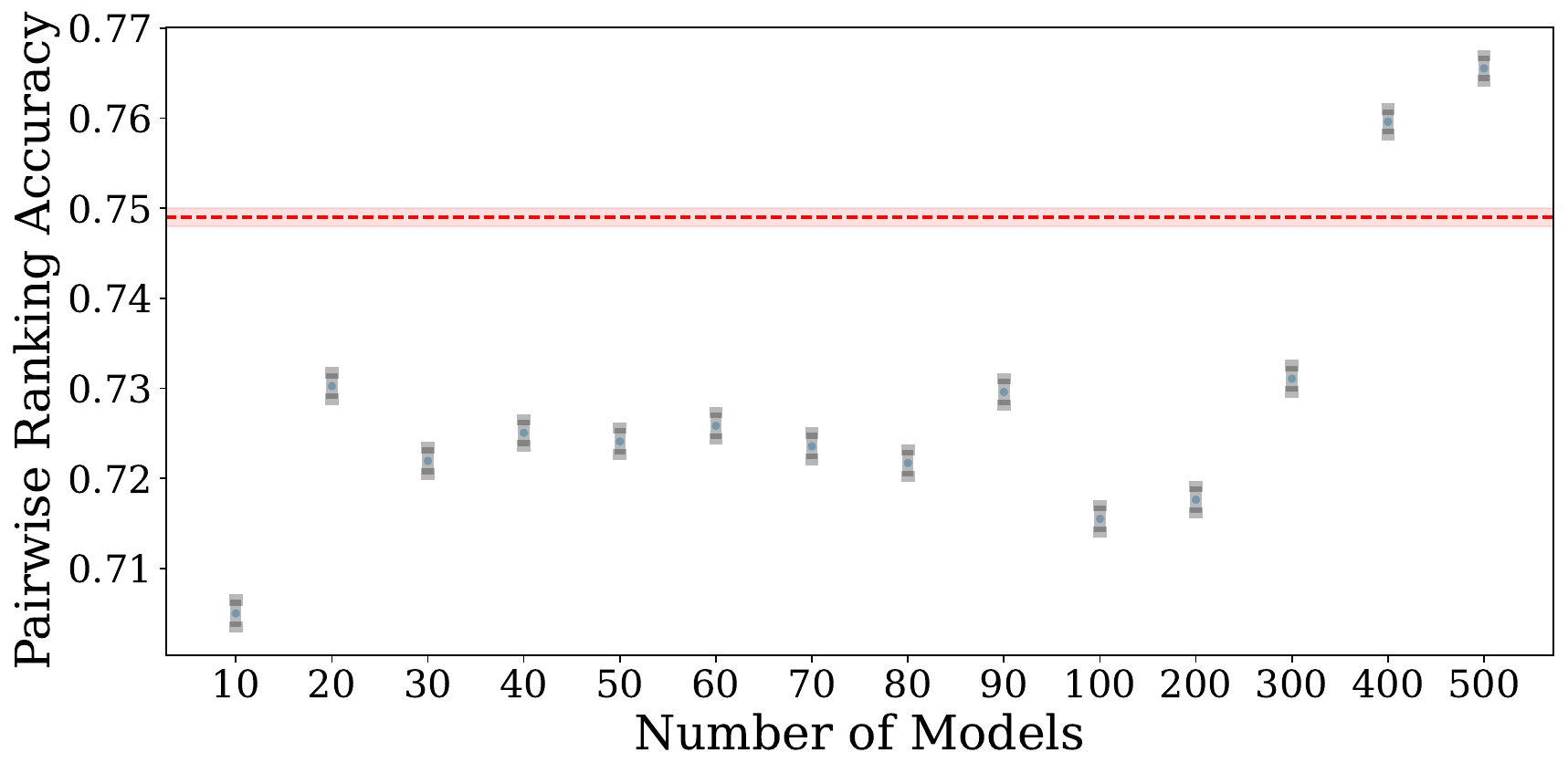}
        \caption{Pairwise Ranking Accuracy ($Acc_{pair}$)}
    \end{subfigure}
    \caption{$\text{DPA}$ - Target 13B+. Aligned benchmarks created with different numbers of models. In (a), each point reports the Spearman Rho ($\rho$) of the best subset of models. In (b), each point shows the Pairwise Ranking Accuracy ($Acc_{pair}$). Error bars obtained using the sample size from the target set. The dotted line represents the results of alignment using all models.}
    \label{fig:bench_align_opt_num_models_13b_honesty_rm2}
\end{figure*}

\xhdr{Target: 30B+} Figures \ref{fig:bench_align_opt_num_models_30b_honesty_rm2} and \ref{fig:bench_align_opt_num_models_30b_honesty_rm2} represent the results of all subset of models and the best subset by number of models, respectively.

\begin{figure*}[t]
    \centering
    \begin{subfigure}[b]{0.49\linewidth}
        \centering
        \includegraphics[width=\linewidth]{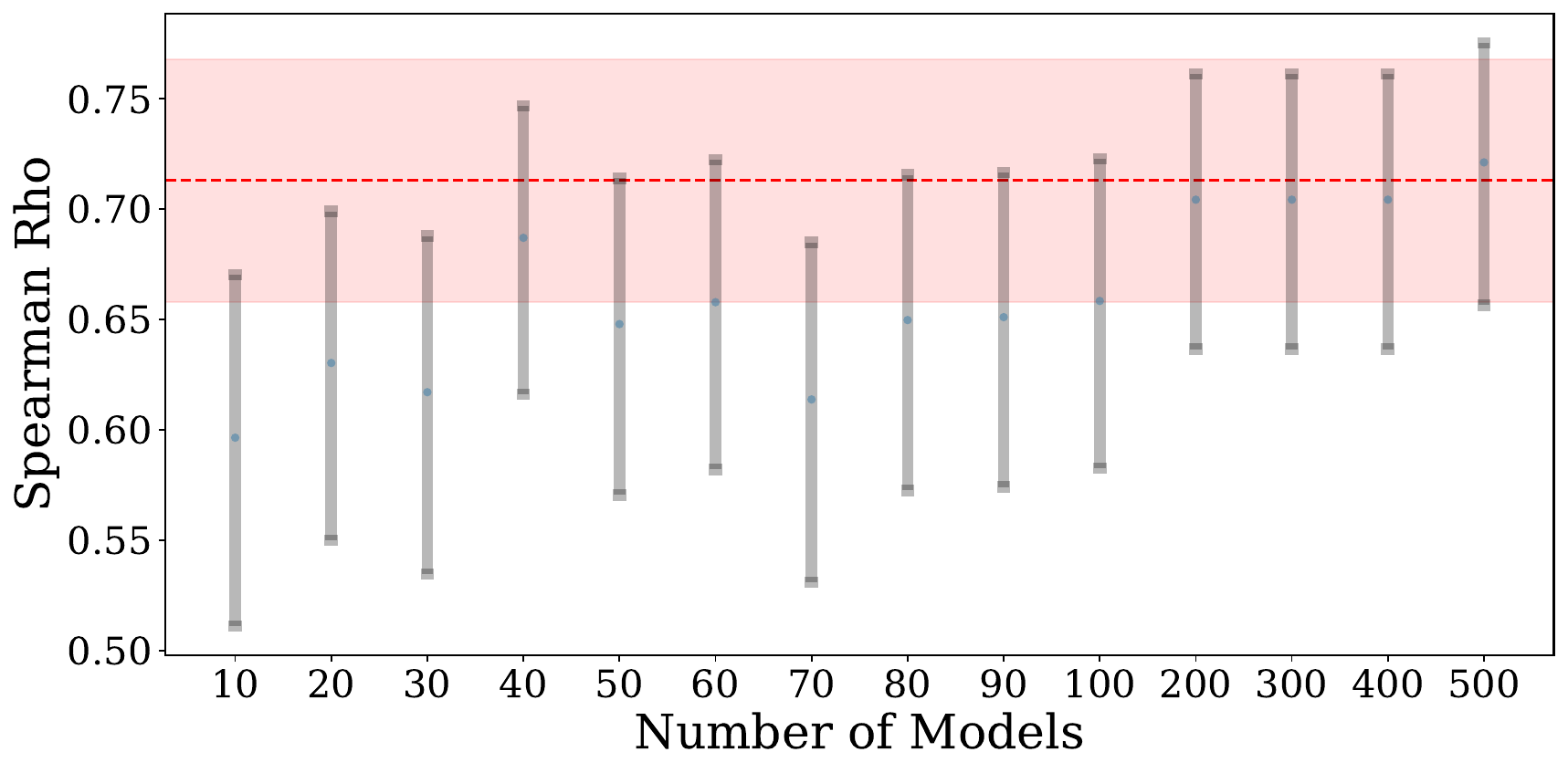}
        \caption{Spearman Rho ($\rho$)}
    \end{subfigure}
    \hfill
    \begin{subfigure}[b]{0.49\linewidth}
        \centering
        \includegraphics[width=\linewidth]{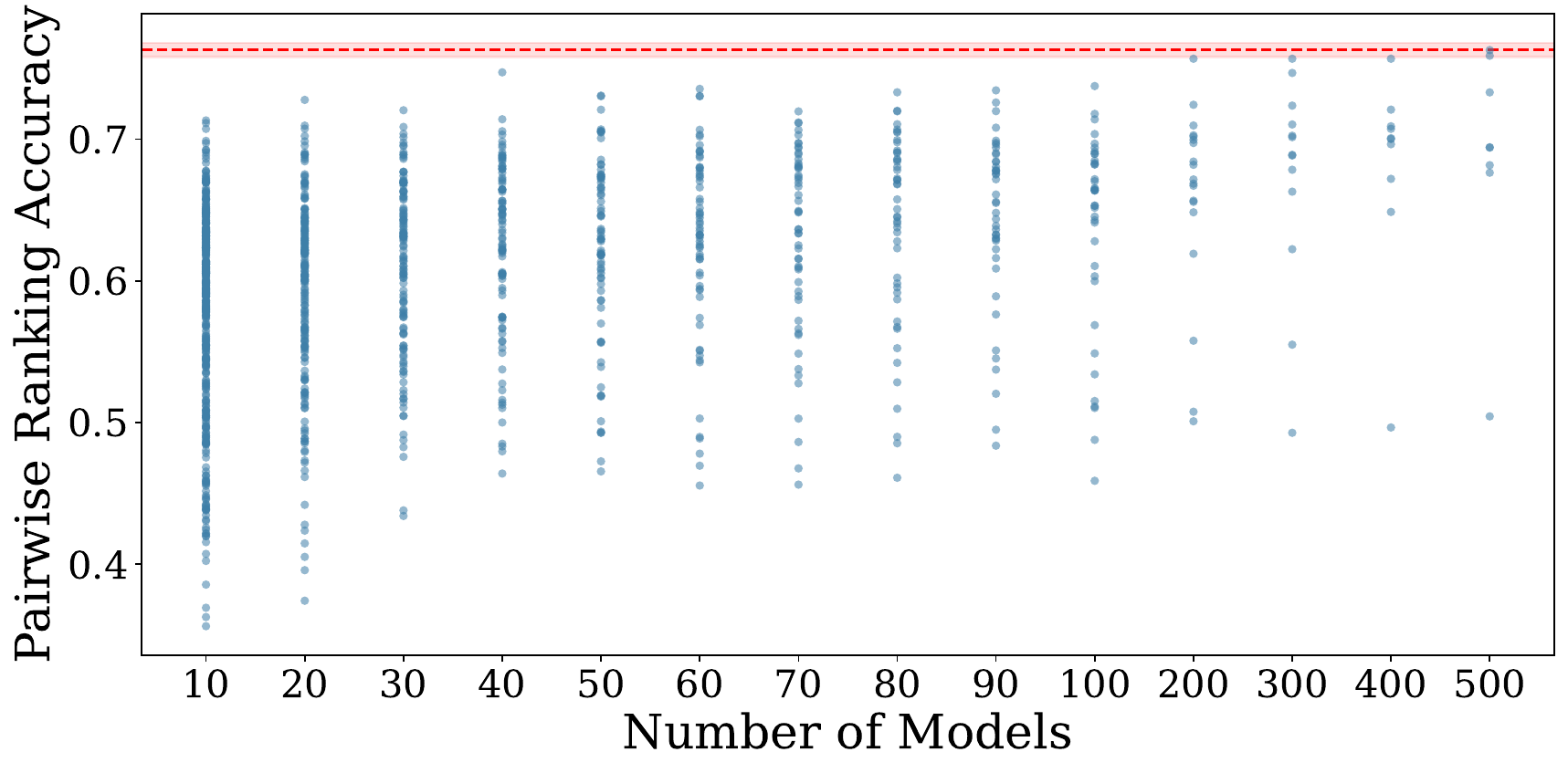}
        \caption{Pairwise Ranking Accuracy ($Acc_{pair}$)}
    \end{subfigure}
    \caption{$\text{DPA}$ - Target 30B+. Aligned benchmarks created with different numbers of models. In (a), each point shows the Spearman Rho for one subset of models. In (b), each point shows the Pairwise Ranking Accuracy ($Acc_{pair}$). Error bars obtained using the sample size from the target set. The dotted line represents the results of alignment using all models.}
    \label{fig:bench_align_opt_num_models_30b_honesty_rm2}
\end{figure*}

\begin{figure*}[t]
    \centering
    \begin{subfigure}[b]{0.49\linewidth}
        \centering
        \includegraphics[width=\linewidth]{images/model_window_best_exps/dpa-ultrafeedback-honesty/30b_70b/pdfs/dpa-ultrafeedback-honesty_30b_70b_rho_best_vs_window.pdf}
        \caption{Spearman Rho ($\rho$)}
    \end{subfigure}
    \hfill
    \begin{subfigure}[b]{0.49\linewidth}
        \centering
        \includegraphics[width=\linewidth]{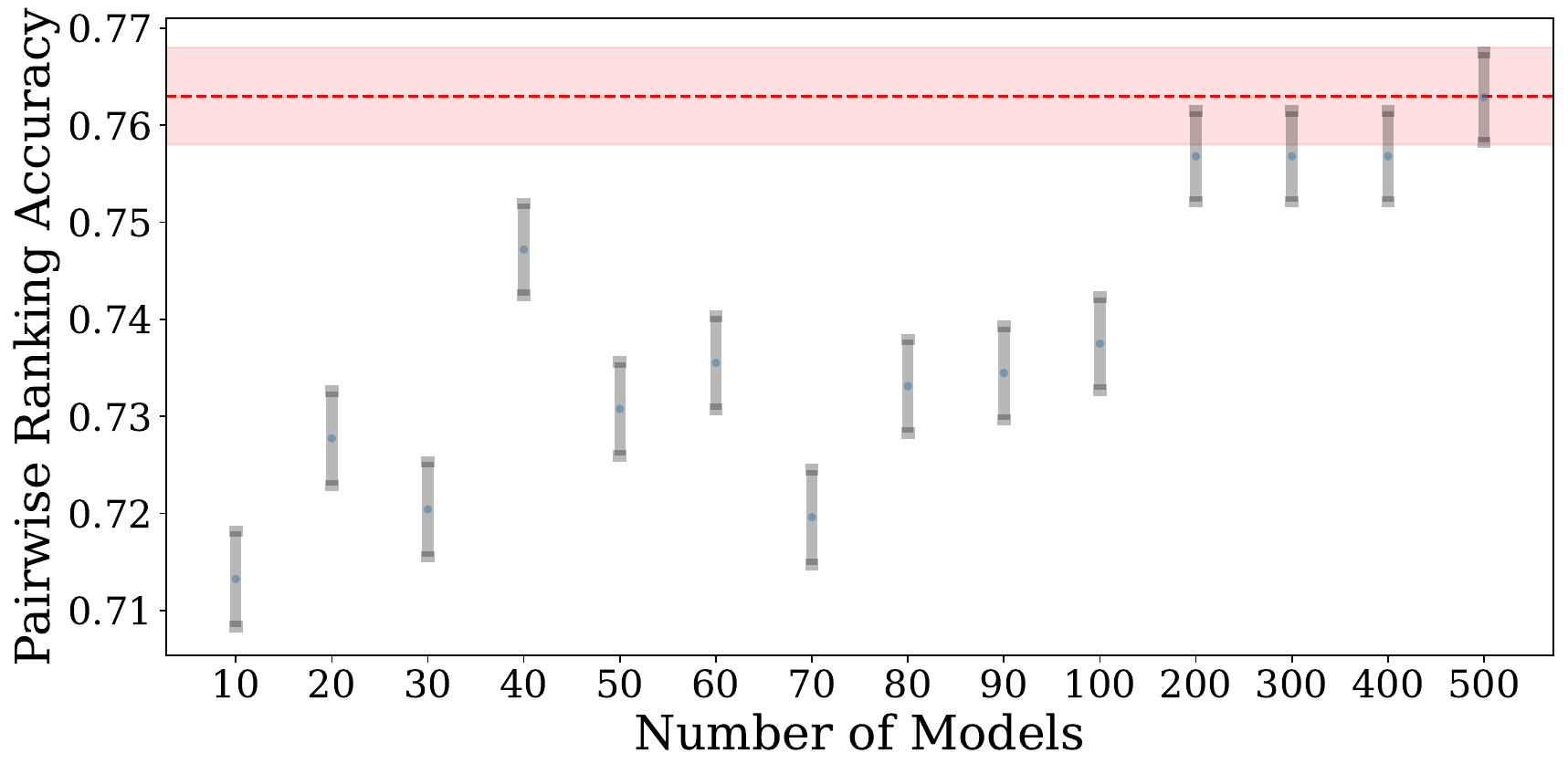}
        \caption{Pairwise Ranking Accuracy ($Acc_{pair}$)}
    \end{subfigure}
    \caption{$\text{DPA}$ - Target 30B+. Aligned benchmarks created with different numbers of models. In (a), each point reports the Spearman Rho ($\rho$) of the best subset of models. In (b), each point shows the Pairwise Ranking Accuracy ($Acc_{pair}$). Error bars obtained using the sample size from the target set. The dotted line represents the results of alignment using all models.}
    \label{fig:bench_align_opt_num_models_30b_honesty_rm2}
\end{figure*}

\xhdr{Target: 70B+} Figures \ref{fig:bench_align_opt_num_models_70b_honesty_rm1} and \ref{fig:bench_align_opt_num_models_70b_honesty_rm1} represent the results of all subset of models and the best subset by number of models, respectively.

\begin{figure*}[t]
    \centering
    \begin{subfigure}[b]{0.49\linewidth}
        \centering
        \includegraphics[width=\linewidth]{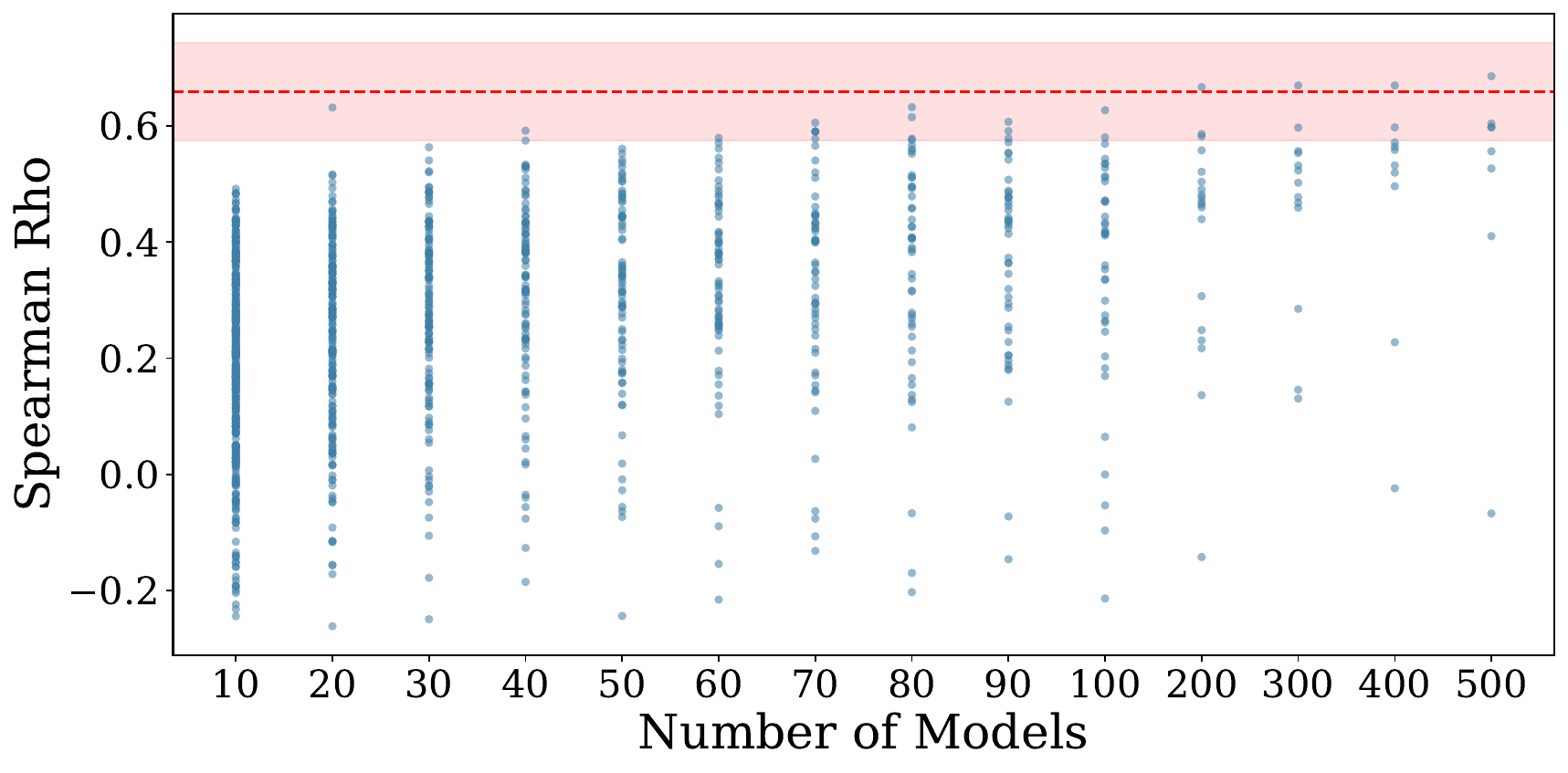}
        \caption{Spearman Rho ($\rho$)}
    \end{subfigure}
    \hfill
    \begin{subfigure}[b]{0.49\linewidth}
        \centering
        \includegraphics[width=\linewidth]{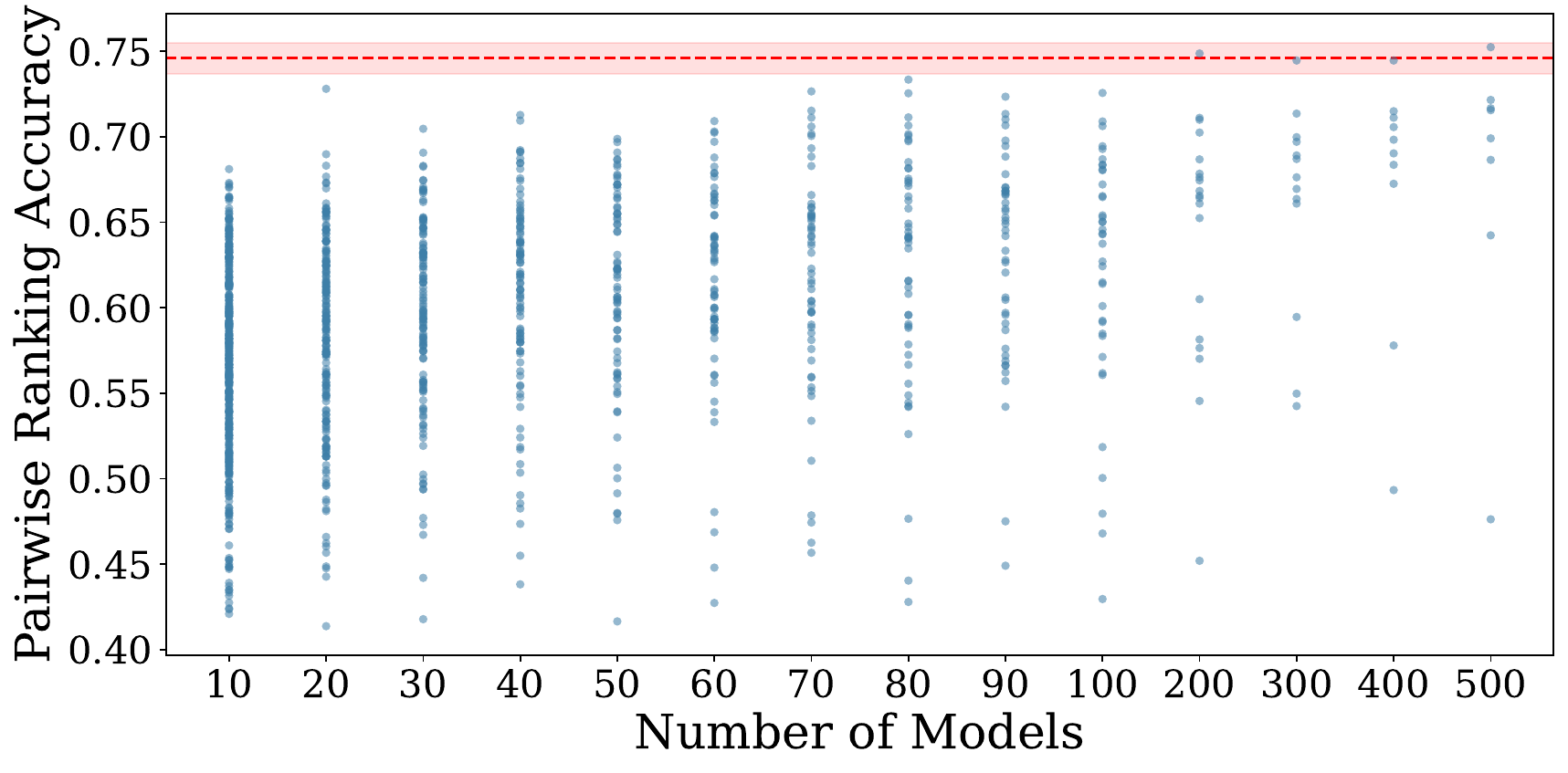}
        \caption{Pairwise Ranking Accuracy ($Acc_{pair}$)}
    \end{subfigure}
    \caption{$\text{DPA}$ - Target 70B+. Aligned benchmarks created with different numbers of models. In (a), each point shows the Spearman Rho for one subset of models. In (b), each point shows the Pairwise Ranking Accuracy ($Acc_{pair}$). Error bars obtained using the sample size from the target set. The dotted line represents the results of alignment using all models.}
    \label{fig:bench_align_opt_num_models_70b_honesty_rm2}
\end{figure*}

\begin{figure*}[t]
    \centering
    \begin{subfigure}[b]{0.49\linewidth}
        \centering
        \includegraphics[width=\linewidth]{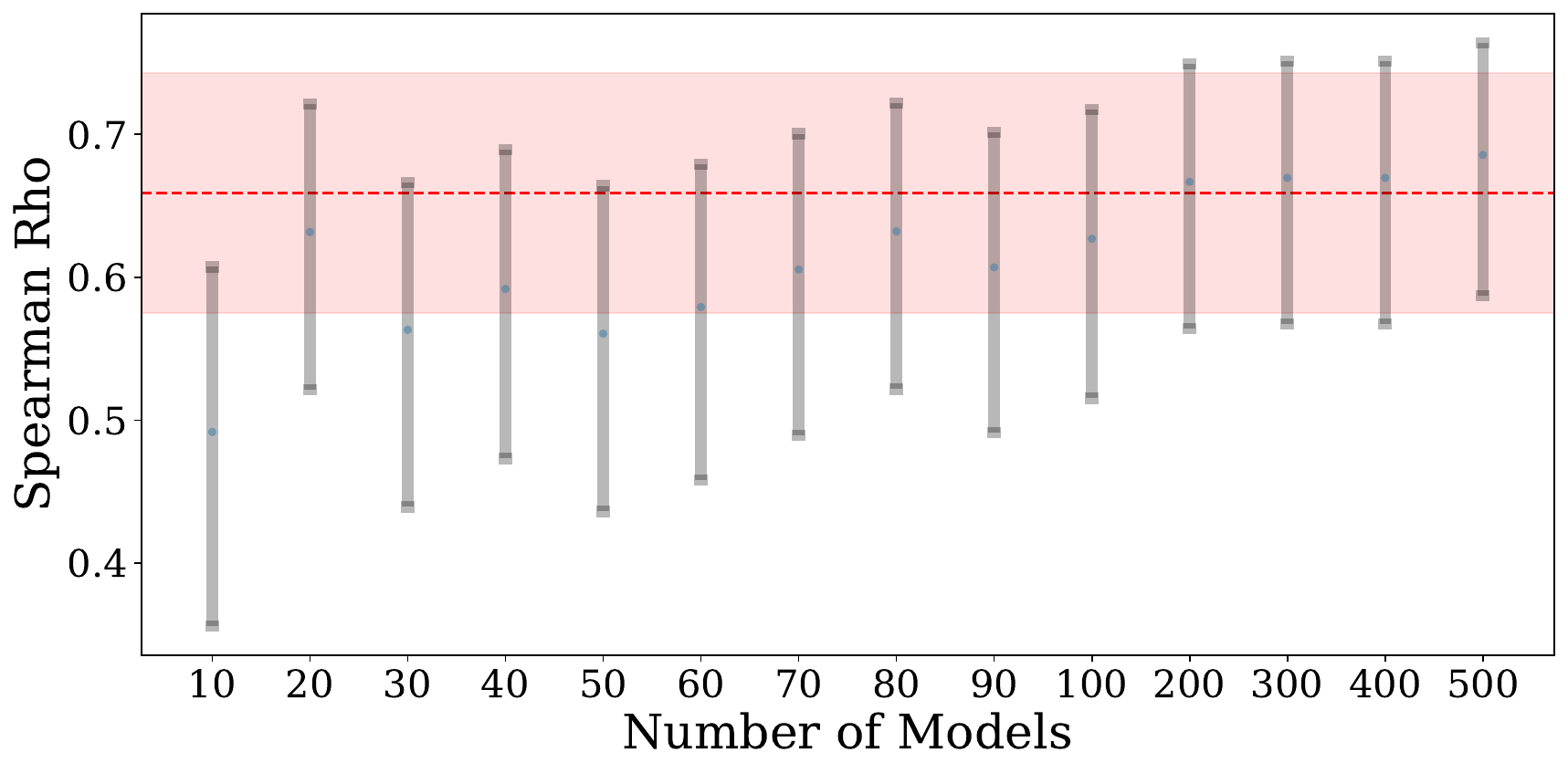}.
        \caption{Spearman Rho ($\rho$)}
    \end{subfigure}
    \hfill
    \begin{subfigure}[b]{0.49\linewidth}
        \centering
        \includegraphics[width=\linewidth]{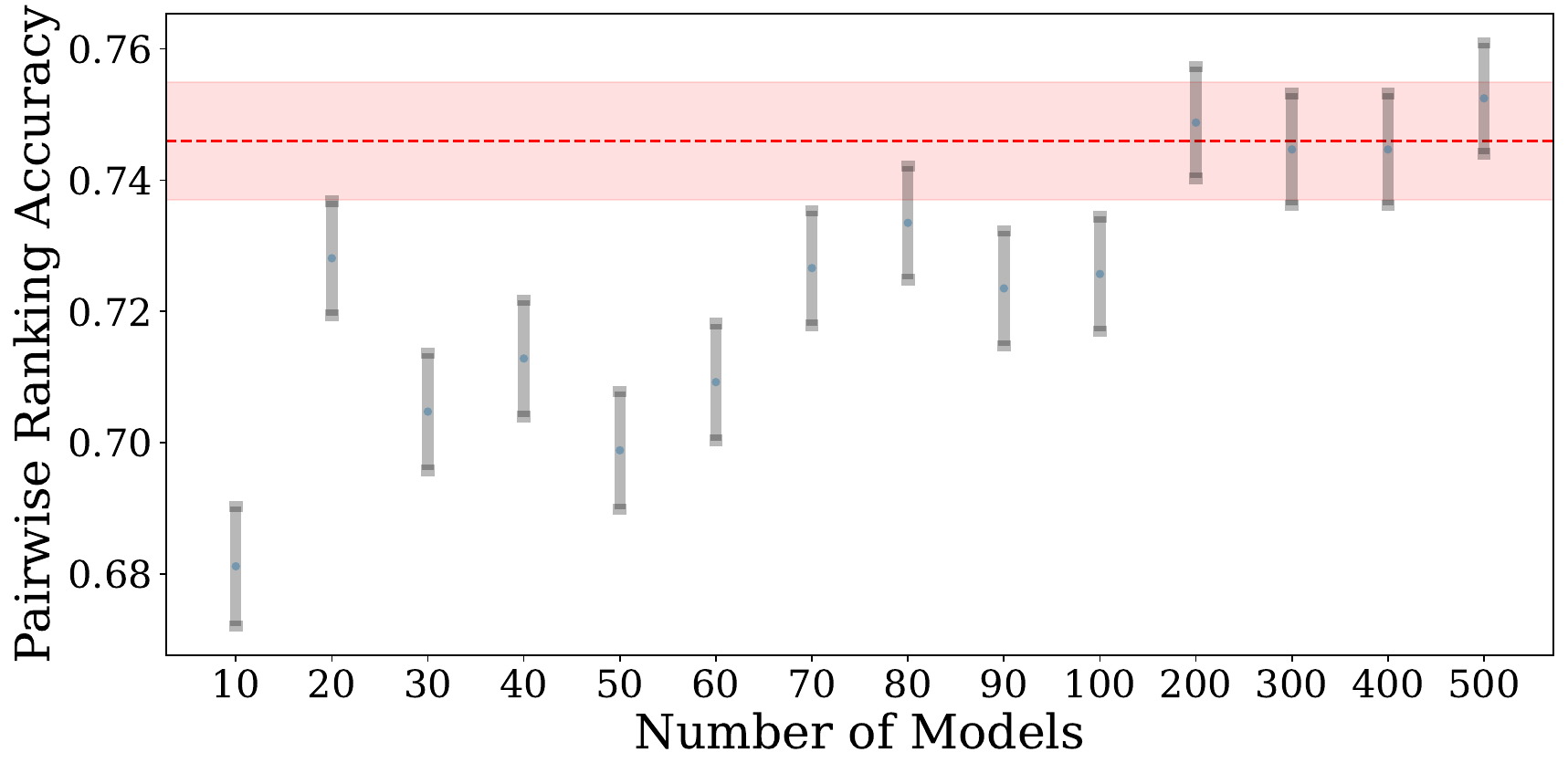}
        \caption{Pairwise Ranking Accuracy ($Acc_{pair}$)}
    \end{subfigure}
    \caption{$\text{DPA}$ - Target 70B+. Aligned benchmarks created with different numbers of models. In (a), each point reports the Spearman Rho ($\rho$) of the best subset of models. In (b), each point shows the Pairwise Ranking Accuracy ($Acc_{pair}$). Error bars obtained using the sample size from the target set. The dotted line represents the results of alignment using all models.}
    \label{fig:bench_align_opt_num_models_70b_honesty_rm2}
\end{figure*}

\subsection{Number of Questions}\label{ap:opt_num_questions}
We extend the analysis from Section~\ref{sec:perf_num_questions} by evaluating \textsc{BenchAlign} in combination with \textsc{TinyBenchmarks} and \textsc{MetaBench} on every preference and target considered in Section \ref{sec:perf_model_size_splits} and Appendix \ref{ap:pref_heterog}. Both approaches probe how few question items are sufficient to retain alignment, but they do so at very different scales: \textsc{TinyBenchmarks} keeps a fixed set of $100$ items per benchmark (around $5000$ items in total across all OpenLLMLeaderboard benchmarks), while \textsc{MetaBench} selects only $1500$ items in total across all of the benchmarks.

\xhdr{Model scale splits} Table~\ref{tab:numq_model_size_splits_helpsteer_ultrafeedback} reports for all model-scale splits. Distillation-plus-alignment combinations remain far above the \textsc{No Alignment} baseline at every target. Simultaneously, the gap between using all questions for alignment and the two distillation variants narrows as the target grows: at $70$B+, \textsc{BenchAlign} achieves the best results in all columns, but at $13$B+ \textsc{TinyBenchmarks+BenchAlign} matches or outperforms all baselines in most cases using around $5000$ items. These results suggest that the small fraction of questions obtained by benchmark distillation contain sufficient preference signals to enable alignment.

\begin{table*}[t]
\centering
\small
\setlength{\tabcolsep}{5pt}
\renewcommand{\arraystretch}{1.05}

\resizebox{\textwidth}{!}{
\begin{tabular}{c c c c c c  c c c c}
\toprule
 \multirow{4}{*}{\makecell{\textbf{Target}\\\textbf{Models}}} & \multirow{4}{*}{\textbf{Method}} & \multicolumn{4}{c }{\textbf{Helpfulness}} & \multicolumn{4}{c}{\textbf{Honesty}} \\
\cmidrule(lr){3-6}  \cmidrule(lr){7-10}

& & \multicolumn{2}{c}{$Acc_{pair}$}
& \multicolumn{2}{c}{$\rho$}
& \multicolumn{2}{ c}{$Acc_{pair}$}
& \multicolumn{2}{c}{$\rho$} \\
\cmidrule(lr){3-4}  \cmidrule(lr){5-6} \cmidrule(lr){7-8}  \cmidrule(lr){9-10} 

& & $\text{ArmoRM}$ & $\text{GPT2}$ & $\text{ArmoRM}$ & $\text{GPT2}$ & $\text{ArmoRM}$ & $\text{DPA}$ & $\text{ArmoRM}$ & $\text{DPA}$ \\ \midrule

\multirow{4}{*}{70B+}& \textsc{No Alignment}& $0.542\pm0.010$& $0.480\pm0.010$& $0.154\pm0.165$& $0.044\pm0.251$& $0.559\pm0.010$& $0.567\pm0.010$& $0.199\pm0.164$&$0.188\pm0.164$ \\
& \textsc{MetaBench + BenchAlign} & 
$0.709\pm0.009$ & $0.594\pm0.010$ & $0.570\pm0.102$ & $0.274\pm0.146$ & $0.699\pm0.009$& $0.668\pm0.009$& $0.548\pm0.106$ & $0.477\pm0.118$\\
& \textsc{TinyBenchmarks + BenchAlign} & $0.725\pm0.009$ & $0.585\pm0.010$ & $0.594\pm0.097$ & $0.244\pm0.149$ & $0.733\pm0.009$ & $0.702\pm0.009$ & $0.622\pm0.092$ & $0.555\pm0.104$ \\
& \textsc{BenchAlign} & $\mathbf{0.778 \pm 0.008}$& $\mathbf{0.620 \pm 0.010}$& $\mathbf{0.707 \pm 0.074}$& $\mathbf{0.333 \pm 0.139}$& $\mathbf{0.774 \pm 0.008}$& $\mathbf{0.746 \pm 0.009}$& $\mathbf{0.706 \pm 0.074}$& $\mathbf{0.659 \pm 0.084}$\\
\midrule

\multirow{4}{*}{30B+}& \textsc{No Alignment}& $0.605\pm0.005$& $0.477\pm0.005$& $0.331\pm0.114$& $0.052\pm0.226$& $0.618\pm0.005$& $0.609\pm0.005$& $0.368\pm0.111$&$0.319\pm0.115$ \\
& \textsc{MetaBench + BenchAlign} & 
$0.711\pm0.005$ & $0.564\pm0.005$ & $0.585\pm0.075$ & $0.191\pm0.116$ & $0.725\pm0.005$ & $0.720\pm0.005$ &  $0.622\pm0.070$ & $0.608\pm0.072$\\
& \textsc{TinyBenchmarks + BenchAlign} & $0.749\pm0.005$ & $0.612\pm0.005$ & $0.666\pm0.063$ & $0.320\pm0.106$ & $0.764\pm0.005$ & $0.756\pm0.005$ & $0.706\pm0.057$ & $0.694\pm0.059$ \\
& \textsc{BenchAlign} & $\mathbf{0.765 \pm 0.005}$& $\mathbf{0.637 \pm 0.005}$& $\mathbf{0.710 \pm 0.056}$& $\mathbf{0.387 \pm 0.100}$&  $\mathbf{0.773 \pm 0.005}$&  $\mathbf{0.763 \pm 0.005}$&  $\mathbf{0.730 \pm 0.053}$&  $\mathbf{0.713 \pm 0.055}$\\
\midrule

\multirow{4}{*}{13B+}& \textsc{No Alignment}& $0.583\pm0.002$& $0.531\pm0.002$& $0.245\pm0.065$& $0.094\pm0.067$& $0.588\pm0.002$& $0.577\pm0.002$& $0.251\pm0.064$&$0.231\pm0.065$ \\
& \textsc{MetaBench + BenchAlign} & 
$0.744\pm0.001$& $0.653\pm0.002$& $0.677\pm0.035$& $0.442\pm0.053$&  
$0.757\pm0.001$&  $0.750\pm0.001$ & $0.706\pm0.032$ & $0.695\pm0.033$\\
& \textsc{TinyBenchmarks + BenchAlign} & $\mathbf{0.754\pm0.001}$ & $0.677\pm0.002$ & $\mathbf{0.700\pm0.033}$ & $0.505\pm0.049$ & $\mathbf{0.770\pm0.001}$ & $\mathbf{0.770\pm0.001}$ & $\mathbf{0.735\pm0.030}$ & $\mathbf{0.741\pm0.029}$ \\
& \textsc{BenchAlign} & $0.741 \pm 0.001$& $\mathbf{0.696 \pm 0.002}$& $0.674 \pm 0.035$& $\mathbf{0.552 \pm 0.045}$&  $0.748 \pm 0.001$ & $0.749 \pm 0.001$&  $0.686 \pm 0.034$&  $0.691 \pm 0.034$\\
\bottomrule
\end{tabular}
}
\caption{Alignment with reduced question pools under model scale splits.}
\label{tab:numq_model_size_splits_helpsteer_ultrafeedback}
\end{table*}

\xhdr{Heterogeneous preferences}
Table~\ref{tab:eval_heterogeneous_distill} shows that benchmark alignment plus distillation under heterogeneous preferences shows similar patterns to the ones observed in Table \ref{tab:numq_model_size_splits_helpsteer_ultrafeedback}. Both distillation-plus-alignment variants recover most of the gains provided by full \textsc{BenchAlign} over \textsc{No Alignment}. \textsc{Metabench+BenchAlign} improves $Acc_{pair}$ by roughly $14$--$19$ points and raises $\rho$ to the $0.6$--$0.8$ range across all targets, despite using only $1{,}500$ items. \textsc{TinyBenchmarks+BenchAlign} is still the best alternative as it consistently achieves alignment with a reduced number of questions. Overall, the carefully selected subsets recover nearly all of the alignment gains achieved using the full item pool under heterogenous preferences.

\begin{table*}[t]
\centering
\small
\setlength{\tabcolsep}{5pt}
\renewcommand{\arraystretch}{1.05}

\resizebox{\textwidth}{!}{
\begin{tabular}{c c c c c c  |c c c c}
\toprule
\multirow{2}{*}{Target}& \multirow{2}{*}{Method}& \multicolumn{4}{c}{$Acc_{pair}$}& \multicolumn{4}{c}{$\rho$}\\
\cline{3-10}& & $RM_{1}^{Judge}$& $RM_{2}^{Judge}$& $RM_{3}^{Judge}$& $RM_{4}^{Judge}$& $RM_{1}^{Judge}$& $RM_{2}^{Judge}$& $RM_{3}^{Judge}$& $RM_{4}^{Judge}$\\
\midrule

\multirow{4}{*}{2\%}
& \textsc{No Alignment}& $0.578\pm0.001$ & $0.743\pm0.007$& $0.638\pm0.015$& $0.684\pm0.015$ & $0.272\pm0.207$& $0.700\pm0.122$& $0.390\pm0.191$ &$0.511\pm0.174$\\
& \textsc{Metabench + BenchAlign}& $0.722\pm0.014$ & $0.808\pm0.013$ & $0.813\pm0.012$ & $0.747\pm0.014$ & $0.601\pm0.120$ & $0.801\pm0.065$ & $0.812\pm0.061$ & $0.648\pm0.107$\\
& \textsc{TinyBenchmarks + BenchAlign}& $0.750\pm0.014$ & $0.822\pm0.012$ & $0.819\pm0.012$ & $0.753\pm0.014$ & $0.676\pm0.100$ & $0.830\pm0.056$ & $0.820\pm0.059$ & $0.675\pm0.100$\\
& \textsc{BenchAlign}& $\mathbf{0.768\pm0.014}$ & $\mathbf{0.842\pm0.012}$ & $\mathbf{0.820\pm0.012}$ & $\mathbf{0.773\pm0.013}$ & $\mathbf{0.698\pm0.127}$ & $\mathbf{0.855\pm0.065}$ & $\mathbf{0.824\pm0.082}$ & $\mathbf{0.693\pm0.128}$\\
\midrule

\multirow{4}{*}{5\%}
& \textsc{No Alignment} & $0.559\pm0.001$& $0.713\pm0.006$& $0.610\pm0.006$& $0.616\pm0.006$& $0.193\pm0.131$& $0.606\pm0.091$& $0.327\pm0.124$&$0.336\pm0.123$\\
& \textsc{Metabench + BenchAlign}& $0.717\pm0.006$ & $0.826\pm0.005$ & $0.779\pm0.005$ & $0.760\pm0.005$ & $0.597\pm0.079$ & $0.836\pm0.036$ & $0.744\pm0.054$ & $0.687\pm0.064$\\
& \textsc{TinyBenchmarks + BenchAlign}& $0.760\pm0.005$ & $0.841\pm0.005$ & $0.808\pm0.005$ & $\mathbf{0.803\pm0.005}$ & $0.694\pm0.063$ & $0.859\pm0.031$ & $0.801\pm0.043$ & $\mathbf{0.781\pm0.047}$\\
& \textsc{BenchAlign}& $\mathbf{0.770\pm0.005}$ & $\mathbf{0.850\pm0.005}$ & $\mathbf{0.811\pm0.005}$ & $\mathbf{0.803\pm0.005}$ & $\mathbf{0.705\pm0.074}$ & $\mathbf{0.873\pm0.036}$ & $\mathbf{0.806\pm0.052}$ & $0.770\pm0.060$\\
\midrule

\multirow{4}{*}{10\%}
& \textsc{No Alignment}& $0.563\pm0.003$& $0.713\pm0.003$& $0.605\pm0.003$& $0.626\pm0.003$& $0.184\pm0.094$& $0.601\pm0.000$& $0.314\pm0.087$&$0.363\pm0.084$\\
& \textsc{Metabench + BenchAlign}& $0.752\pm0.003$ & $0.826\pm0.002$ & $0.793\pm0.003$ & $0.772\pm0.003$ & $0.687\pm0.047$ & $0.840\pm0.026$ & $0.774\pm0.035$ & $0.724\pm0.042$\\
& \textsc{TinyBenchmarks + BenchAlign}& $0.772\pm0.003$ & $0.846\pm0.002$ & $0.810\pm0.002$ & $\mathbf{0.808\pm0.003}$ & $0.731\pm0.041$ & $0.872\pm0.021$ & $0.810\pm0.030$ & $\mathbf{0.800\pm0.031}$\\
& \textsc{BenchAlign}& $\mathbf{0.782\pm0.003}$ & $\mathbf{0.850\pm0.002}$ & $\mathbf{0.814\pm0.002}$ & $\mathbf{0.808\pm0.003}$ & $\mathbf{0.746\pm0.045}$ & $\mathbf{0.875\pm0.024}$ & $\mathbf{0.814\pm0.034}$ & $0.796\pm0.037$\\
\bottomrule
\end{tabular}
}
\caption{Alignment with reduced question pools under heterogeneous preferences}
\label{tab:eval_heterogeneous_distill}
\end{table*}

\xhdr{Arbitrary model splits} To show the robustness of our findings further, we repeat our distillation experiments using the same targets as \ref{sec:perf_model_size_splits} but under arbitrary splits. Table~\ref{tab:random_splits_distill_benchalign} shows the results across all target models and preferences, \textsc{TinyBenchmarks+BenchAlign} matches full \textsc{BenchAlign} within confidence intervals on both $Acc_{pair}$ and $\rho$, and occasionally slightly surpasses it. \textsc{Metabench+BenchAlign}, despite using only around $1500$ items, remains within roughly $3$--$5$ points of \textsc{BenchAlign}$'$s accuracy while substantially outperforming the \textsc{No Alignment} baseline on both metrics. Together with the model-scale and heterogeneous-preference experiments, these results indicate that benchmark alignment does not require access to the a large number of test items to be performed as long as we know in advance which of them contain the largest preference signals.

\begin{table*}[t]
\centering
\small
\setlength{\tabcolsep}{5pt}
\renewcommand{\arraystretch}{1.05}

\resizebox{\textwidth}{!}{
\begin{tabular}{c c c c c c  c c c c}
\toprule
 \multirow{4}{*}{\makecell{\textbf{Target}\\\textbf{Models}}} & \multirow{4}{*}{\textbf{Method}} & \multicolumn{4}{c }{\textbf{Helpfulness}} & \multicolumn{4}{c}{\textbf{Honesty}} \\
\cmidrule(lr){3-6}  \cmidrule(lr){7-10}

& & \multicolumn{2}{c}{$Acc_{pair}$}
& \multicolumn{2}{c}{$\rho$}
& \multicolumn{2}{ c}{$Acc_{pair}$}
& \multicolumn{2}{c}{$\rho$} \\
\cmidrule(lr){3-4}  \cmidrule(lr){5-6} \cmidrule(lr){7-8}  \cmidrule(lr){9-10} 

& & $\text{ArmoRM}$ & $\text{GPT2}$ & $\text{ArmoRM}$ & $\text{GPT2}$ & $\text{ArmoRM}$ & $\text{DPA}$ & $\text{ArmoRM}$ & $\text{DPA}$ \\ \midrule

\multirow{4}{*}{2\%}
& \textsc{No Alignment}& $0.720\pm0.014$& $0.633\pm0.015$& $0.614\pm0.151$& $0.383\pm0.196$& $0.736\pm0.014$& $0.724\pm0.014$& $0.652\pm0.140$&$0.627\pm0.147$\\
& \textsc{Metabench + BenchAlign}& $0.831\pm0.012$ & $0.779\pm0.013$ & $0.842\pm0.052$ & $0.741\pm0.082$ & $0.852\pm0.011$ & $0.842\pm0.012$ & $0.872\pm0.043$ & $0.863\pm0.045$\\
& \textsc{TinyBenchmarks + BenchAlign}& $\mathbf{0.869\pm0.011}$ & $\mathbf{0.811\pm0.013}$ & $\mathbf{0.895\pm0.035}$ & $0.796\pm0.066$ & $0.881\pm0.010$ & $0.874\pm0.011$ & $0.915\pm0.029$ & $0.910\pm0.030$\\
& \textsc{BenchAlign}& $0.866\pm0.011$ & $0.807\pm0.013$ & $\mathbf{0.895\pm0.035}$ & $\mathbf{0.800\pm0.078}$ & $\mathbf{0.891\pm0.010}$ & $\mathbf{0.891\pm0.010}$ & $\mathbf{0.933\pm0.028}$ & $\mathbf{0.929\pm0.030}$\\
\midrule

\multirow{4}{*}{5\%}
& \textsc{No Alignment}& $0.708\pm0.006$& $0.605\pm0.006$& $0.580\pm0.095$& $0.314\pm0.125$& $0.727\pm0.006$& $0.722\pm0.006$& $0.630\pm0.088$&$0.616\pm0.090$\\
& \textsc{Metabench + BenchAlign}& $0.808\pm0.005$ & $0.746\pm0.006$ & $0.802\pm0.043$ & $0.673\pm0.067$ & $0.827\pm0.005$ & $0.823\pm0.005$ & $0.843\pm0.035$ & $0.834\pm0.037$\\
& \textsc{TinyBenchmarks + BenchAlign}& $\mathbf{0.851\pm0.005}$ & $0.791\pm0.005$ & $\mathbf{0.867\pm0.030}$ & $\mathbf{0.764\pm0.050}$ & $0.864\pm0.004$ & $0.863\pm0.004$ & $0.899\pm0.023$ & $0.897\pm0.023$\\
& \textsc{BenchAlign}& $0.849\pm0.005$ & $\mathbf{0.792\pm0.005}$ & $0.866\pm0.034$ & $0.762\pm0.056$ & $\mathbf{0.870\pm0.004}$ & $\mathbf{0.867\pm0.004}$ & $\mathbf{0.905\pm0.024}$ & $\mathbf{0.899\pm0.026}$\\
\midrule

\multirow{4}{*}{10\%}
& \textsc{No Alignment}& $0.678\pm0.003$& $0.590\pm0.003$& $0.502\pm0.074$& $0.264\pm0.090$& $0.697\pm0.003$& $0.695\pm0.003$& $0.558\pm0.068$&$0.551\pm0.069$\\
& \textsc{Metabench + BenchAlign}& $0.820\pm0.002$ & $0.769\pm0.003$ & $0.825\pm0.028$ & $0.713\pm0.043$ & $0.838\pm0.002$ & $0.836\pm0.002$ & $0.859\pm0.023$ & $0.857\pm0.023$\\
& \textsc{TinyBenchmarks + BenchAlign}& $\mathbf{0.855\pm0.002}$ & $0.808\pm0.003$ & $\mathbf{0.876\pm0.020}$ & $0.793\pm0.032$ & $\mathbf{0.868\pm0.002}$ & $\mathbf{0.866\pm0.002}$ & $\mathbf{0.903\pm0.016}$ & $\mathbf{0.902\pm0.016}$\\
& \textsc{BenchAlign}& $\mathbf{0.855\pm0.002}$ & $\mathbf{0.814\pm0.002}$ & $\mathbf{0.876\pm0.020}$ & $\mathbf{0.804\pm0.033}$ & $0.863\pm0.002$ & $0.862\pm0.002$ & $0.894\pm0.019$ & $0.891\pm0.019$\\
\bottomrule
\end{tabular}
}
\caption{Alignment with reduced question pools under arbitrary model sets}
\label{tab:random_splits_distill_benchalign}
\end{table*}

\clearpage

\section{Finding the Best Subset of Models}\label{ap:best_subset_models}

\subsection{Model Family}

We study whether the model family\footnote{We consider a model family as the one obtained from the HuggingFace metadata entry with the same name.} drives alignment by performing an experiment inspired by the most common architectures found on each target by model scale. We select the top 5 families from each target and created a new group of train/test splits that allow us to test whether the model family is relevant for alignment:

\begin{itemize}
    \item Llama
    \item Qwen2
\end{itemize}

\begin{figure*}[t]
    \centering
    \begin{subfigure}[b]{0.49\linewidth}
        \centering
        \includegraphics[width=\linewidth]{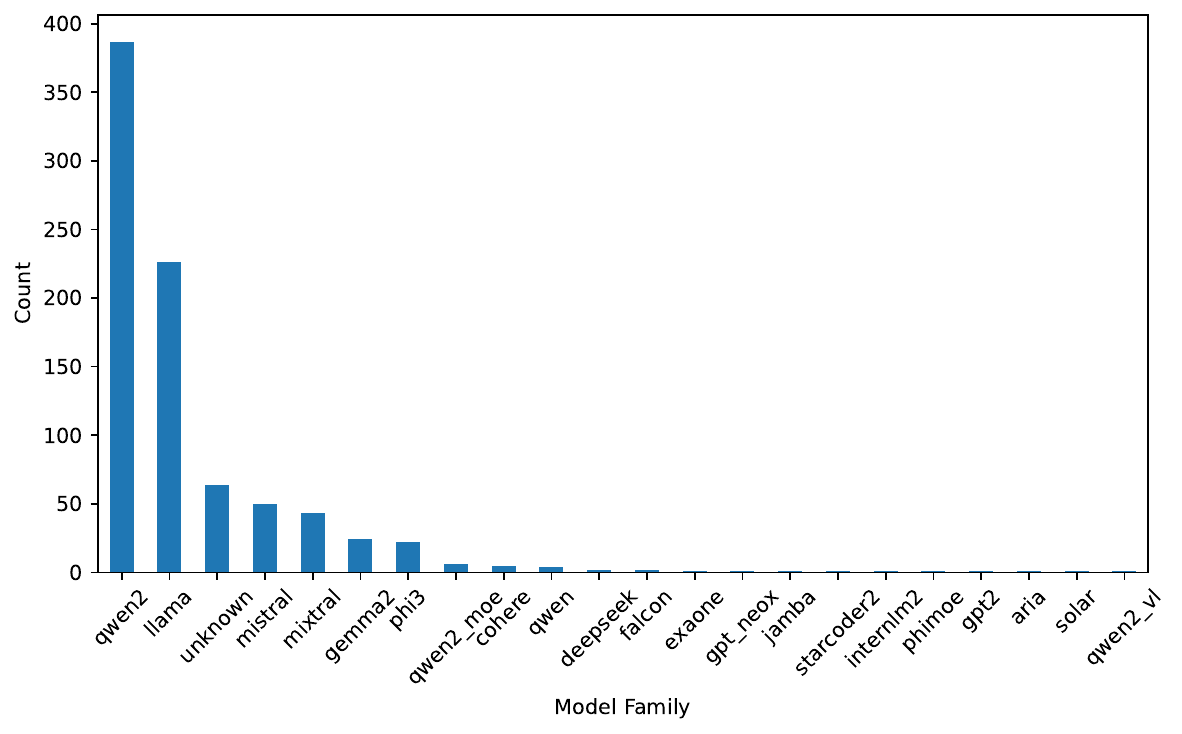}
        \caption{Target 13B+}
    \end{subfigure}
    \hfill
    \begin{subfigure}[b]{0.49\linewidth}
        \centering
        \includegraphics[width=\linewidth]{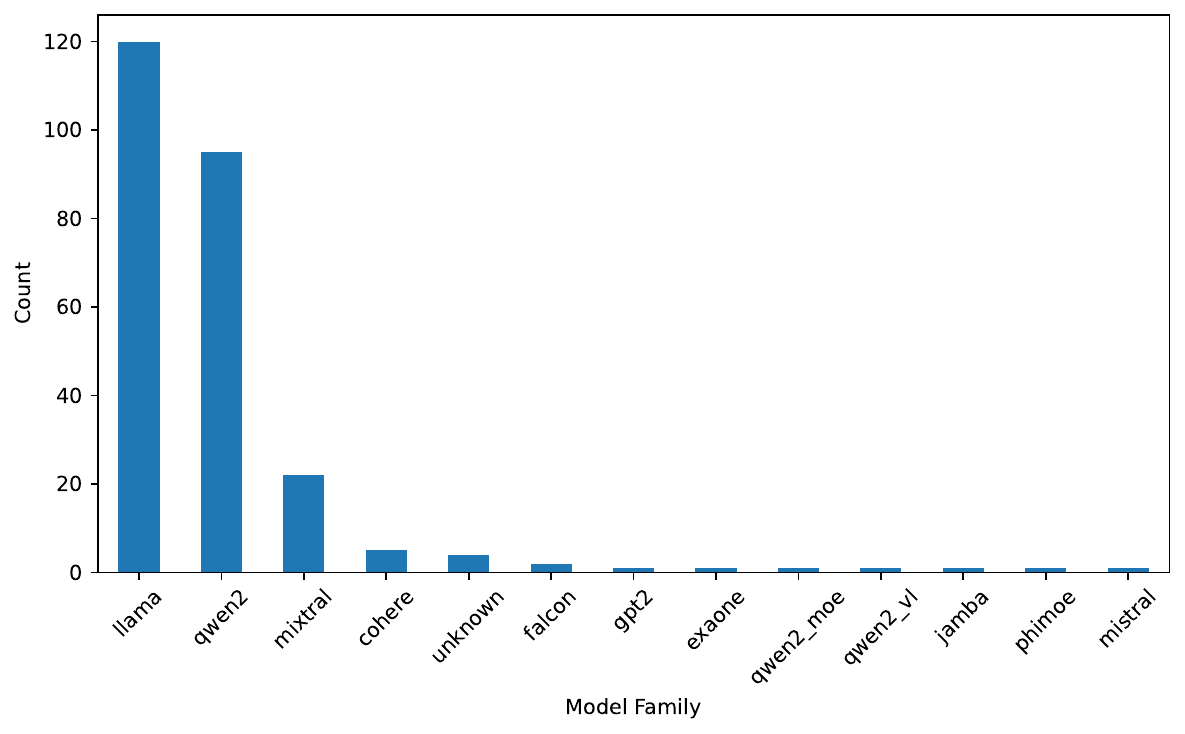}
        \caption{Target 30B+}
    \end{subfigure}
    \hfill
    \begin{subfigure}[b]{0.49\linewidth}
        \centering
        \includegraphics[width=\linewidth]{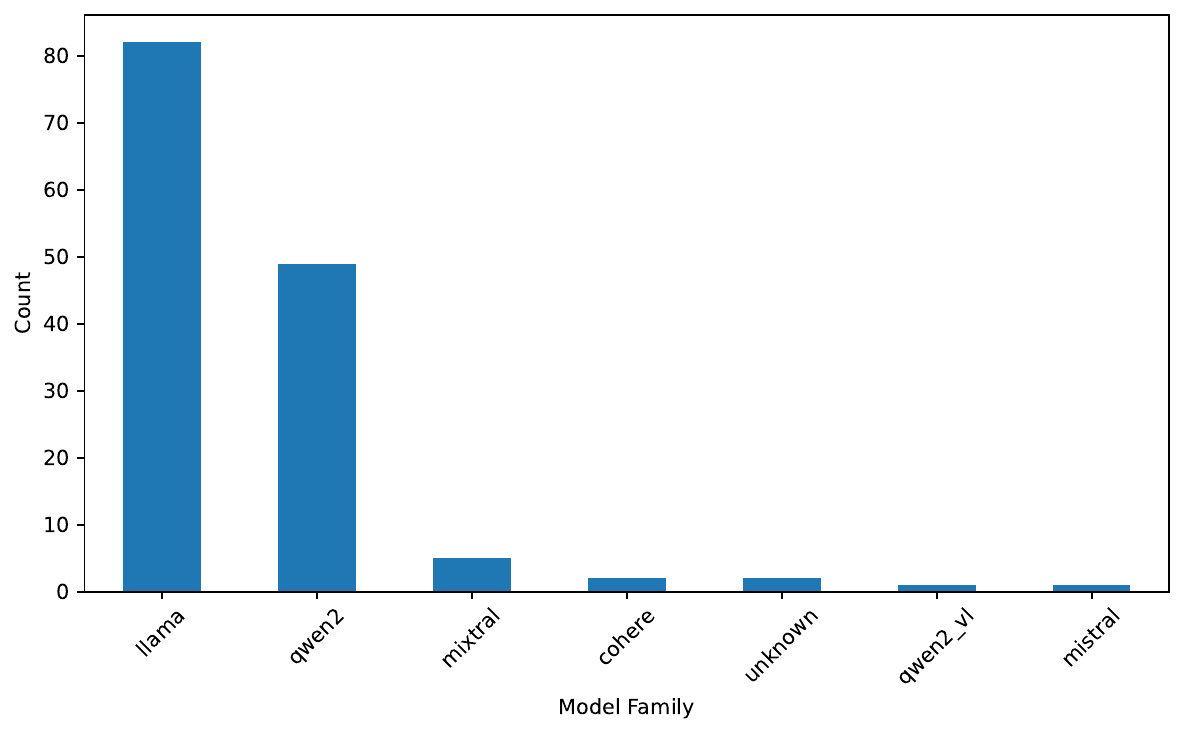}
        \caption{Target 70B+}
    \end{subfigure}
    \caption{Distribution of models by model family across model size targets}
    \label{fig:dist_msize_family}
\end{figure*}

The test samples consisted of all Llama or Qwen models found in the target set as they are the most common across all target sets (See Figure \ref{fig:dist_msize_family}). Then, we find the top 5 biggest families on train:

\begin{itemize}
    \item Qwen2
    \item Llama
    \item Mistral
    \item Gemma
    \item Phi
\end{itemize}

We randomly sample models from each of these 5 groups making sure they all share the same size to ensure all had the same number of models\footnote{We fixed this number as 64 as that is the number of Phi models, the smallest group from the top 5}. Finally, we evaluate for all reward models from Section \ref{sec:perf_model_size_splits} whether the best family for alignment is the same family as the target.

Table \ref{tab:model_fam_exps} shows the results of our experiments. Surprisingly, the model family with the best results coincides with the target in 45.83\% of the experiments. In the Llama experiments, this coincidence happens 41.6\% of the time, while 50\% of the times in the Qwen experiments. This results suggest that the choice of model family for experiments is not very different from choosing a family at random.

\begin{table}
    \centering
    \caption{Best model families in family target experiments}
    \begin{tabular}{cccc}\toprule
         RM&  Target&  Train Family& Test Family\\\midrule
         $\text{ArmoRM}$ (Helpful)&  13B+&  qwen& llama\\
         $\text{ArmoRM}$ (Helpful)&  30B+&  qwen& llama\\
         $\text{ArmoRM}$ (Helpful)&  70B+&  llama& llama\\
         $\text{ArmoRM}$ (Helpful)&  13B+&  mistral& qwen\\
         $\text{ArmoRM}$ (Helpful)&  30B+&  qwen& qwen\\
         $\text{ArmoRM}$ (Helpful)&  70B+&  qwen& qwen\\
         $\text{GPT2}$&  13B+&  llama& llama\\
         $\text{GPT2}$&  30B+&  qwen& llama\\
         $\text{GPT2}$& 70B+& gemma&llama\\
         $\text{GPT2}$& 13B+& llama&qwen\\
         $\text{GPT2}$& 30B+& qwen&qwen\\
         $\text{GPT2}$& 70B+& qwen&qwen\\
         $\text{ArmoRM}$ (Honest)&  13B+&  llama& llama\\
         $\text{ArmoRM}$ (Honest)&  30B+&  qwen& llama\\
         $\text{ArmoRM}$ (Honest)&  70B+&  mistral& llama\\
         $\text{ArmoRM}$ (Honest)&  13B+&  mistral& qwen\\
         $\text{ArmoRM}$ (Honest)&  30B+&  qwen& qwen\\
         $\text{ArmoRM}$ (Honest)&  70B+&  mistral& qwen\\
         $\text{DPA}$&  13B+&  llama& llama\\
         $\text{DPA}$&  30B+&  llama& llama\\
         $\text{DPA}$&  70B+&  mistral& llama\\
         $\text{DPA}$&  13B+&  mistral& qwen\\
         $\text{DPA}$&  30B+&  qwen& qwen\\
         $\text{DPA}$&  70B+&  mistral& qwen\\ \bottomrule
    \end{tabular}
    \label{tab:model_fam_exps}
\end{table}

\subsection{Model Generation}

We use BenchAlign to rerank IFEval using older Llama and Qwen generations to predict future ones. Our results in Tables \ref{tab:subset_llama_qwen}, \ref{tab:subset_qwen} and \ref{tab:subset_llama} show that using models from older generations can create somewhat generalizable benchmarks for future generations. However, our results don't cover the vast majority of model families currently available nor the state-of-the-art Qwen models due to the closure of OpenLLMLeaderboard. Therefore, our findings should be used to guide future research on finding better subset rules rather than direct practical applications.

\begin{itemize}
    \item Source: Llama 2 and 3 models | Target: Llama 3.x models
    \item Source: Qwen 2 models | Target: Qwen 2.x models
    \item Source: Llama 2, 3 and Qwen 2 models | Target: Llama 3.x and Qwen 2.x models
\end{itemize}

\begin{table*}[t]
\centering
\small
\setlength{\tabcolsep}{5pt}
\renewcommand{\arraystretch}{1.05}

\resizebox{\textwidth}{!}{
\begin{tabular}{c c c c c c c c c c}
\toprule
\multirow{4}{*}{\makecell{\textbf{Target}\\\textbf{Models}}}
& \multirow{4}{*}{\textbf{Method}}
& \multicolumn{4}{c}{\textbf{Helpfulness}}
& \multicolumn{4}{c}{\textbf{Honesty}} \\
\cmidrule(lr){3-6} \cmidrule(lr){7-10}

& & \multicolumn{2}{c}{$Acc_{pair}$}
& \multicolumn{2}{c}{$\rho$}
& \multicolumn{2}{c}{$Acc_{pair}$}
& \multicolumn{2}{c}{$\rho$} \\
\cmidrule(lr){3-4} \cmidrule(lr){5-6} \cmidrule(lr){7-8} \cmidrule(lr){9-10}

& & $\text{ArmoRM}$ & $\text{GPT2}$
& $\text{ArmoRM}$ & $\text{GPT2}$
& $\text{ArmoRM}$ & $\text{DPA}$
& $\text{ArmoRM}$ & $\text{DPA}$ \\
\midrule

\multirow{6}{*}{\makecell{Llama\\+ Qwen}}
& \textsc{No Alignment}
& $0.709\pm0.002$ & $0.606\pm0.002$
& $0.597\pm0.045$ & $0.312\pm0.064$
& $0.732\pm0.002$ & $0.738\pm0.002$
& $0.654\pm0.040$ & $0.667\pm0.038$ \\

& \textsc{Random}
& $0.676\pm0.002$ & $0.605\pm0.002$
& $0.393\pm0.060$ & $0.310\pm0.064$
& $0.708\pm0.002$ & $0.711\pm0.002$
& $0.604\pm0.044$ & $0.611\pm0.044$ \\

& \textsc{Individual}
& $0.733\pm0.002$ & $0.644\pm0.002$
& $0.648\pm0.040$ & $0.397\pm0.059$
& $0.757\pm0.002$ & $0.761\pm0.002$
& $0.705\pm0.035$ & $0.712\pm0.034$ \\

& \textsc{Metabench}
& $0.708\pm0.001$ & $0.602\pm0.001$
& $0.594\pm0.047$ & $0.302\pm0.066$
& $0.731\pm0.001$ & $0.736\pm0.001$
& $0.651\pm0.042$ & $0.664\pm0.041$ \\

& \textsc{TinyBenchmarks}
& $0.697\pm0.001$ & $0.600\pm0.001$
& $0.573\pm0.048$ & $0.296\pm0.066$
& $0.720\pm0.001$ & $0.727\pm0.001$
& $0.633\pm0.043$ & $0.646\pm0.042$ \\

& \textsc{BenchAlign}
& $\mathbf{0.803\pm0.002}$ & $\mathbf{0.702\pm0.002}$
& $\mathbf{0.805\pm0.024}$ & $\mathbf{0.583\pm0.046}$
& $\mathbf{0.819\pm0.001}$ & $\mathbf{0.828\pm0.001}$
& $\mathbf{0.838\pm0.020}$ & $\mathbf{0.850\pm0.019}$ \\
\bottomrule
\end{tabular}
}

\caption{Alignment with Llama and Qwen older generation models vs newer ones}
\label{tab:subset_llama_qwen}
\end{table*}

\begin{table*}[t]
\centering
\small
\setlength{\tabcolsep}{5pt}
\renewcommand{\arraystretch}{1.05}

\resizebox{\textwidth}{!}{
\begin{tabular}{c c c c c c c c c c}
\toprule
\multirow{4}{*}{\makecell{\textbf{Target}\\\textbf{Models}}}
& \multirow{4}{*}{\textbf{Method}}
& \multicolumn{4}{c}{\textbf{Helpfulness}}
& \multicolumn{4}{c}{\textbf{Honesty}} \\
\cmidrule(lr){3-6} \cmidrule(lr){7-10}

& & \multicolumn{2}{c}{$Acc_{pair}$}
& \multicolumn{2}{c}{$\rho$}
& \multicolumn{2}{c}{$Acc_{pair}$}
& \multicolumn{2}{c}{$\rho$} \\
\cmidrule(lr){3-4} \cmidrule(lr){5-6} \cmidrule(lr){7-8} \cmidrule(lr){9-10}

& & $\text{ArmoRM}$ & $\text{GPT2}$
& $\text{ArmoRM}$ & $\text{GPT2}$
& $\text{ArmoRM}$ & $\text{DPA}$
& $\text{ArmoRM}$ & $\text{DPA}$ \\
\midrule

\multirow{6}{*}{Qwen}
& \textsc{No Alignment}
& $0.730\pm0.003$ & $0.614\pm0.004$
& $0.651\pm0.054$ & $0.323\pm0.087$
& $0.756\pm0.003$ & $0.757\pm0.003$
& $0.716\pm0.046$ & $0.721\pm0.045$ \\

& \textsc{Random}
& $0.714\pm0.003$ & $0.618\pm0.004$
& $0.394\pm0.082$ & $0.333\pm0.087$
& $0.729\pm0.003$ & $0.732\pm0.003$
& $0.661\pm0.053$ & $0.670\pm0.052$ \\

& \textsc{Individual}
& $0.767\pm0.003$ & $0.642\pm0.004$
& $0.702\pm0.048$ & $0.393\pm0.082$
& $0.779\pm0.003$ & $0.791\pm0.003$
& $0.753\pm0.041$ & $0.762\pm0.039$ \\

& \textsc{Metabench}
& $0.736\pm0.002$ & $0.616\pm0.003$
& $0.655\pm0.058$ & $0.320\pm0.091$
& $0.759\pm0.002$ & $0.762\pm0.002$
& $0.715\pm0.050$ & $0.720\pm0.049$ \\

& \textsc{TinyBenchmarks}
& $0.712\pm0.002$ & $0.611\pm0.003$
& $0.624\pm0.062$ & $0.313\pm0.091$
& $0.738\pm0.002$ & $0.739\pm0.002$
& $0.692\pm0.053$ & $0.698\pm0.052$ \\

& \textsc{BenchAlign}
& $\mathbf{0.839\pm0.003}$ & $\mathbf{0.684\pm0.003}$
& $\mathbf{0.858\pm0.024}$ & $\mathbf{0.546\pm0.067}$
& $\mathbf{0.856\pm0.003}$ & $\mathbf{0.864\pm0.003}$
& $\mathbf{0.892\pm0.019}$ & $\mathbf{0.904\pm0.017}$ \\
\bottomrule
\end{tabular}
}

\caption{Alignment with Qwen older generation models vs newer ones}
\label{tab:subset_qwen}
\end{table*}

\begin{table*}[h]
\centering
\small
\setlength{\tabcolsep}{5pt}
\renewcommand{\arraystretch}{1.05}

\resizebox{\textwidth}{!}{
\begin{tabular}{c c c c c c c c c c}
\toprule
\multirow{4}{*}{\makecell{\textbf{Target}\\\textbf{Models}}}
& \multirow{4}{*}{\textbf{Method}}
& \multicolumn{4}{c}{\textbf{Helpfulness}}
& \multicolumn{4}{c}{\textbf{Honesty}} \\
\cmidrule(lr){3-6} \cmidrule(lr){7-10}

& & \multicolumn{2}{c}{$Acc_{pair}$}
& \multicolumn{2}{c}{$\rho$}
& \multicolumn{2}{c}{$Acc_{pair}$}
& \multicolumn{2}{c}{$\rho$} \\
\cmidrule(lr){3-4} \cmidrule(lr){5-6} \cmidrule(lr){7-8} \cmidrule(lr){9-10}

& & $\text{ArmoRM}$ & $\text{GPT2}$
& $\text{ArmoRM}$ & $\text{GPT2}$
& $\text{ArmoRM}$ & $\text{DPA}$
& $\text{ArmoRM}$ & $\text{DPA}$ \\
\midrule

\multirow{6}{*}{Llama}
& \textsc{No Alignment}
& $0.727\pm0.004$ & $0.676\pm0.004$
& $0.623\pm0.060$ & $0.510\pm0.073$
& $0.736\pm0.003$ & $0.740\pm0.003$
& $0.648\pm0.057$ & $0.649\pm0.057$ \\

& \textsc{Random}
& $0.714\pm0.004$ & $0.661\pm0.004$
& $0.516\pm0.073$ & $0.445\pm0.080$
& $0.699\pm0.004$ & $0.703\pm0.004$
& $0.563\pm0.067$ & $0.569\pm0.067$ \\

& \textsc{Individual}
& $0.741\pm0.003$ & $0.691\pm0.004$
& $0.643\pm0.058$ & $0.535\pm0.071$
& $0.750\pm0.003$ & $0.756\pm0.003$
& $0.665\pm0.054$ & $0.671\pm0.054$ \\

& \textsc{Metabench}
& $0.726\pm0.002$ & $0.673\pm0.003$
& $0.624\pm0.064$ & $0.506\pm0.078$
& $0.735\pm0.002$ & $0.737\pm0.002$
& $0.650\pm0.061$ & $0.649\pm0.061$ \\

& \textsc{TinyBenchmarks}
& $0.710\pm0.002$ & $0.669\pm0.003$
& $0.588\pm0.067$ & $0.489\pm0.078$
& $0.717\pm0.002$ & $0.723\pm0.002$
& $0.612\pm0.064$ & $0.615\pm0.064$ \\

& \textsc{BenchAlign}
& $\mathbf{0.795\pm0.003}$ & $\mathbf{0.748\pm0.003}$
& $\mathbf{0.777\pm0.038}$ & $\mathbf{0.694\pm0.051}$
& $\mathbf{0.805\pm0.003}$ & $\mathbf{0.811\pm0.003}$
& $\mathbf{0.801\pm0.035}$ & $\mathbf{0.814\pm0.033}$ \\
\bottomrule
\end{tabular}
}

\caption{Alignment with Llama older generation models vs newer ones}
\label{tab:subset_llama}
\end{table*}

\clearpage
\section*{NeurIPS Paper Checklist}

\begin{enumerate}

\item {\bf Claims}
    \item[] Question: Do the main claims made in the abstract and introduction accurately reflect the paper's contributions and scope?
    \item[] Answer: \answerYes{} 
    \item[] Justification: Yes, the claims made in the abstract and introduction are accurately reflected throughout the paper.

\item {\bf Limitations}
    \item[] Question: Does the paper discuss the limitations of the work performed by the authors?
    \item[] Answer: \answerYes{} 
    \item[] Justification: Yes, please see Section \ref{sec:limitations} for the limitations.

\item {\bf Theory assumptions and proofs}
    \item[] Question: For each theoretical result, does the paper provide the full set of assumptions and a complete (and correct) proof?
    \item[] Answer: \answerNA{} 
    \item[] Justification: We do not include theoretical results.

    \item {\bf Experimental result reproducibility}
    \item[] Question: Does the paper fully disclose all the information needed to reproduce the main experimental results of the paper to the extent that it affects the main claims and/or conclusions of the paper (regardless of whether the code and data are provided or not)?
    \item[] Answer: \answerYes{} 
    \item[] Justification: We explain the datasets, model architecture and hyperparameters for our experiments in Section \ref{sec:experiments}. Code will be released following the review process.

\item {\bf Open access to data and code}
    \item[] Question: Does the paper provide open access to the data and code, with sufficient instructions to faithfully reproduce the main experimental results, as described in supplemental material?
    \item[] Answer: \answerYes{} 
    \item[] Justification: We use datasets and reward models publicly available. Information to reproduce our experiments can be found in Section \ref{sec:experiments}.

\item {\bf Experimental setting/details}
    \item[] Question: Does the paper specify all the training and test details (e.g., data splits, hyperparameters, how they were chosen, type of optimizer) necessary to understand the results?
    \item[] Answer: \answerYes{} 
    \item[] Justification: We specify all data splits, hyperparameters and optimizer in Section \ref{sec:experiments}.

\item {\bf Experiment statistical significance}
    \item[] Question: Does the paper report error bars suitably and correctly defined or other appropriate information about the statistical significance of the experiments?
    \item[] Answer: \answerYes{} 
    \item[] Justification: All of our experiments report error bars where applicable.

\item {\bf Experiments compute resources}
    \item[] Question: For each experiment, does the paper provide sufficient information on the computer resources (type of compute workers, memory, time of execution) needed to reproduce the experiments?
    \item[] Answer: \answerYes{} 
    \item[] Justification: Please see appendix \ref{ap:runtime_analysis}.
    
\item {\bf Code of ethics}
    \item[] Question: Does the research conducted in the paper conform, in every respect, with the NeurIPS Code of Ethics \url{https://neurips.cc/public/EthicsGuidelines}?
    \item[] Answer: \answerYes{} 
    \item[] Justification: We followed the NeurIPS Code of Ethics.

\item {\bf Broader impacts}
    \item[] Question: Does the paper discuss both potential positive societal impacts and negative societal impacts of the work performed?
    \item[] Answer: \answerYes{} 
    \item[] Justification: Please see Appendix \ref{ap:broader_impacts}.
    
\item {\bf Safeguards}
    \item[] Question: Does the paper describe safeguards that have been put in place for responsible release of data or models that have a high risk for misuse (e.g., pre-trained language models, image generators, or scraped datasets)?
    \item[] Answer: \answerNA{} 
    \item[] Justification: We do not introduce data or models that pose a high risk of misuse.

\item {\bf Licenses for existing assets}
    \item[] Question: Are the creators or original owners of assets (e.g., code, data, models), used in the paper, properly credited and are the license and terms of use explicitly mentioned and properly respected?
    \item[] Answer: \answerYes{} 
    \item[] Justification: The creators or original asset owners have been credited in appropriate ways.

\item {\bf New assets}
    \item[] Question: Are new assets introduced in the paper well documented and is the documentation provided alongside the assets?
    \item[] Answer: \answerNA{} 
    \item[] Justification: In this work, no new assets are released.

\item {\bf Crowdsourcing and research with human subjects}
    \item[] Question: For crowdsourcing experiments and research with human subjects, does the paper include the full text of instructions given to participants and screenshots, if applicable, as well as details about compensation (if any)? 
    \item[] Answer: \answerNA{} 
    \item[] Justification: This work does not involve crowdsourcing nor research with human subjects.

\item {\bf Institutional review board (IRB) approvals or equivalent for research with human subjects}
    \item[] Question: Does the paper describe potential risks incurred by study participants, whether such risks were disclosed to the subjects, and whether Institutional Review Board (IRB) approvals (or an equivalent approval/review based on the requirements of your country or institution) were obtained?
    \item[] Answer: \answerNA{} 
    \item[] Justification: This work does not involve crowdsourcing nor research with human subjects.

\item {\bf Declaration of LLM usage}
    \item[] Question: Does the paper describe the usage of LLMs if it is an important, original, or non-standard component of the core methods in this research? Note that if the LLM is used only for writing, editing, or formatting purposes and does \emph{not} impact the core methodology, scientific rigor, or originality of the research, declaration is not required.
    \item[] Answer: \answerNA{} 
    \item[] Justification: The core methodology, scientific rigorousness, or originality of the research does not involve the usage of LLMs.

\end{enumerate}

\end{document}